\pdfoutput=1

\documentclass[11pt]{article}

\usepackage[final]{acl}

\usepackage{times}
\usepackage{latexsym}

\usepackage[T1]{fontenc}

\usepackage[utf8]{inputenc}

\usepackage{microtype}

\usepackage{inconsolata}

\usepackage{graphicx}

\usepackage{hyperref}
\usepackage{url}

\usepackage{algorithm}
\usepackage{algorithmic}

\usepackage{float}
\usepackage{graphicx}
\usepackage{amsmath}
\usepackage{amssymb}
\usepackage{booktabs}

\usepackage{makecell}
\usepackage{colortbl}
\usepackage{xcolor}

\usepackage{graphicx}
\usepackage{multicol}
\usepackage{multirow}
\usepackage{blindtext}
\usepackage{dirtree}
\usepackage{tikz}

\usepackage{amsthm}
\usepackage{pbox}
\usepackage{chngpage}
\usepackage{minitoc}
\usepackage{longtable}

\usepackage{amsmath, amsfonts, amsthm, stackrel, amssymb} 

\usepackage{tcolorbox}
\usepackage{caption}
\usepackage{subcaption}
\usepackage{hyperref}

\usepackage{graphicx}
\usepackage{multicol}
\usepackage{multirow}
\usepackage{blindtext}
\usepackage{dirtree}
\usepackage{tikz}

\usepackage{amsmath}
\usepackage{amsthm}
\usepackage{pbox}
\usepackage{chngpage}
\usepackage{minitoc}
\usepackage{longtable}

\usepackage{booktabs} 
\usepackage{multirow} 
\usepackage{graphicx}

\newtheorem{definition}{Definition}

\title{GraphInsight: Unlocking Insights in Large Language Models\\ for Graph Structure Understanding}

\author{
	Yukun Cao$^{1,2}$\thanks{These authors contributed equally to this work.} \quad Shuo Han$^{1,2}$\footnotemark[1] \quad Zengyi Gao$^{1,2}$ \\
	\textbf{Zezhong Ding}$^{1,2}$ \quad \textbf{Xike Xie}$^{1,2}$\thanks{Corresponding author} \quad \textbf{S. Kevin Zhou}$^{1,3}$ \\
	$^{1}$ University of Science and Technology of China, China \\
	$^{2}$ Data Darkness Lab, MIRACLE Center, USTC, China \\
	$^{3}$ MIRACLE Center, USTC, China \\
	\texttt{\{ykcho, shuo.han, gzy02, zezhongding\}@mail.ustc.edu.cn} \\
	\texttt{\{xkxie, skevinzhou\}@ustc.edu.cn}
}

\begin{document}
\maketitle

\begin{abstract}

	Although Large Language Models (LLMs) have demonstrated potential in processing graphs, they struggle with comprehending graphical structure information through prompts of graph description sequences, especially as the graph size increases. We attribute this challenge to the uneven memory performance of LLMs across different positions in graph description sequences, known as ``Positional bias''. To address this, we propose {\bf GraphInsight}, a novel framework aimed at improving LLMs' comprehension of both macro- and micro-level graphical information. GraphInsight is grounded in two key strategies: 1) placing critical graphical information in positions where LLMs exhibit stronger memory performance, and 2) investigating a lightweight external knowledge base for regions with weaker memory performance, inspired by retrieval-augmented generation (RAG). Moreover, GraphInsight explores integrating these two strategies into LLM agent processes for composite graph tasks that require multi-step reasoning. Extensive empirical studies on benchmarks with a wide range of evaluation tasks show that GraphInsight significantly outperforms all other graph description methods (e.g., prompting techniques and reordering strategies) in understanding graph structures of varying sizes.
\end{abstract}

\section{Introduction}

Large language models (LLMs) have demonstrated remarkable capabilities in natural language processing (NLP) \cite{shen2024understanding, naveed2023comprehensive, DBLP:conf/nips/GeHMJTXLZ23}, enabling their initial applications across various data domains, such as graphs \cite{chen2024exploring,DBLP:conf/aaai/WangCOWHSGXZCLZ24, DBLP:conf/aaai/BestaBKGPGGLNNH24}, time-series data \cite{jin2023time, yu2023temporal}, tabular data \cite{sui2024table, hegselmann2023tabllm}, and other structured or semi-structured data types \cite{ko2024filling, perozzi2024let}.
Among these domains, leveraging LLMs to tackle applications involving graphs has emerged as a burgeoning field of research, as graphs represent fundamental structures that capture intricate relationships and interactions in the real world \cite{wang2021learning, xu2021understanding}.
For example, \citeauthor{fatemi2023talk} have explored the potential of LLMs by converting various types of graphs, such as knowledge graphs \cite{Baek2023, DBLP:journals/tkde/PanLWCWW24} and social network graphs \cite{DBLP:conf/sigir/Santra24,  DBLP:conf/ekgllm/Babic23}, into natural language descriptions, thereby enabling LLMs to perform question-answering tasks related to these graphs.

\setlength{\textfloatsep}{5pt}
\begin{figure}[t]
	\centering
	\includegraphics[width=\columnwidth]{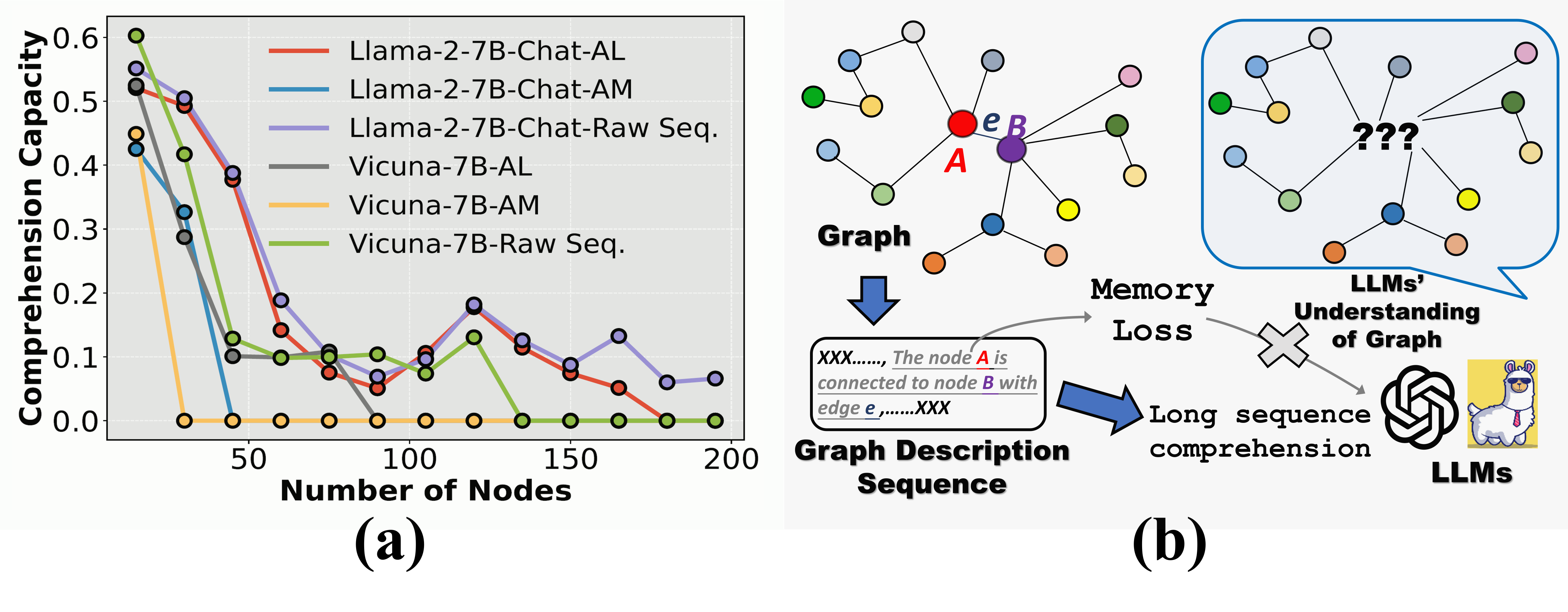}
	\vspace{-10mm}
	\caption{\small Capabilities of LLMs on Graph Structure Understanding}
	\label{fig:graphtollms}
\end{figure}

A {\bf key observation} is that enhancing LLM performance in graph-related applications depends critically on LLMs' ability to comprehend graph structures through natural language descriptions.
Existing studies \cite{shang2024survey, DBLP:journals/corr/abs-2311-12399} primarily utilize two direct methods to transform graphs into text inputs for LLMs: the {\it structural format transforming}, such as adjacency matrices (termed as {\it AM}) or lists (termed as {\it AL}) and the {\it sequential format transforming}, such as edge-by-edge descriptions (termed as {\it Raw Seq}).
However, extensive empirical studies \cite{yuan2024gracore} have shown that LLMs face significant challenges in understanding and reasoning about graph structures using current graph transformation methods, especially as graph size increases, leading to a ``comprehension collapse''.
As shown in Figure 1 (a), several common LLMs perform poorly on graph structure understanding tasks (see benchmarks in Section~\ref{exp:setup}), and their comprehension declines sharply as the graph size increases, ultimately leading to complete failure.

In this paper, we focus on enhancing the ability of LLMs to understand graph structures by developing new graph-transforming methods and other novel techniques. To achieve this, we first analyze the inherent nature of graph understanding tasks in LLMs and the reasons behind the poor performance of existing methods. At its core, the challenge of LLM understanding graph structures can be viewed as a \emph{stringent long sequence comprehension problem} \cite{liu2024lost, wang2024}.
This challenge arises primarily due to two factors, as illustrated in Figure 1(b). First, graph structural information is conveyed to LLMs solely through language descriptions. As the graph size increases, their descriptions become longer, challenging LLMs' ability to comprehend long-sequence inputs.                                                                                                                          Second, to accurately understand the graph's structure, LLMs must retain every detail in the natural language description. Any memory lapse, especially regarding critical nodes or edges (e.g., central nodes, bridging edges), notably impairs the LLMs' ability to infer the correct structure from the description.

However, the challenge is further exacerbated by the uneven memory performance of LLMs across different positions within long sequences—a phenomenon known as \emph{``positional bias''} \cite{tang2023found, hsieh2024found, zhang2024attention}, which mainly stems from limitations in the attention mechanism \cite{xiao2023efficient} and the internal memory capacity \cite{an2024make} of LLMs. For example, as shown in Figure~2(a), many studies indicate that LLMs generally perform better at understanding the head and tail of sequences (i.e., {\it strong memory regions}), with a noticeable dip in performance in middle parts (i.e., {\it weak memory regions}), termed as \emph{``lost-in-the-middle''} \cite{liu2024lost}. Consequently, LLMs struggle to meet the stringent requirements for understanding graph structures from descriptive sequences, due to they always lose some positions in the sequence. This issue becomes particularly pronounced as the graph size increases (i.e., sequence lengthens),  leading to ``comprehension collapse,'' as mentioned in Figure 1(a).

\setlength{\textfloatsep}{5pt} %
\begin{figure}[t]
	\centering
	\includegraphics[width=\columnwidth]{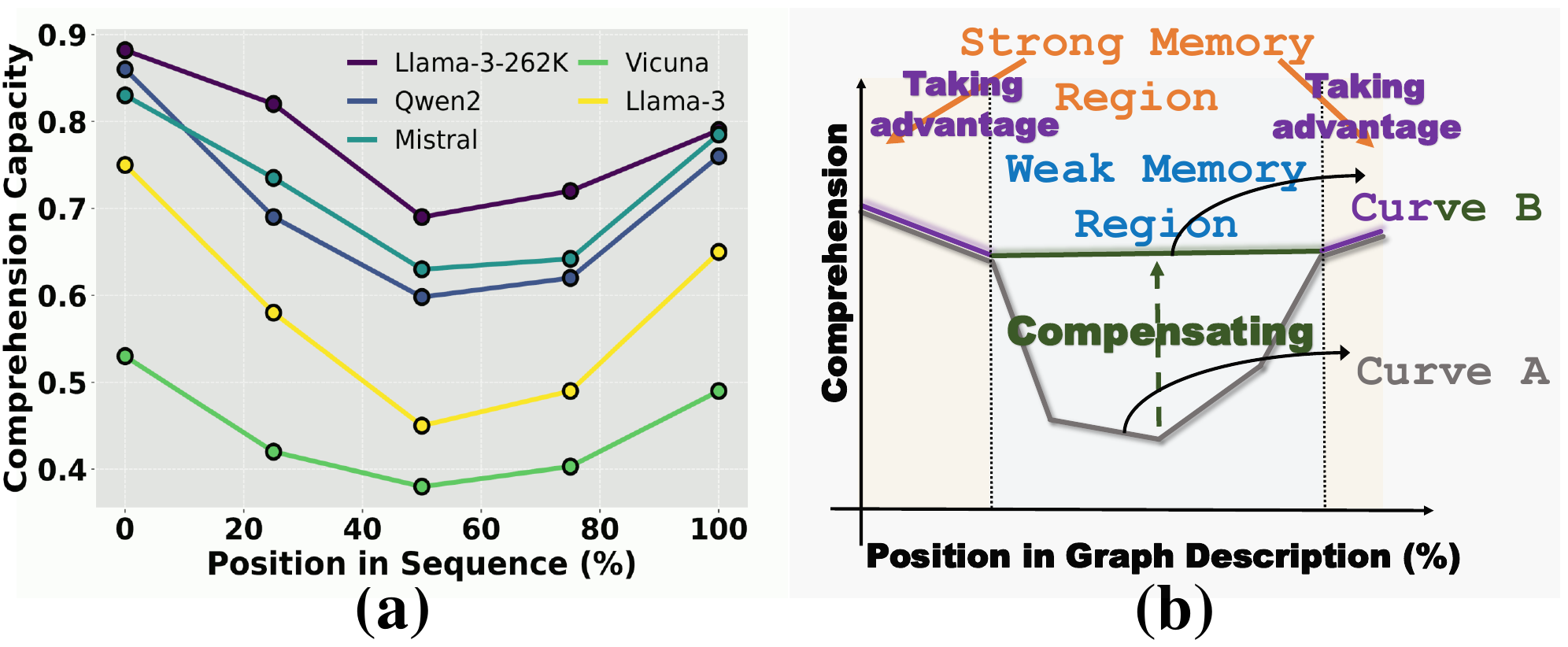}
	\vspace{-8mm}
	\caption{Analysis on Positional bias of LLMs}
	\label{fig:graphtollms}
\end{figure}

Thus, improving LLMs' ability to comprehend graph structures relies on improving their ability to retain information across different sequence positions in graph descriptions, thereby reducing the impact of positional bias.
To this end, our work is grounded in two high-level insights: \textbf{1) Taking advantage of the Strong Memory Regions:} For sequence positions where LLMs demonstrate strong memory capabilities, we strategically place descriptions of critical nodes or edges in these locations. \textbf{2) Compensating for the Weak Memory Regions:} we construct a lightweight external knowledge base for positions where LLMs show weak memory performance, inspired by the concept of retrieval-augmented generation (RAG).
As illustrated in Figure 1(b), following the above ``{\it different horses for different courses}'' principle, the LLMs' comprehension of graph description sequences improves from Curve $A$ to Curve $B$.

Starting from the premises, we propose \textit{GraphInsight}, a framework aimed at enhancing the ability of LLMs in graph comprehension tasks through a series of innovative optimization techniques applied to graph description sequences. 
Our framework incorporates two key techniques tailored for macro-level and micro-level graph understanding tasks, respectively.
Specifically, For Macro-level tasks (i.e., issues related to global graph structures), we  reconstruct the original graph description sequence according to the relative importance of local graph structures, ensuring that critical graph components are aligned with the LLMs' strong memory regions, thereby improving the overall memory retention and understanding of the graph.
For Micro-level tasks (i.e., issues related to local detailed graph structures), we  build a lightweight knowledge base for the graph description sequences corresponding to the LLMs' weak memory regions, enhancing the LLMs' comprehension in fine-grained graph tasks by enabling efficient retrieval of relevant information.
Moreover, we explore integrating two techniques into LLMs agent processes to tackle composite  graph understanding tasks that require multi-step reasoning, which involves multiple interconnected micro-level tasks.

\newcommand{\squishlist}{
	\begin{list}{$\bullet$}
		{   \setlength{\itemsep}{0pt}
			\setlength{\parsep}{3pt}
			\setlength{\topsep}{3pt}
			\setlength{\partopsep}{0pt}
			\setlength{\leftmargin}{1.5em}
			\setlength{\labelwidth}{1em}
			\setlength{\labelsep}{0.5em} } }
	\newcounter{Lcount}
	\newcommand{\squishlisttwo}{
		\begin{list}{\arabic{Lcount}. }
			{ \usecounter{Lcount}
				\setlength{\itemsep}{0pt}
				\setlength{\parsep}{0pt}
				\setlength{\topsep}{0pt}
				\setlength{\partopsep}{0pt}
				\setlength{\leftmargin}{2em}
				\setlength{\labelwidth}{1.5em}
				\setlength{\labelsep}{0.5em} } }
		\newcommand{\squishend}{\end{list} }

The main contributions of this paper are listed as follows:
\begin{list}{$\bullet$}
	{   \setlength{\itemsep}{0pt}
		\setlength{\parsep}{3pt}
		\setlength{\topsep}{3pt}
		\setlength{\partopsep}{0pt}
		\setlength{\leftmargin}{1.5em}
		\setlength{\labelwidth}{1em}
		\setlength{\labelsep}{0.5em} }
	\item We conduct a pioneering analysis of the fundamental issues, challenges, and high-level solutions for LLMs in understanding graph structures based on natural language graph descriptions.
	\item To the best of our knowledge, from the perspective of the ``positional bias'' inherent in LLMs, we propose the first framework, {\bf GraphInsight}, designed to enhance LLMs' ability to understand graph structures, integrating a series of innovative techniques.
	\item We introduce \textbf{GraphSQA}, a benchmark designed to evaluate LLMs' ability to understand and reason about graph structures across two levels of tasks, from macro-level to micro-level comprehension.
	\item Extensive empirical studies across various evaluation tasks and LLMs demonstrate the effectiveness and superiority of our framework.
\end{list}


\vspace{-2.5mm}
\section{Related Work}
\vspace{-1.5mm}

\subsection{LLMs' Understanding of Graph Structures Through Description Sequences}
\vspace{-0.5mm}
Research on the ability of LLMs to understand graph structures by inputting natural language descriptions of graphs into these models has become an emerging area of study. Existing research has focused on developing evaluation benchmarks for graph structure understanding tasks and analyzing their results. For instance, \citeauthor{wang2024can} and \citeauthor{guo2023gpt4graph} conducted preliminary empirical assessments of LLMs on coarse-grained graph structure tasks, indicating the nascent stage of this field. Subsequent benchmarks, such as GraphEval2000 \cite{wu2024grapheval2000}, GraphArena \cite{tang2024grapharena}, GraphInstruct \cite{luo2024graphinstruct}, and GraCoRe \cite{yuan2024gracore}, have shifted focus to fine-grained tasks, including graph theory problems. \citeauthor{ge2024sequential} also explored the impact of the sequence of graph descriptions on LLMs comprehension, though their study was limited to a few tasks and lacked deep analysis. Unlike previous work, GraphInsight aims to enhance LLMs' capabilities in graph structure understanding tasks.

\vspace{-0.5mm}
\subsection{Positional Bias in Long Sequences for LLMs}
\vspace{-0.25mm}
Extensive research \cite{tang2023found, hsieh2024found} has shown that the position of inputs and the order of answer choices can significantly impact LLMs' performance and output generation (i.e., ``positional bias''). \citeauthor{liu2024lost} analyzed how LLMs utilize information in long input sequences and identified the ``lost-in-the-middle'' problem. Additionally, \citeauthor{xiao2023efficient} discovered that attention scores tend to be biased towards initial tokens of  input sequences  due to the Softmax operation.

Some recent works have attempted to mitigate the positional bias issue in LLMs. For instance, \citeauthor{zhang2024attention, hsieh2024found} introduced attention instructions to guide the LLMs to focus on specific sequence segments.  \citeauthor{an2024make} proposed training with a synthesized long-sequence QA dataset to mitigate the "lost-in-the-middle" problem. \citeauthor{wu2024never} proposed interpolating positional encodings via index adjustment, requiring fine-tuning within the pre-trained context window. Note that these broad strategies are not specifically designed for graphs and do not directly apply to graph description sequences; instead, they are generally implemented during the training/fine-tuning phases of LLMs. Therefore, these methods are orthogonal to GraphInsight and could further enhance its performance if computational resources permit.

\vspace{-3mm}
\section{Preliminaries}  
\label{sec:preliminaries}  
\vspace{-2mm}
In this section, we define the basic format of graph description sequences, outline our assessment of LLM capabilities on these sequences, and introduce two levels of graph understanding tasks as the foundation of our framework.


There are two primary methods for converting graphs into description sequences for LLMs: {\bf 1)} structural format transforming (e.g., adjacency matrices/lists) and {\bf 2)} sequential format transforming  (e.g., edge-by-edge descriptions). Since empirical studies (see Section~\ref{exp:performance}) show that sequential format is more conducive to LLMs' understanding  and can seamlessly integrate additional semantic information (e.g., node and edge attributes, labels, etc.), our framework focuses on optimizing graph description sequences under this format. Therefore, for a given graph \( G \), we define the standard graph description sequences  $\mathcal{T}$ as follows:
\vspace{-2.5pt}
\begin{definition}[\small \textbf{Sequential Format Graph Description}]
	For a graph $G$, consisting of $V = \{v_1, v_2, ...\}$ and $E = \{e_{ij}\}$. The sequential format transforming is to transform $G$ to $\mathcal{T}$:
\vspace{-5pt}
\begin{tcolorbox}[colback=white, colframe=black, width=\linewidth, boxrule=0.5pt, rounded corners, left=1mm, right=1mm, top=1mm, bottom=1mm, boxsep=1mm]
	\small
	\textit{This graph is described as follows:}\\
	\hspace*{1em}\textit{
		Node $v_i$ is connected to node $v_j$ by edge $e_{ij}$ with weight $w_{ij}$;
		\ldots}
\end{tcolorbox}
	\label{def:trans}
\end{definition}
\vspace{-8pt}

The description $\mathcal{T}$ consists of the descriptions of all edges $e_{ij} \in E$. Based on Definition~\ref{def:trans}, the LLMs' ability to understand $\mathcal{T}$ can be quantified as shown in  Definition~\ref{def:LLMcapacity}.

\vspace{-4pt}
\begin{definition}[\small \bf LLMs' Capacity for Graph Understanding]
	\label{def:LLMcapacity}
	The capacity of LLM for graph understanding can be quantified by $C_{{\text{LLM}}}(\mathcal{T}) = \int C(\mathcal{T}, p) \, dp$, where $C(\mathcal{T}, p)$ is the comprehension ability at a specific position $p$ within $\mathcal{T}$.
	
\end{definition}	

\vspace{-3pt}

Thus, the \(C(\mathcal{T}, p)\) at position \(p\), can be modeled as following a specific distribution $\Psi(p)$:
\begin{equation}
\setlength\abovedisplayskip{3pt}
\setlength\belowdisplayskip{3pt}
	C(\mathcal{T}, p) \sim \Psi(p)
	\label{eq:comprehension_ability}
\end{equation}
Here, \(\Psi(p)\) denotes the positional bias curve, which represents the distribution of an LLM's inherent comprehension at position \(p\).
As mentioned in Figure~\ref{fig:graphtollms}, \(\Psi(p)\) typically displays stronger comprehension at the head and tail of a sequence, with weaker comprehension in the middle,  forming a U-shape curve.
In this paper, we follow the above assumption and define the positions corresponding to the head \( \alpha\% \) and tail \( \beta\% \) of the sequence as strong memory regions, and the rest as weak memory regions.

Next, we introduce two levels of graph understanding tasks, as shown in Definition~\ref{def:gut}.

\vspace{-3.5pt}
\begin{definition}[\small \bf Graph Understanding Tasks]
	\label{def:gut}
	Integrating on existing benchmarks~\cite{wang2024can,yuan2024gracore}, we categorize LLM graph structure understanding into two levels: \textbf{1) Macro-level}, involving coarse-grained reasoning related to the overall graph structure (e.g., node counting, connectivity detection, cycle detection); \textbf{2) Micro-level}, focusing on fine-grained reasoning related to local structures (e.g., direct connection detection, node degree calculation, leaf node identification, neighbor recognition). Moreover, some composite graph understanding tasks may require multiple micro-level reasoning steps (e.g., complete subgraph verification, and third-order neighbor identification).	
\end{definition}
\vspace{-4pt}
Following Definition~\ref{def:gut}, we build the GraphSQA benchmark (see Section~\ref{exp:setup}) to comprehensively and fairly evaluate LLMs' performance in graph structure understanding.

\vspace{-4.5pt}
\section{Methodology} \label{Methodology}
\vspace{-3.5pt}

\setlength{\textfloatsep}{5pt} %
\begin{figure}[t]
	\centering
	\includegraphics[width=1\columnwidth]{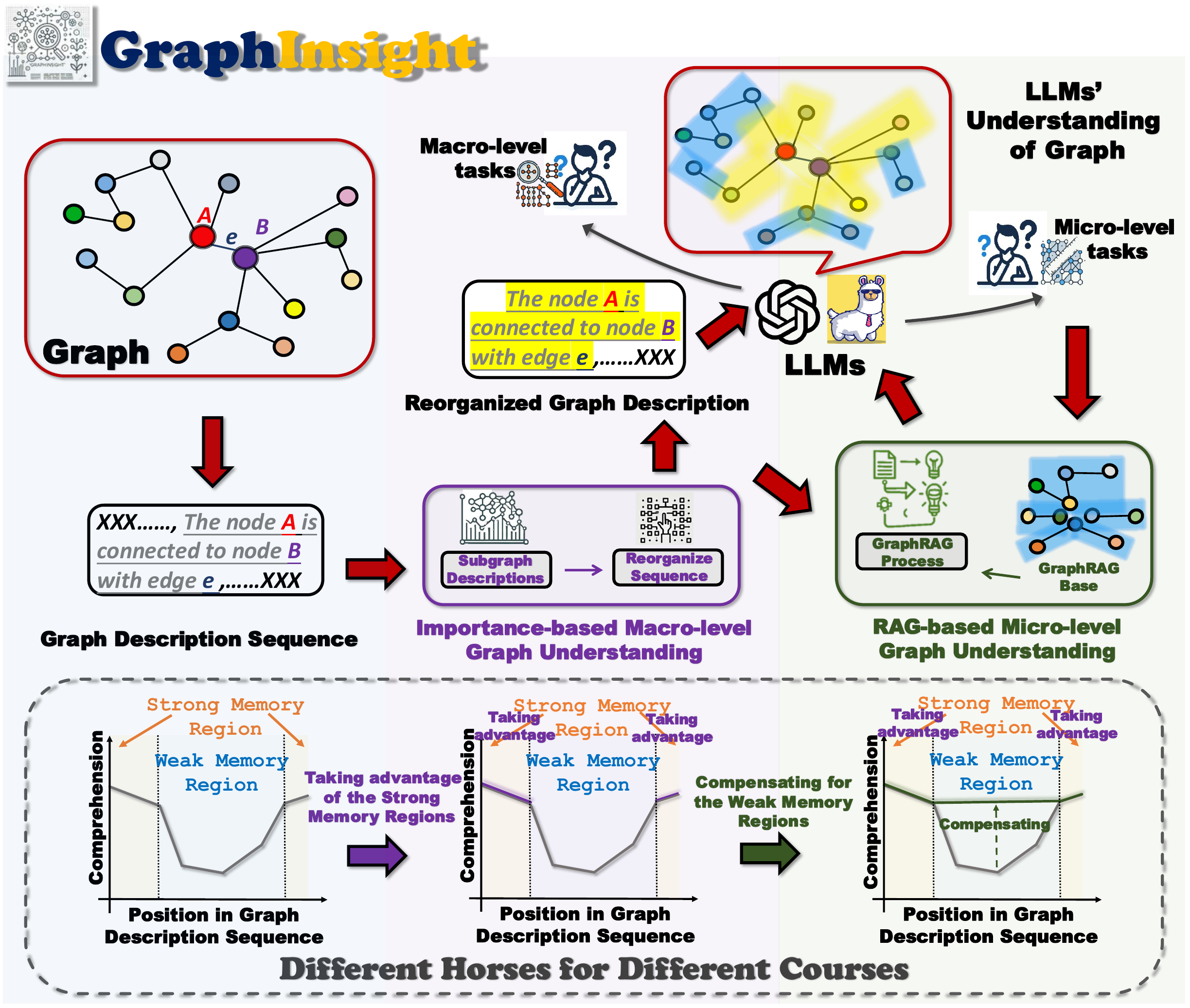}
	\vspace{-7.5mm}
	\caption{Framework of GraphInsight.}
	\label{fig:GI}
\end{figure}

\vspace{-4pt}

\subsection{Overview}
\vspace{-3pt}

In this section, we present the \emph{GraphInsight} framework, as shown in Figure~\ref{fig:GI}, which consists of two key techniques that enhance LLMs' graph comprehension.

\vspace{-3pt}
\subsection{Importance-based Macro-level Graph Understanding}
\label{sec:ibmgu}
\vspace{-2.5pt}

For macro-level graph understanding tasks, guiding LLMs to focus on the memory and comprehension of key elements within the graph structure can potentially enhance their reasoning performance. Thus, the intuitive idea is to align the LLMs' comprehension distribution across different sequence positions (i.e., $\Psi(p)$) with the importance distribution of these positions in the graph description sequence.
Let $\Phi(p)$ represent the importance of each corresponding position $p$ in the sequence $\mathcal{T}$ relative to the graph structure. In this context, our objective is:
\begin{equation}
\small
\setlength\abovedisplayskip{3pt}
\setlength\belowdisplayskip{1.5pt}
	\text{argmin}_{\Psi} \, D_{\text{KL}} \left(\Phi(p) \parallel \Psi(p)\right), \quad \text{for all } p \in \mathcal{T}
	\label{eq:alignment_kl}
\end{equation}
where \(D_{\text{KL}}\) represents the Kullback-Leibler divergence, quantifying the alignment between \(\Phi(p)\) and \(\Psi(p)\).

From theory to practice, we need to address two key challenges: 1) defining and quantifying \(\Phi(p)\); and 2) aligning \(\Phi(p)\) with \(\Psi(p)\). Given that in this study we assume the comprehension distribution of LLMs follows a U-shaped curve~\footnote{Advancements in pre-training techniques and corpus optimization may alter the exact shape of this curve. Nevertheless, the variation in LLM comprehension across different sequence positions is likely to persist. Our framework is not restricted to a U-shaped curve; once the precise curve shape of a particular LLM is empirically estimated, our framework can be easily adjusted and adapted.}, we adopt a straightforward approach to achieve this alignment.
Overall, we decompose the graph description $\mathcal{T}$ into a series of mutually exclusive subgraph descriptions  $\mathcal{T}_{s}:\{t_1, t_2, ...\}$, with the importance of each subgraph $	\mathcal{I}(t_i)$ measured by the PageRank score (calculated over the entire graph) of its highest-degree node. Finally, we reorder these subgraph descriptions based on their importance and reorganize them within the strong memory regions of the LLMs (i.e., the head and tail of the graph description), resulting in a new graph description sequence $\mathcal{\hat{T}}$.

Specifically, for the graph description decomposition, given a graph \( G \) and its corresponding graph description sequence \( \mathcal{T} \), we first iteratively calculate the PageRank score \( \text{PR}(v) \)\footnote{PR can be replaced with others describe the importance of graph structures. Designing graph structure importance remains an open problem \cite{openproblem,  DBLP:journals/entropy/LiuG23a,  DBLP:conf/cikm/GengWZMDJG022}, and is orthogonal to our work.}  for each node \( v \in V \) in \( G \) as follows:
\begin{equation}
\setlength\abovedisplayskip{2.5pt}
\setlength\belowdisplayskip{2pt}
	\small
	\text{PR}(v) = \lambda \sum_{u \in \text{InNb}(v)} \frac{\text{PR}(u)}{\text{OutDeg}(u)} + (1-\lambda)\frac{1}{|V|}
	\label{eq:pr}
\end{equation}
where \( \lambda \) is the damping factor, typically set to 0.85, \( \text{InNb}(v) \) represents the set of nodes with edges directed towards \( v \), and \( \text{OutDeg}(u) \) is the number of edges leaving node \( u \).

After calculating the PageRank scores, we sort the nodes in descending order. Starting from the node with the highest score, we iteratively construct subgraphs centered on each node, including directly connected neighbors with lower degrees, provided the connecting edges haven't been used in other subgraphs.
This ensures that each edge is only included in one subgraph, guaranteeing that the total length of the merged subgraph descriptions equals the original graph description sequence. This is because, as per Definition 1, we describe the graph edge by edge. The descriptions for these subgraphs \( G_1, G_2, \dots, G_k \) are denoted as \( t_1, t_2, \dots, t_k \).
Then, each subgraph description \( t_i \) is formed by combining the descriptions of all edges within the subgraph \( G_i \), according to the graph description defined previously in Definition~\ref{def:trans}. The importance \( \mathcal{I}(t_i) \) of each subgraph description \( t_i \) is defined as the PageRank score of its central node \( v_{c_i} \): $\mathcal{I}(t_i) = \text{PR}(v_{c_i})$.

Finally, the subgraph descriptions are reordered based on their importance and organized within the strong memory regions of the LLM, as defined in Definition~\ref{def:LLMcapacity}. Specifically, the head \( \alpha\% \) and the tail \( \beta\% \) of the graph description sequence are designated as the strong memory regions. Consequently, the most important subgraph descriptions are prioritized and placed in these regions, while the remaining descriptions, sorted by importance, occupy the middle and are considered the weak memory region. The final graph description sequence \( \mathcal{\hat{T}} \) is:
\begin{equation}
\setlength\abovedisplayskip{3pt}
\setlength\belowdisplayskip{3pt}
	\label{eq:reorg}
		\small
	\mathcal{\hat{T}} = \mathcal{T}[:\alpha\%] \cup \mathcal{T}[\alpha\%:(100-\beta)\%] \cup \mathcal{T}[(100-\beta)\%:]
\end{equation}
where  \( \mathcal{T}[:\alpha\%] \) and \( \mathcal{T}[(100-\beta)\%:] \) contain the most important subgraph descriptions.

\setlength{\textfloatsep}{0pt} 
\begin{algorithm}[t]
	\footnotesize 
	\captionsetup{font=footnotesize} 
	\caption{Importance-based Description Reorganization}
	\label{alg:ibdr}
	\begin{algorithmic}[1]
		\REQUIRE Graph \( G = (V, E) \) and corresponding description \( \mathcal{T} \), \( \lambda = 0.85 \), Memory regions \( \alpha\% \), \( \beta\% \)
		\ENSURE Reorganized sequence \( \mathcal{\hat{T}} \)
		
		\STATE \textbf{1. Compute PageRank:}
		\FOR{each \( v \in V \)}
		\STATE Calculate \(\text{PR}(v)\) based on Equation~\ref{eq:pr}
		\ENDFOR
		\STATE Sort nodes by \( \text{PR}(v) \) in descending order, and get sort list $V_s$
		
		\STATE \textbf{2.  Get Subgraph Descriptions :}
		\STATE Initialize \( E_{\text{used}} = \emptyset \), $\mathcal{T}_{s}= \emptyset$
		\FOR{each \( v_{c_i} \) in sorted list $V_s$}
		\STATE Initial subgraph \(G_i:\{V_i, E_i\} \)
		\FOR{each edge \( (v_{c_i}, u) \) where \( u \in \text{Neighbors}(v_{c_i}) \)}
		\IF{\( (v_{c_i}, u) \notin E_{\text{used}} \)}
		\STATE \( E_i \) $\gets$ \( E_i  \cup   (v_{c_i}, u) \),  \( V_i\) $\gets$ \(  V_i \cup  u \)	
		\STATE \(E_{used}\) $\gets$ \( E_{used} \cup (v_{c_i}, u) \)		
		\ENDIF
		\STATE \( t_i = \) Description of \( G_i \) according Definition 1.
		\STATE {\bf Add} $t_i$ in $\mathcal{T}_{s}$  with importance \( \mathcal{I}(t_i) = \text{PR}(v_{c_i}) \)
		\ENDFOR
		\ENDFOR
		\STATE \textbf{3. Reorganize Graph Description Sequence:}
		\STATE Sort \( \mathcal{T}_{s} \) by \( \mathcal{I}(t_i) \), reorganize sequence according Equation~\ref{eq:reorg}	
		\RETURN \( \mathcal{\hat{T}} \)
	\end{algorithmic}
\end{algorithm}
\setlength{\textfloatsep}{5pt}

\vspace{-4pt}
\subsection{RAG-based Micro-level Graph Understanding}
\vspace{-4pt}
For micro-level graph understanding tasks, when these tasks involve information about nodes or edges within graph description sequences that correspond to the weak memory regions of the LLMs, the LLMs are inevitably prone to forgetting this information, thereby failing to generate accurate responses.
To address this, inspired by the commonly used Retrieval-Augmented Generation (RAG) idea for enhancing the LLM reasoning capabilities, we propose establishing a lightweight, optional-scale RAG knowledge base for the nodes and edges in weak memory regions. This knowledge base, powered by RAG algorithms, retrieves relevant node and edge information for specific graph understanding tasks, thereby enhancing the comprehension of the LLMs.

Next, we introduce the construction of our framework's RAG knowledge base, termed ``{\it GraphRAG base}",  and the corresponding RAG process, termed the ``{\it GraphRAG process}". Note that existing RAG techniques \cite{DBLP:conf/aaai/GhoshAJ0CS24, DBLP:conf/icde/RorsethGGSS24, DBLP:conf/sigir/SojitraJ0JG24, DBLP:conf/wsdm/DasSVWX24} and optimizations are orthogonal to our framework and can further enhance RAG quality, but here we focus only on the most basic RAG methods.

\vspace{-6pt}
\paragraph{GraphRAG Base.}
Conventional RAG base typically require substantial storage and rely on extensive structured or unstructured data (e.g., documents, knowledge graphs). In contrast, our GraphRAG base, denoted as $\mathcal{K}$, demands minimal storage overhead. Specifically, for the graph description sequence  $\mathcal{\hat{T}}$  generated by the importance-based description reorganization, the  nodes and edges of the subgraph structures corresponding to the weak memory regions within  $\mathcal{\hat{T}}$ will be stored as the GraphRAG base.
The proportion of stored subgraph structures denoted as \( \gamma\% \), is adjustable. Since the subgraph structures corresponding to the weak memory regions have already been ranked by importance, we can conveniently select and store only the top \( \gamma\% \) of these structures. 
Moreover, to facilitate efficient RAG retrieval, we store the node and edge information of the subgraph structures that need to be included in the GraphRAG base separately: \( \mathcal{K} = \{\mathcal{K}_{node}, \mathcal{K}_{edge}\} \). Among them,
\begin{equation}
	\small
\setlength\abovedisplayskip{2pt}
\setlength\belowdisplayskip{2pt}
	\mathcal{K}_{node} = \{ (v, \deg(v)) \mid v \in V' \}
\end{equation}
where \( V' \subseteq V \) represents the set of nodes in the GraphRAG base, and \( \deg(v) \) is the degree of node \( v \).
\begin{equation}
	\small
\setlength\abovedisplayskip{1pt}
\setlength\belowdisplayskip{2pt}
	\mathcal{K}_{edge} = \{ (u, v, w(u, v)) \mid (u, v) \in E' \}
\end{equation}
where \( E' \subseteq E \) are the edges in the GraphRAG base, and \( w(u, v) \) is the edge weight between nodes \( u \) and \( v \).

\vspace{-4pt}

\paragraph{GraphRAG Process.}

\vspace{-2pt}
Based on the GraphRAG base $\mathcal{K}$, when  augmented graph information is needed for a micro-level graph understanding task $q$,  we first extract/identify all nodes from \( q \) by their names or identifiers. This can be accomplished using various entity recognition \cite{DBLP:journals/tkde/LiSHL22} or semantic parsing models \cite{lewis2020retrieval, DBLP:conf/aaai/0001WW024}:  $V_q = \text{EntityRecognition}(q)$, where $V_q$ represents the set of node entities extracted from the task $q$. Next, based on the $V_q$, we perform retrieval operations on $\mathcal{K}_{edge}$. Specifically, for $\mathcal{K}_{\text{node}}$, we directly retrieve each node and its degree information based on the nodes in $V_q$, resulting in the augmented information set $\mathcal{K}_{\text{node}}^q$:
\begin{equation}
	\small
\setlength\abovedisplayskip{2pt}
\setlength\belowdisplayskip{2pt}
	\mathcal{K}_{\text{node}}^q = \{ (v, \deg(v)) \mid v \in V_q \cap V' \}
\end{equation}
For $\mathcal{K}_{\text{edge}}$, we retrieve all edges information associated with the nodes in $V_q$, resulting in the augmented information set:
\begin{equation}
\setlength\abovedisplayskip{2pt}
\setlength\belowdisplayskip{1pt}
	\small
	\mathcal{K}_{\text{edge}}^q = \{ (u, v, w(u, v)) \mid u \in V_q \text{ or } v \in V_q, (u, v) \in E' \}
\end{equation}

Finally, the two parts of the augmented information are organized into a prompt, which is then input into LLMs to assist in the reasoning process for the task $q$:
\begin{equation}
\setlength\abovedisplayskip{1pt}
\setlength\belowdisplayskip{1pt}
		\small
	\text{Response}_{q} = \text{LLMs}\left(\text{Prompt}(\mathcal{K}_{\text{node}}^q, \mathcal{K}_{\text{edge}}^q, q)\right)
\end{equation}

\vspace{-2pt}
\paragraph{Composite  Tasks.}
\vspace{-2pt}
For composite tasks requiring multi-step reasoning, the common approach in LLMs is to design agents \cite{zhao2024expel,kirk2024improving} that use multi-step prompts to aid reasoning. For example, identifying a node's third-order neighbors involves first finding its direct (first-order) neighbors, then their neighbors (second-order), and finally the neighbors of those (third-order). This task essentially consists of three micro-level graph understanding tasks.

GraphInsight framework incorporates two key techniques that can be seamlessly integrated into the agent-based processes of LLMs to enhance the performance of such multi-step reasoning tasks:
	\vspace{-13.5pt}
\squishlist 
	\item Initially, during the LLMs agent process's inception phase, our framework's importance-based description reorganization method can be applied to the sequence of graph descriptions input into the LLMs. This enhances the LLMs' overall comprehension of the graph structure. 
	\item Subsequently, in the multi-step reasoning phase of the LLMs agent process, our framework's GraphRAG method can provide LLMs with enriched information relevant to each step of the reasoning process, thereby improving the quality of the reasoning.
\squishend
	\vspace{-3pt}

\setlength{\textfloatsep}{0pt} %
\begin{table*}[thb]\centering
	\caption{Analysis on Macro- and Micro-level Tasks with Different Baseline Methods and Models.}
	\vspace{-3.8mm}
	\label{tab:model_performance_transposed}
	\small
	\renewcommand{\arraystretch}{0.5} 
	\resizebox{\textwidth}{!}{
		\small 
		\begin{tabular}{c|c|c|c|ccc|ccc|cc|c}
			\toprule
			Models & Tasks & Raw Seq. & GraphToken & \multicolumn{3}{c|}{Prompting Methods} & \multicolumn{3}{c|}{Reordering Methods} & \multicolumn{2}{c|}{Structural Methods} & GraphInsight \\
			& & & & COT & FS & BAG & BFS & DFS & SP & AL & AM & \\
			\midrule
			\multirow{3}{*}{Mistral-7B} & Overall & 0.5476 & 0.2644 & 0.5315 & 0.4671 & 0.5145 & \underline{0.5481} & 0.5482 & {{0.5486}} & 0.5320 & 0.1922 & {{\bf0.6811}} \\
			& Macro  & 0.5222 & 0.3342 & 0.4323 & {\underline{0.5439}} & 0.4249 & 0.5212 & 0.5205 & 0.5222 & 0.5001 & 0.1808 & {\bf0.5846} \\
			& Micro   & 0.5635 & 0.2208 & {\underline{0.5935}} & 0.4192 & 0.5705 & 0.5650 & 0.5655 & 0.5651 & 0.5519 & 0.1994 & {\bf0.7425} \\
			\midrule
			\multirow{3}{*}{Llama-3-8B} & Overall & 0.4513 & 0.3480 & 0.4504 & 0.4173 & 0.4358 & 0.4551 & 0.4591 & \underline{0.4559} & 0.4048 & 0.1434 & {{\bf0.6765}} \\
			& Macro  & 0.4379 & 0.3989 & 0.4187 & 0.3702 & 0.4060 & 0.4381 & 0.4376 & \underline{0.4404} & 0.4336 & 0.1198 & {\bf0.5422} \\
			& Micro  & 0.4597 & 0.3161 & \underline{0.4702} & 0.4468 & 0.4545 & 0.4658 & {{0.4726}} & 0.4656 & 0.3869 & 0.1581 & {\bf0.7605} \\
			\midrule
			\multirow{3}{*}{Qwen2-7B} & Overall & 0.5640 & 0.3550 & {{0.5663}} & 0.5400 & {\underline{0.5677}} & 0.5381 & 0.5290 & 0.5292 & 0.5595 & 0.2129 & {\bf0.7695} \\
			& Macro  & 0.5644 & 0.4493 & 0.5423 & 0.5366 & {\underline{0.6447}} & 0.6017 & 0.5913 & 0.5958 & 0.5223 & 0.1915 & {\bf0.6587} \\
			& Micro   & 0.5637 & 0.2960 & {{0.5813}} & 0.5421& 0.5196 & 0.4984 & 0.4901 & 0.4876 & {\underline{0.5828}} & 0.2262 & {\bf0.8387} \\
			\midrule
			\multirow{3}{*}{Llama-3-8B-262k} & Overall & {\underline{0.7183}} & 0.2703 & 0.5279 & 0.5825 & 0.6231 & 0.7150 & 0.7166 & 0.7147 & 0.6603 & 0.5038 & {\textbf{0.8285}} \\
			& Macro  & {{0.6218}} & 0.4198 & 0.5751 & 0.4652 & 0.5115 & 0.6189 & \underline{0.6221} & 0.6183 & 0.5292 & 0.3529 & {\bf0.6928} \\
			& Micro   & {\underline{0.7786}} & 0.1770 & 0.4985 & 0.6559 & 0.6928 & {{0.7752}} & 0.7757 & 0.7750 & 0.7422 & 0.5982 & {\bf0.9401} \\
			\midrule
			\multirow{3}{*}{Vicuna-7B} & Overall & {{0.1273}} & \underline{0.1794} & 0.1147 & 0.1041 & 0.1066 & 0.1276 & {{0.1277}} & {0.1277} & 0.0862 & 0.0345 & {\bf0.3557} \\
			& Macro  & 0.1267 & {\underline{0.1575}} & 0.1164 & 0.0877 & 0.0995 & 0.1266 & 0.1259 & 0.1263 & 0.0567 & 0.0409 & {\bf0.2296} \\
			& Micro   & {{0.1276}} & \underline{0.2016} & 0.1137 & 0.1144 & 0.1111 & {{0.1282}} & 0.1289 & 0.1285 & 0.1046 & 0.0306 & {\bf 0.4346} \\
			\bottomrule                        
			
		\end{tabular}
	}
	
	\vspace{-10px}
\end{table*}

\vspace{-6pt}
\section{Evaluation}
\label{Experiment}
\vspace{-6pt}
We conduct experiments with GraphInsight on two level graph understanding tasks: (i) macro understanding, (ii) micro understanding. Section~\ref{exp:setup} summarizes the experimental setup. Section~\ref{exp:performance} demonstrates the advantages of GraphInsight in enhancing LLMs' understanding of graphs. Ablation study and hyperparameter analysis are on Sections \ref{exp:as} and \ref{exp:hp}. Each reported value is the average of three runs.

\vspace{-7pt}
\subsection{Experimental Setup}
\label{exp:setup}
\vspace{-1pt}
	\vspace{-3pt}
\subsubsection{Baselines.}
	\vspace{-2.5pt}

We compare GraphInsight with 10 baselines:  
(a) raw sequence input method without any processing (Raw Seq.),  
(b) GNN encoding-based method\footnote{GNN encoding-based methods like GraphToken rely on implicit graph embeddings, which are orthogonal to GraphInsight's explicit graph description approach. Despite this, we include them to ensure a comprehensive comparison.} (GraphToken) \cite{perozzi2024let},  
(c) prompting methods including Build-a-Graph Prompting (BAG) \cite{wang2024can}, Chain-of-Thought (COT) \cite{wei2022chain}, and Few-Shot (FS) \cite{brown2020language},  
(d) reordering methods \cite{ge2024sequential} based on breadth-first-search order (BFS), depth-first-search order (DFS), and shortest-path order (SP),  
(e) structural methods including adjacency lists (AL) and adjacency matrices (AM).

\setlength{\textfloatsep}{0pt} %
\begin{figure}[t]
	\centering
	\includegraphics[width=0.234\textwidth]{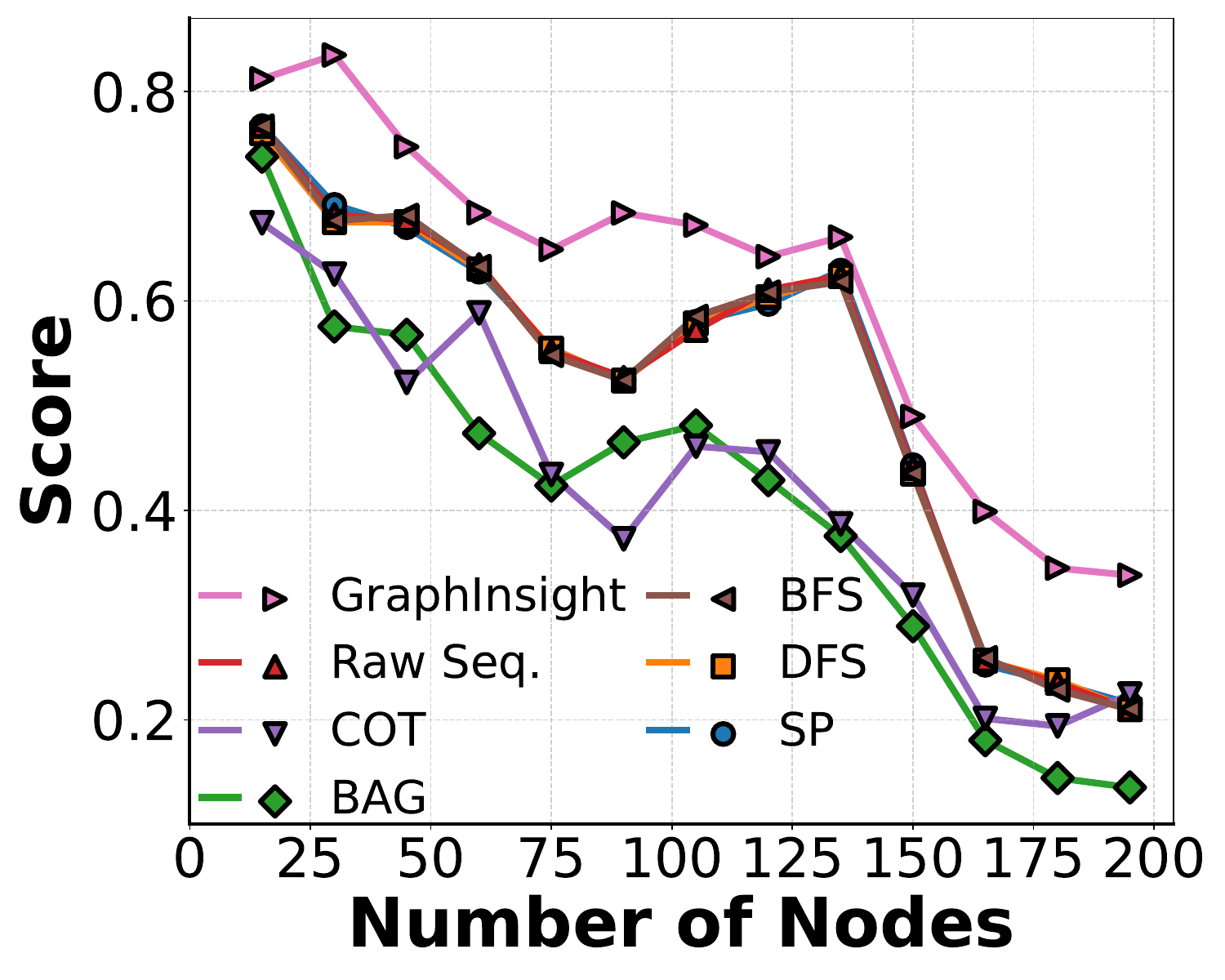}
	\includegraphics[width=0.234\textwidth]{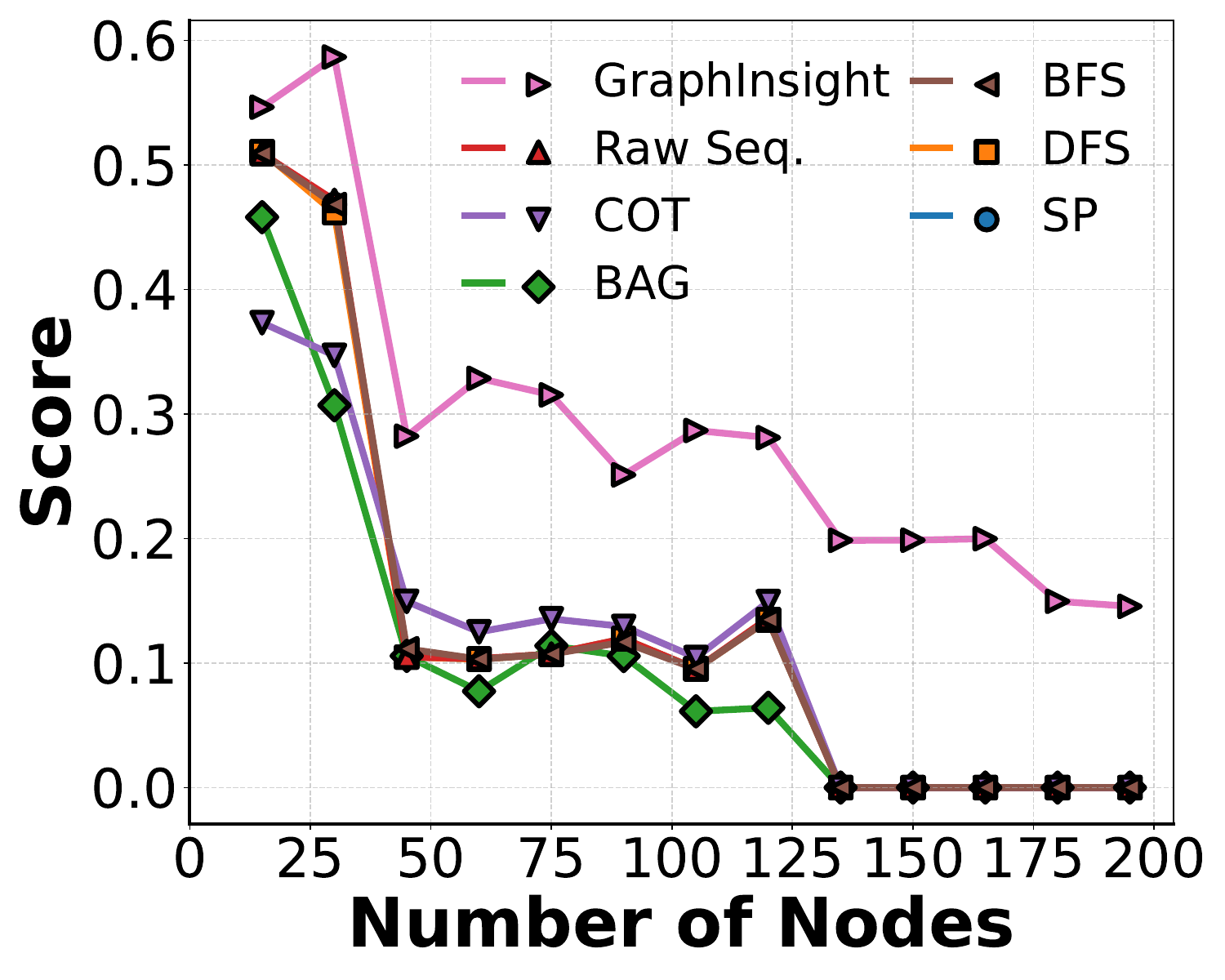}
	\raisebox{7.5mm}{\makebox[0.225\textwidth]{\small (a) Macro-level (Mistral-7B)}}
	\raisebox{7.5mm}{\makebox[0.225\textwidth]{\small (b) Macro-level (Vicuna-7B)}}
	
	\vspace{-6.5mm}
	
	\includegraphics[width=0.234\textwidth]{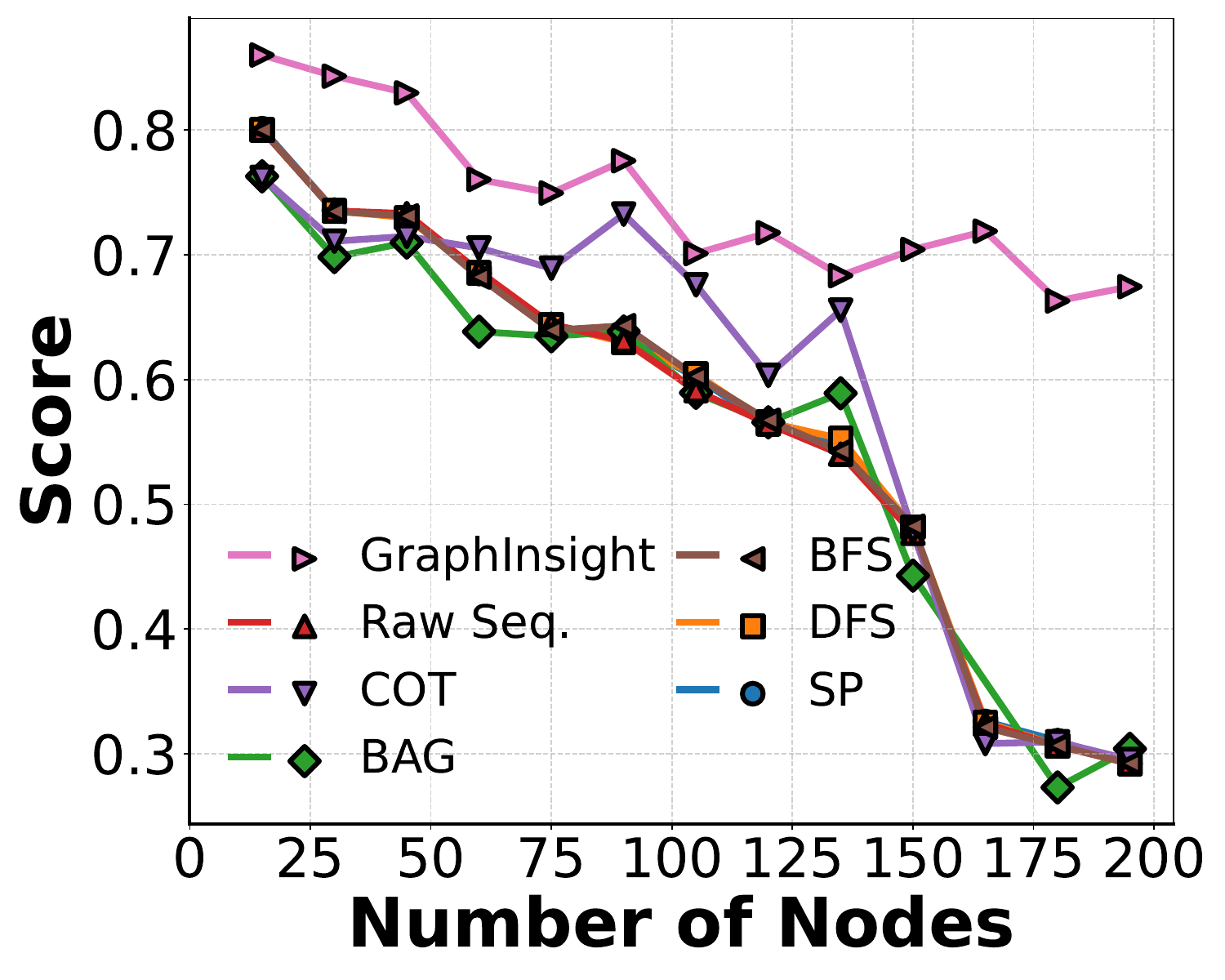}
	\includegraphics[width=0.234\textwidth]{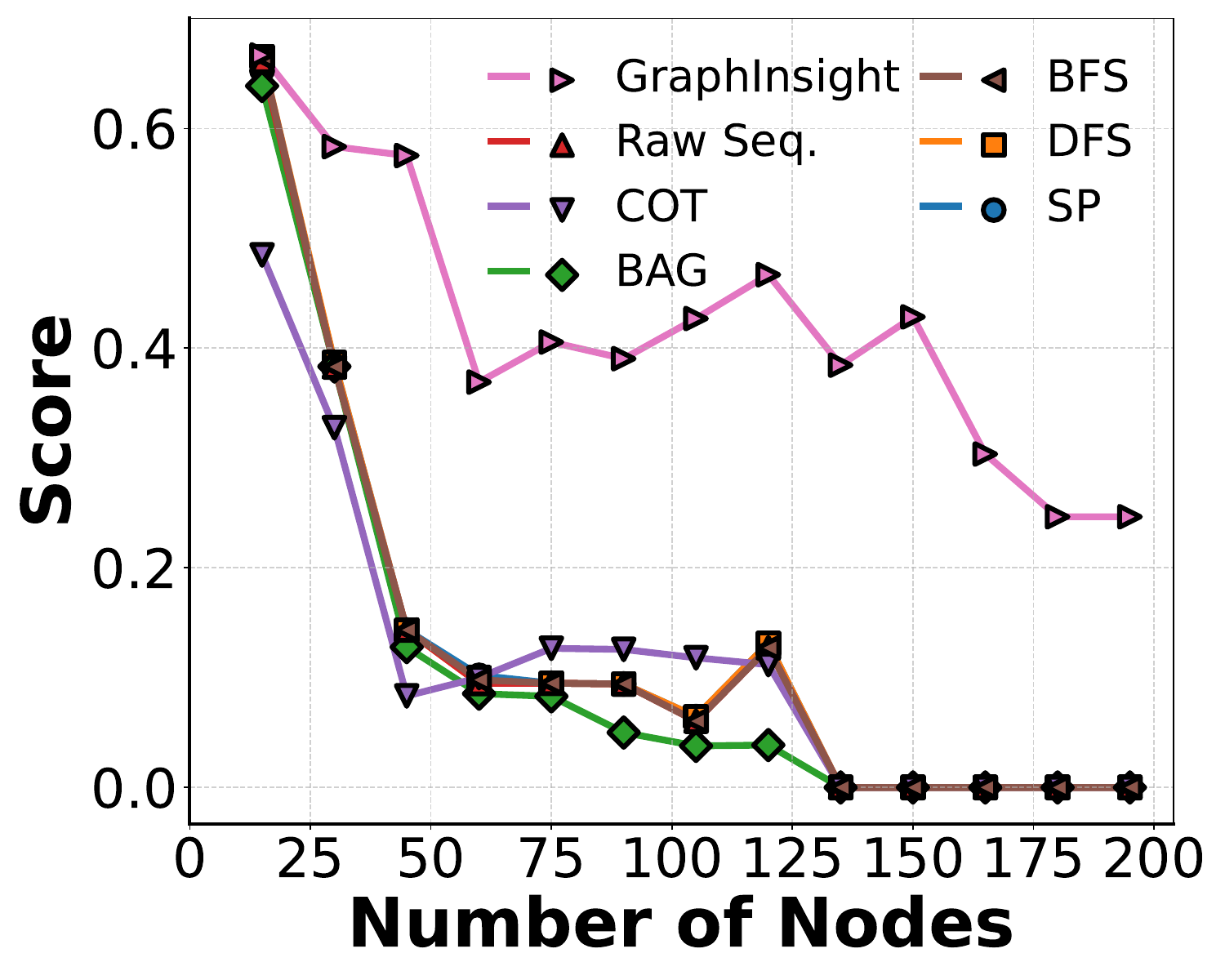}
	\raisebox{7.5mm}{\makebox[0.225\textwidth]{\small (c) Micro-level (Mistral-7B)}}
	\raisebox{7.5mm}{\makebox[0.225\textwidth]{\small (d) Micro-level (Vicuna-7B)}}
	\vspace{-8mm}
	\caption{Analysis on Graphs with Different $|V|$ }
	\label{fig:analysisonV}
\end{figure}

\vspace{-10pt}

\subsubsection{Benchmarks and Evaluation Tasks.} 

\vspace{-6pt}
Existing graph understanding benchmarks suffer from limited node coverage, unclear task definitions, a lack of structural diversity, and insufficient support for multi-step graph reasoning. To address these issues, we introduce GraphSQA, a comprehensive benchmark designed to evaluate the capabilities of LLMs in understanding graph structures, which encompasses a wide range of task types found in existing benchmarks \cite{wang2024can,yuan2024gracore}. GraphSQA includes a broad spectrum of node counts and features diverse graph structures, such as multi-edges and self-loops. It encompasses two types of graph understanding tasks: (1) 5 macro-level graph tasks and (2) 15 micro graph tasks including 7 composite tasks. Further details are provided in \textbf{Appendix}~\ref{Benchmark}. In our experiments, we evaluate the average performance of various methods and models on diverse graph tasks with $|V|$ ranging from 15 to 200 nodes, covering 10,400 tasks in total—2,600 macro-level and 7,800 micro-level.

\vspace{-6pt}
\subsubsection{Metric.}
\vspace{-3.5pt}
GraphSQA employs three answer types: boolean, numerical, and set. 
For boolean answers, the score is 1 if correct, otherwise 0. 
For numerical answers, we use one minus the relative error \cite{wang2024can}. 
For set answers, the score is the Jaccard similarity \cite{ji2013min} between the answer and the ground truth.

\vspace{-4pt}

\subsubsection{Models.}
\vspace{-3pt}
We employed a diverse selection of open-source LLMs, encompassing both long-sequence models, such as Mistral-7B and Llama3-8B-256K (fine-tuned for long sequences), and their counterparts, including Llama3-8B, Qwen2-7B, and Vicuna-7B.

\vspace{-5pt}

\subsection{Performance}
\label{exp:performance}
\vspace{-1pt}
\vspace{-2.5pt}
\subsubsection{Macro-level Tasks.}
\vspace{-3.5pt}

GraphInsight outperforms all other methods across all LLMs for macro-tasks, as shown in Table~\ref{tab:model_performance_transposed}. Specifically, GraphInsight can achieve up to a 4.61$\times$ increase in score compared to AM on Vicuna-7B. Structural methods show the smallest improvement for macro-level tasks. The other two types of methods, i.e., prompting and structural methods, result in only minimal improvements in macro-level tasks. For example, on Llama-3-8B, they can achieve at most a minimal improvement of 5.7\%, from 0.4379 to 0.4404. However, GraphInsight can provide a substantial increase of 23.82\%, from 0.4379 to 0.5422. Also, as shown in Figures~\ref{fig:analysisonV}(a)-(b), across macro-level tasks with varying $|V|$, GraphInsight consistently outperforms other methods. As $|V|$ increases, the understanding capability tends to decrease, but GraphInsight shows the smallest decline. 

\vspace{-5pt}
\subsubsection{Micro-level Tasks.} 
\vspace{-4pt}
For the micro-level tasks, GraphInsight still outperforms the others, as shown in Table~\ref{tab:model_performance_transposed}. Notably, GraphInsight can achieve up to an 14.02$\times$ increase in score compared to AL on  Vicuna-7B. Similar to macro-level tasks, baseline methods offer minimal improvement in the large model's understanding. For example, on Qwen2-7B, they can achieve at most a minimal improvement of 3.4\%, from 0.5637 to 0.5828. However, GraphInsight can provide a substantial increase of 48.78\%, from 0.5637 to 0.8387. In Figures~\ref{fig:analysisonV}(c)-(d), as $|V|$ increases, GraphInsight causes the least decline in the model's understanding. For example, on Mistral-7B, when $|V|$ increases from 135 to 195, the score decreases only slightly from 0.68 to 0.67. In contrast, the best baseline's score drops more significantly, from 0.54 to 0.29. We also conduct experiments on composite tasks, e.g., common neighbor finding (CN) and $k$-order neighbor finding ($k$-ON) shown in Figure~\ref{fig:fig_s_vs_l}. GraphInsight can provide an improvement of 12.75$\times$ and 3.25$\times$ on Vicuna, respectively.

\setlength{\textfloatsep}{3pt} %
\begin{figure}[t]
	\centering
	\includegraphics[width=0.234\textwidth]{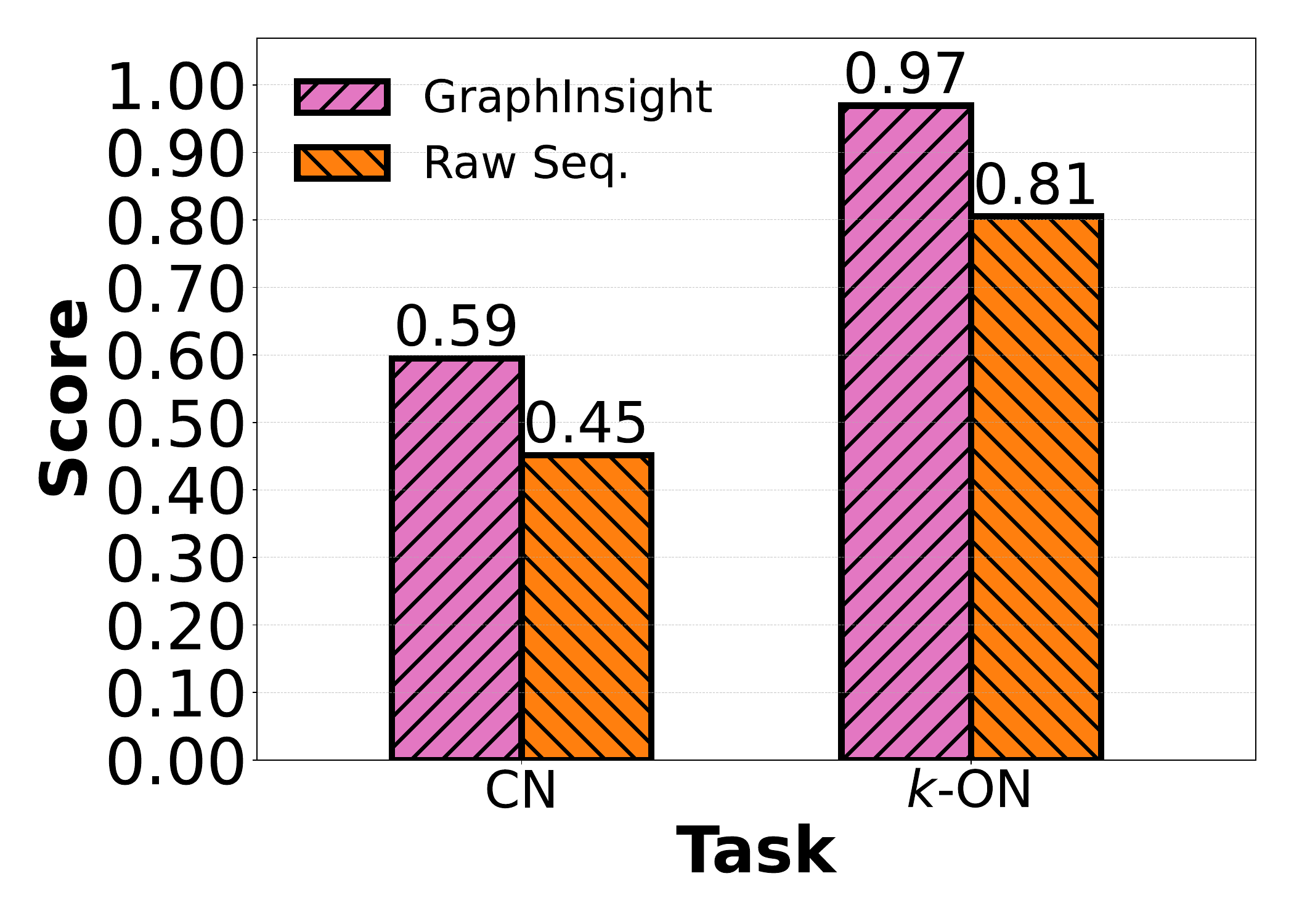}
	\includegraphics[width=0.234\textwidth]{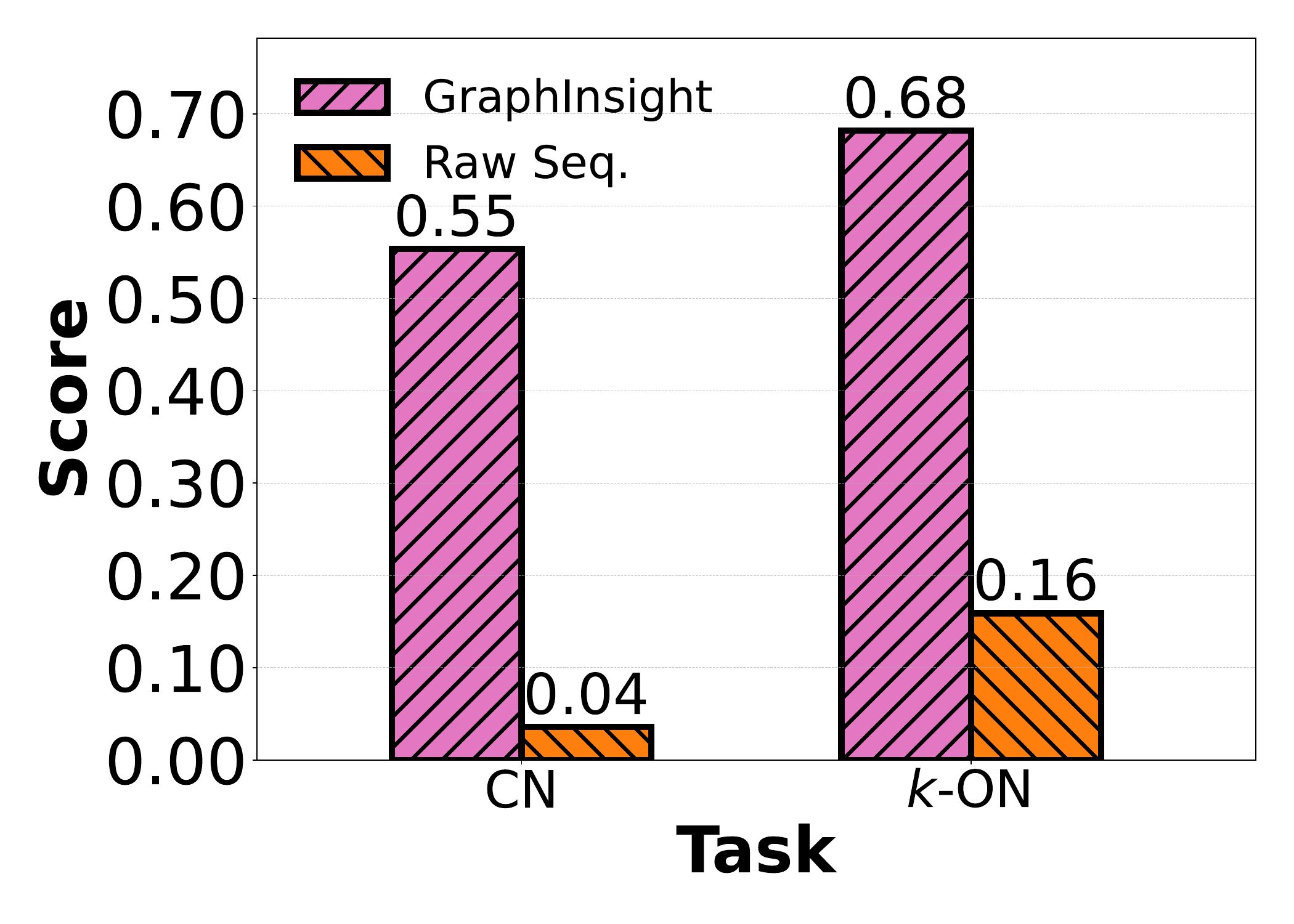}
	\
	\raisebox{9.5mm}{\makebox[0.225\textwidth]{\small (a) Qwen2-7B}}
	\raisebox{9.5mm}{\makebox[0.225\textwidth]{\small (b) Vicuna-7B}}
	\vspace{-11mm}
	\caption{Analysis on Composite Tasks}
	\label{fig:fig_s_vs_l}
\end{figure}

\setlength{\textfloatsep}{3pt} %
\begin{figure}[t]
	\centering
	
	\includegraphics[width=0.225\textwidth]{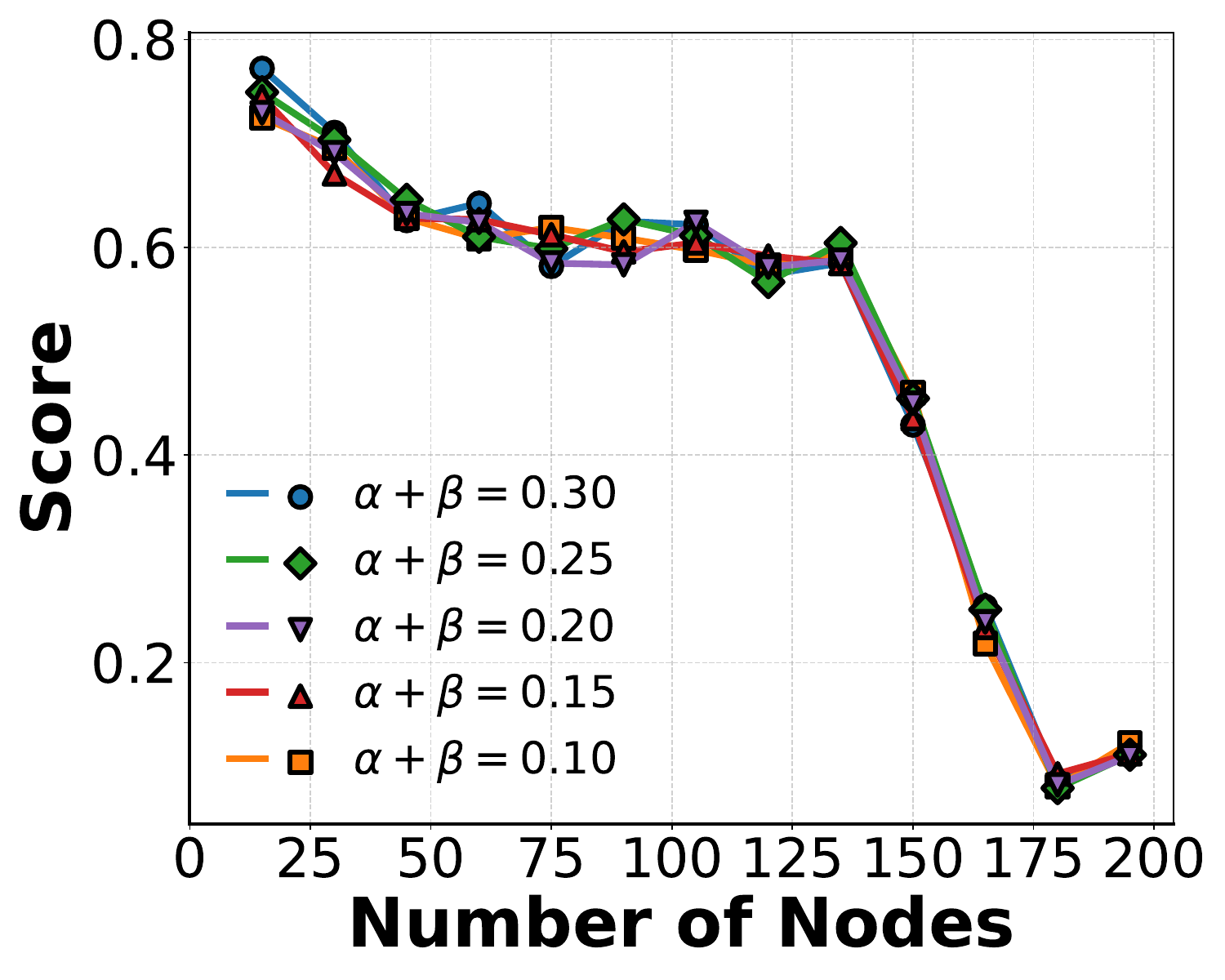}
	\includegraphics[width=0.225\textwidth]{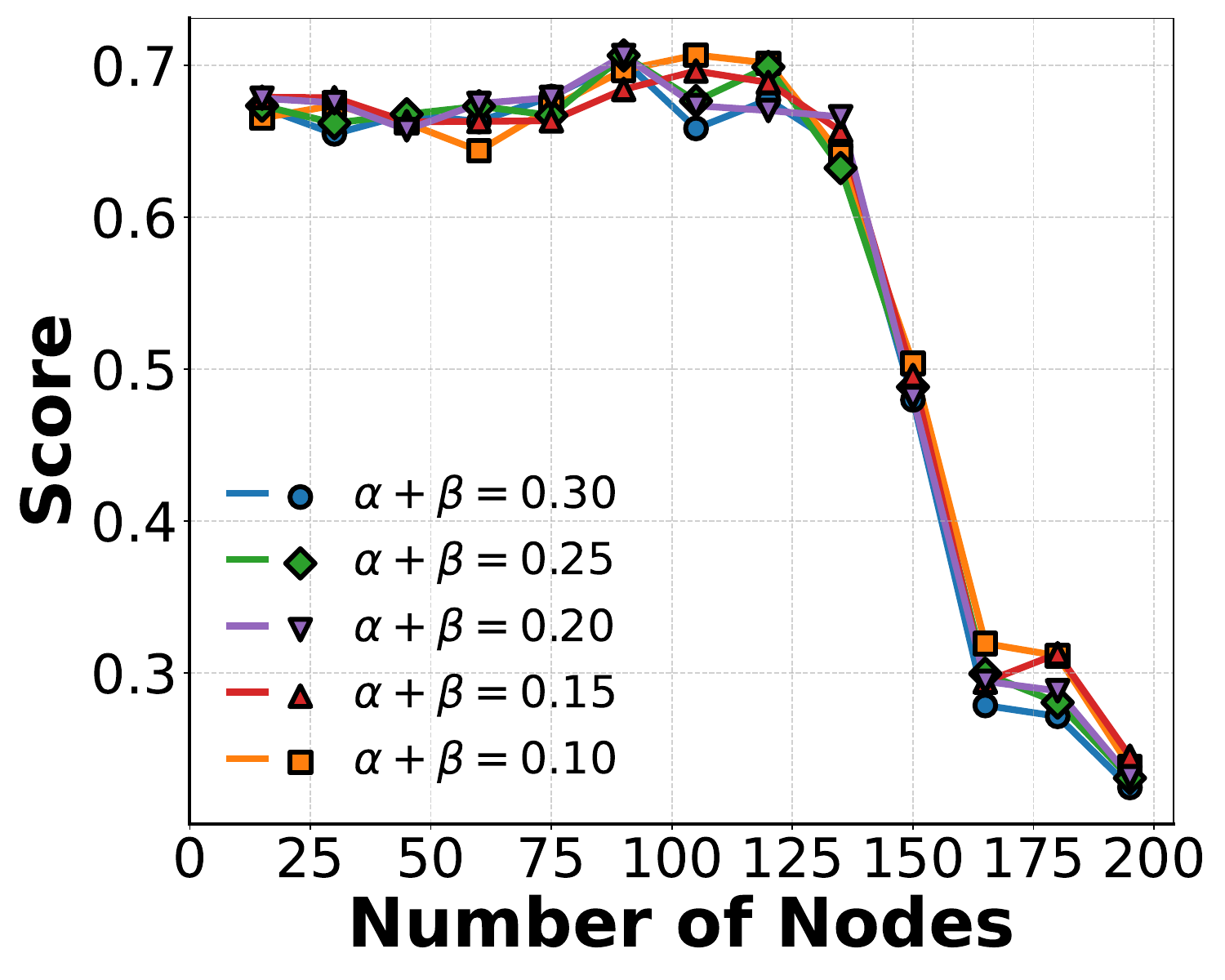}
	\
	\vspace{-6.5mm}
	\raisebox{7.5mm}{\makebox[0.225\textwidth]{\small (a) Mistral-7B ($\frac{\alpha}{\beta} = \frac{3}{7}$)}}
	\raisebox{7.5mm}{\makebox[0.225\textwidth]{\small (b) Qwen2-7B ($\frac{\alpha}{\beta} = \frac{3}{7}$)}}
	
	\includegraphics[width=0.225\textwidth]{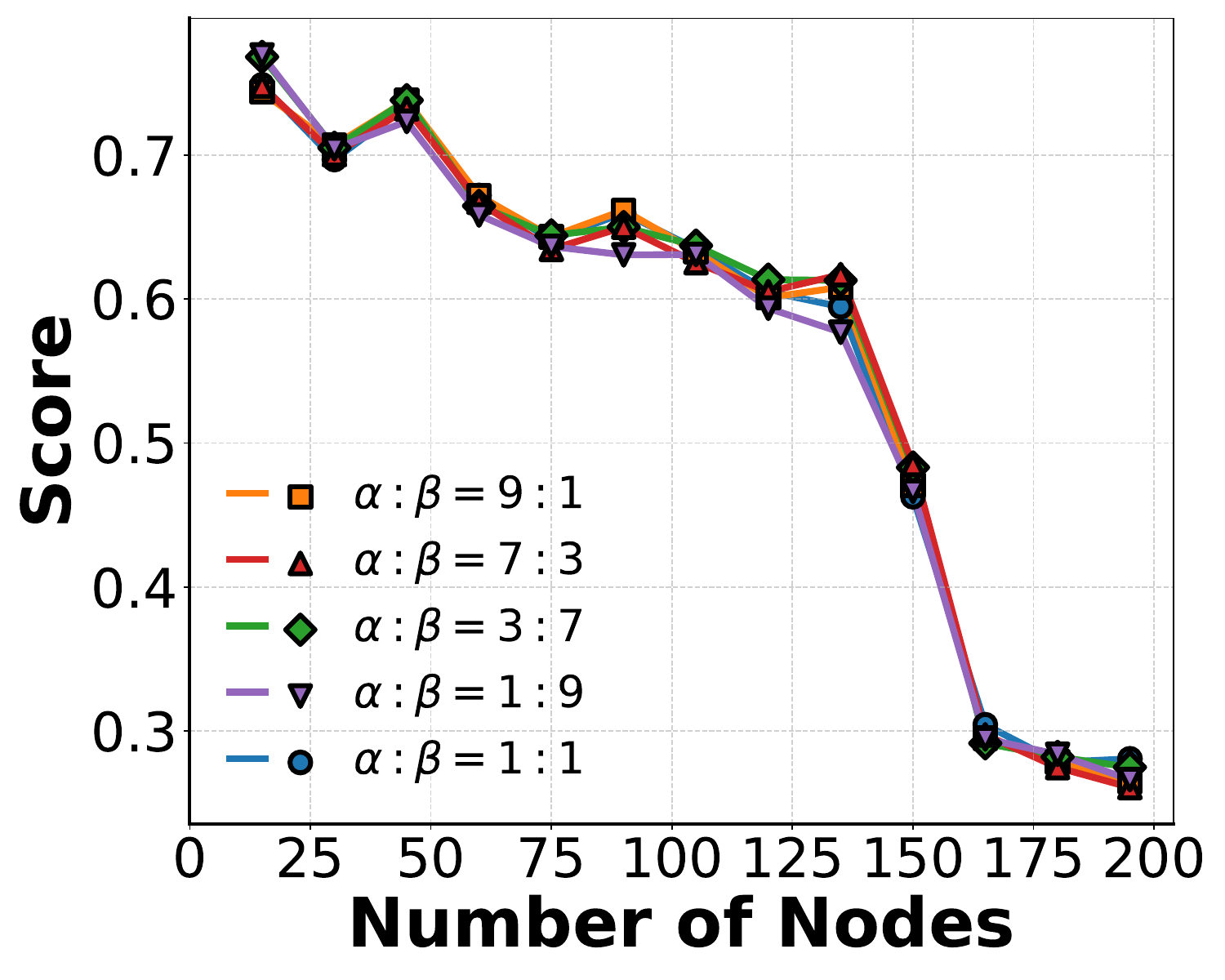}
	\includegraphics[width=0.225\textwidth]{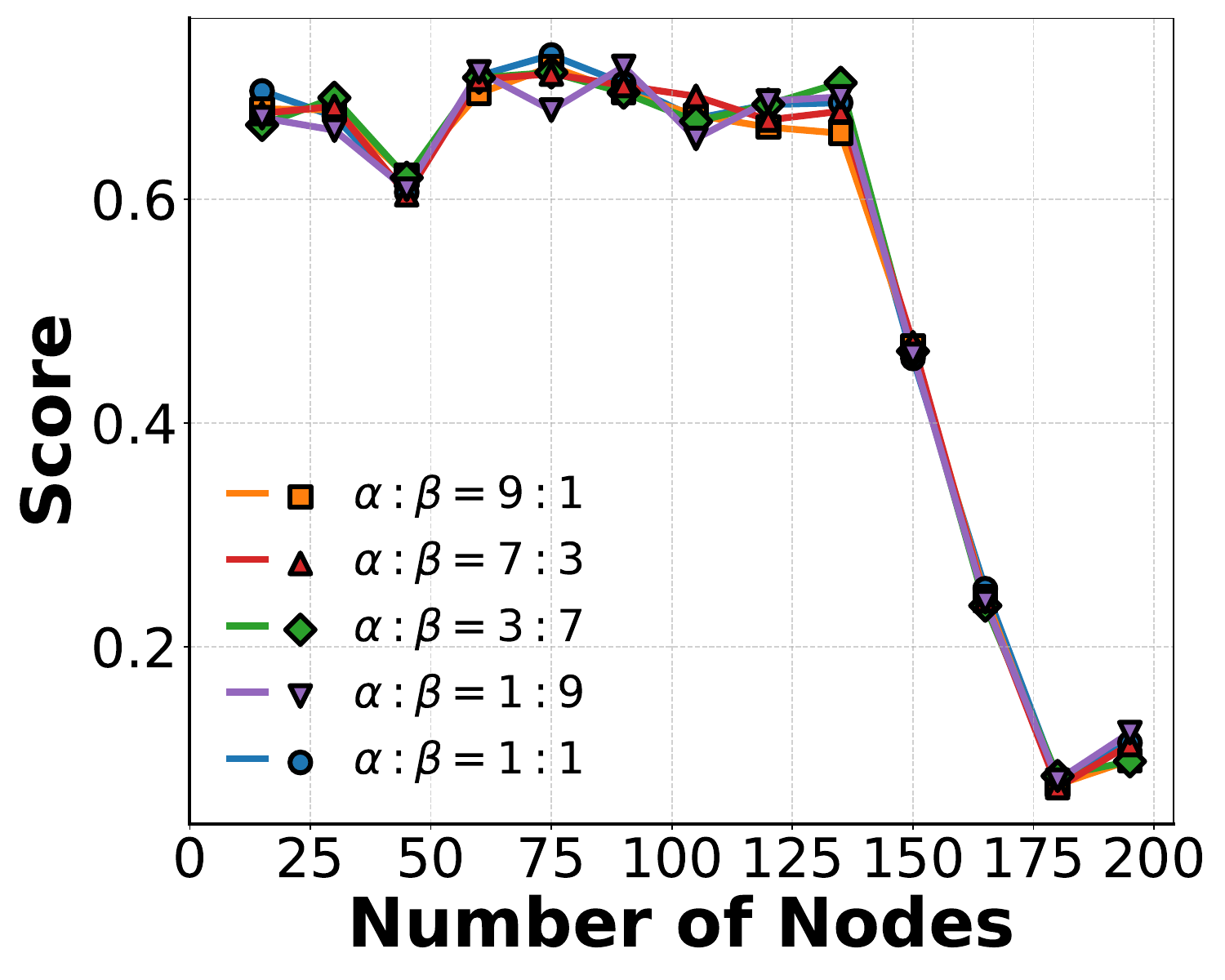}
	\
	\raisebox{7.5mm}{\makebox[0.225\textwidth]{\small (c) Mistral-7B {($\alpha+\beta=15$)}}}
	\raisebox{7.5mm}{\makebox[0.225\textwidth]{\small (d) Qwen2-7B {($\alpha+\beta=15$)}}}
	\vspace{-10mm}
	\caption{Head and Tail Strong Region Thresholds $\alpha$ and $\beta$.}
	\label{fig:split}
\end{figure}

\setlength{\textfloatsep}{3pt} %
\begin{figure}[t]
	\centering
	\includegraphics[width=0.225\textwidth]{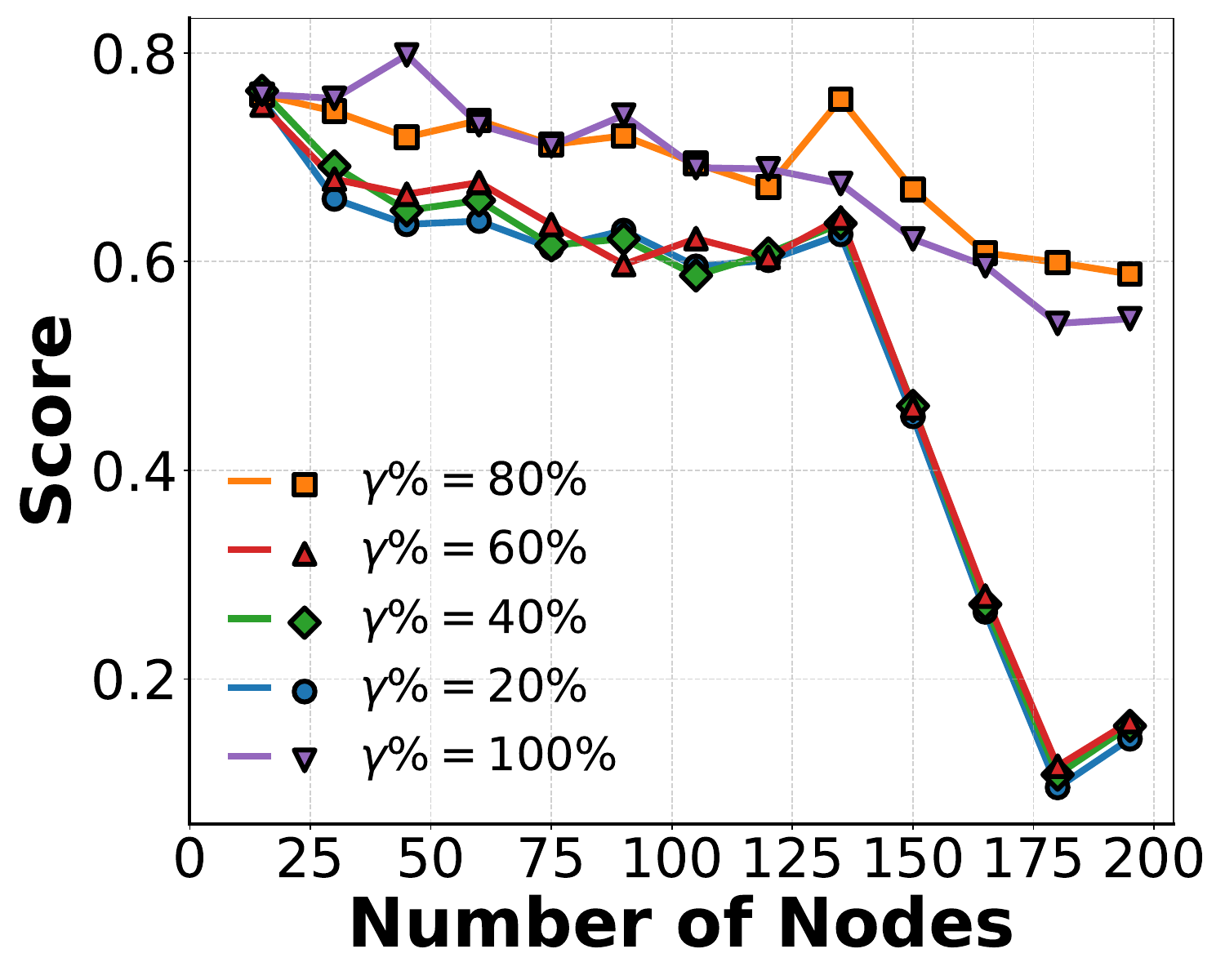}
	\includegraphics[width=0.225\textwidth]{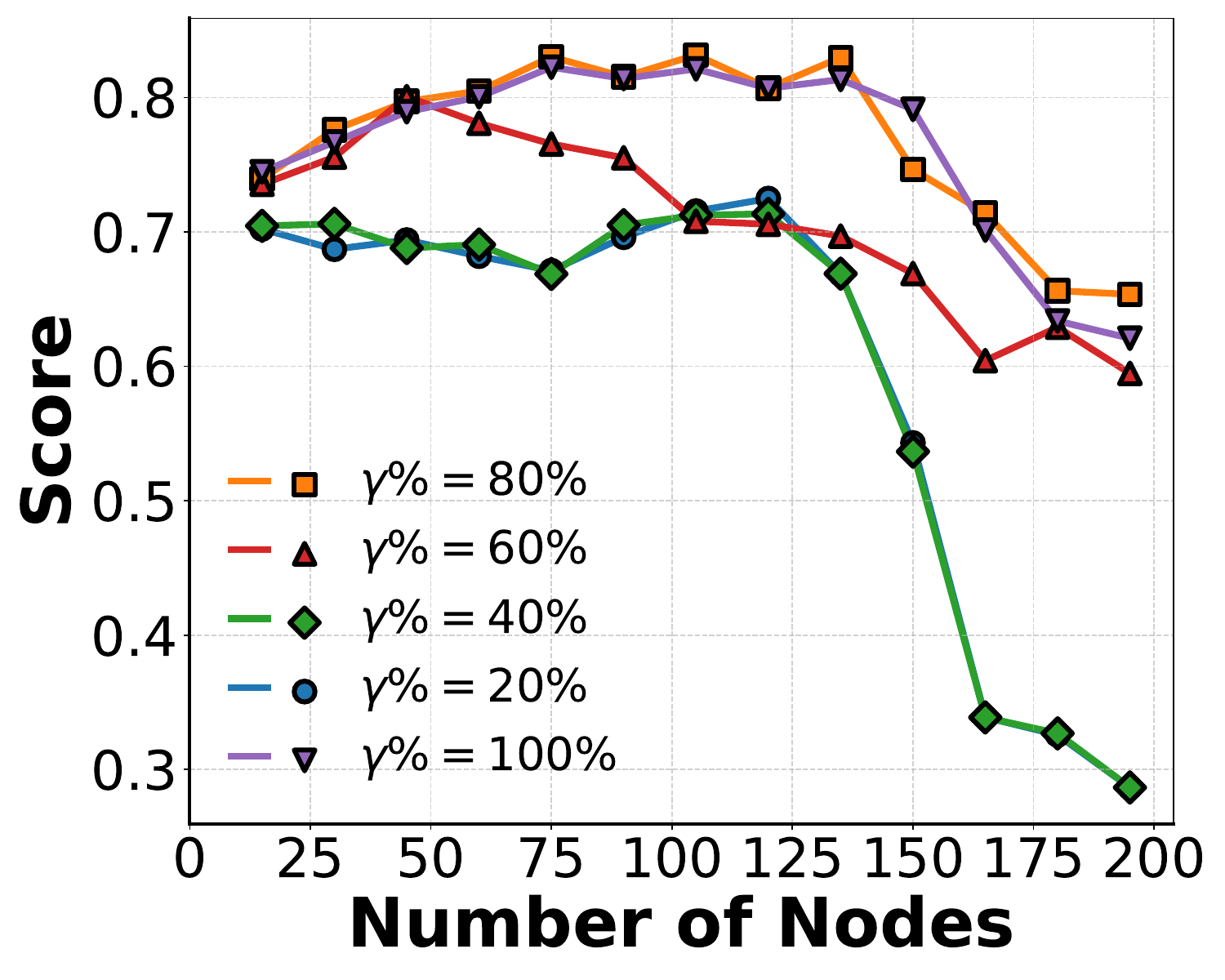}
	\raisebox{7.5mm}{\makebox[0.225\textwidth]{\small (a) Mistral-7B}}
	\raisebox{7.5mm}{\makebox[0.225\textwidth]{\small (b) Qwen2-7B}}
	\vspace{-10mm}
	\caption{GraphRAG Rate $\gamma\%$.}
	\label{fig:fig15}
\end{figure}

\setlength{\textfloatsep}{3pt}
\setlength{\floatsep}{3pt}
\setlength{\intextsep}{3pt}
\begin{table}[t]\centering
	\caption{Ablation Study.}
		\vspace{-4mm}
	\label{tab:ablation}
	\small
	\renewcommand{\arraystretch}{0.02} 
	\resizebox{\columnwidth}{!}{
		\begin{tabular}{c|c|cc}
			\toprule
			Methods & Models & Macro & Micro \\
			\midrule
			Raw Seq. & \multirow{3}{*}{Mistral-7B} & 0.5222 & 0.5635 \\
			w/ Reorganization & & {\bf0.5851} & 0.5639 \\
			w/ Reorganization and GraphRAG & & - & {\bf0.7425} \\
			\midrule
			Raw Seq. & \multirow{3}{*}{Llama-3-8B} & 0.4379 & 0.4597 \\
			w/ Reorganization & & {\bf0.5029} & 0.5298 \\
			w/ Reorganization and GraphRAG & & - & {\bf0.7605} \\
			\midrule
			Raw Seq. & \multirow{3}{*}{Qwen2-7B} & 0.5644 & 0.5637 \\
			w/ Reorganization & & {\bf0.6293} & 0.5676 \\
			w/ Reorganization and GraphRAG & & - & {\bf0.8387} \\
			\midrule
			Raw Seq. & \multirow{3}{*}{Llama-3-8B-262K} & 0.6218 & 0.7786 \\
			w/ Reorganization & & {\bf0.6928} & 0.8477 \\
			w/ Reorganization and GraphRAG & & - & {\bf0.9401} \\
			\midrule
			Raw Seq. & \multirow{3}{*}{Vicuna-7B} & 0.1267 & 0.1276 \\
			w/ Reorganization & & {\bf0.1571} & 0.1588 \\
			w/ Reorganization and GraphRAG & & - & {\bf0.4346} \\
			\bottomrule
		\end{tabular}
	}
\end{table}

\vspace{-10pt}

\subsection{Ablation Study}
\label{exp:as}
\vspace{-4pt}
We conduct the ablation study on graph description organization and GraphRAG to verify the effectiveness of them. As shown in Table~\ref{tab:ablation}, graph description organization is the optimization primarily for macro-level tasks as well as  for micro-level tasks. 
For example,  on Llama-3-8B-262K, reorganization can achieve an 11.4\% increase  for macro-level tasks and  an 8.9\% increase for micro-level tasks.
On the other hand, GraphRAG focuses on micro-level tasks and can provide further notably improvements for micro-level tasks on top of the gains achieved by reorganization.

\vspace{-5pt}

\subsection{Hyperparameter Analysis}
\label{exp:hp}
\vspace{-4pt}

\subsubsection{Strong Region Thresholds $\alpha\%$ and $\beta\%$.}
\label{sec:srt} 
\vspace{-1.5pt}
We conduct experiments with $\alpha\% + \beta\% = 15\%$ (resp. $\frac{\alpha\%}{\beta\%} = \frac{3}{7}$) to assess how variations in $\frac{\alpha\%}{\beta\%}$ (resp. $\alpha\% + \beta\%$) affect LLMs' graph understanding. As shown in Figure~\ref{fig:split}, the performance remains stable in both cases, indicating insensitivity to these parameters. We recommend setting $\frac{\alpha\%}{\beta\%} = \frac{3}{7}$ and $\alpha\% + \beta\% = 15\%$ for simplicity.

\vspace{-3pt}

\subsubsection{GraphRAG Base Rate $\gamma\%$.}
\vspace{-2.5pt}
We examine the impact of $\gamma\%$ on the size of the GraphRAG base. As $\gamma\%$ increases, the base retains more critical graph information, improving micro-level understanding. However, as shown in Figure~\ref{fig:split}, the improvement diminishes when $\gamma\%$ exceeds 80\%. We recommend setting $\gamma\%$ to 80\%.

\vspace{-6pt}
\section{Conclusion}
\vspace{-4pt}
In this paper, we introduce GraphInsight, the first framework to address the challenge of graph structure comprehension in LLMs through their inherent ``positional bias.'' By leveraging strong memory in certain graph sequence regions and compensating for weaker ones, GraphInsight enhances graph understanding. It employs two key techniques: importance-based reorganization and lightweight RAG, optimized for macro- and micro-level tasks. Experiments show that GraphInsight outperforms all baselines across graph sizes. Future work will focus on improving LLMs' understanding of more complex graph types, including those with labels and semantics.

\newpage
\section*{Limitations}

The proposed \textbf{GraphInsight} framework significantly enhances LLMs' ability to comprehend graph structures, yet it presents several limitations that could inspire future research directions. 

First, the quantification of subgraph importance remains a task-dependent challenge, as different graph characteristics and task objectives require distinct importance measures. Future work could explore more task-aware strategies for subgraph importance quantification to improve the framework's adaptability across various graph-related tasks. Additionally, while the GraphRAG component addresses memory limitations effectively, its optimization in terms of retrieval efficiency and storage overhead remains an open area for improvement. Future research could focus on refining these aspects to enhance retrieval speed and reduce memory consumption. Moreover, the framework's approach contrasts with other orthogonal LLM-based methods, such as instruction fine-tuning or implicit embedding techniques. Integrating these methods with GraphInsight could complement its capabilities and foster further advancements in graph structure comprehension tasks. 

\bibliography{custom}

\begin{thebibliography}{53}
\providecommand{\natexlab}[1]{#1}

\bibitem[{An et~al.(2024)An, Ma, Lin, Zheng, and Lou}]{an2024make}
Shengnan An, Zexiong Ma, Zeqi Lin, Nanning Zheng, and Jian-Guang Lou. 2024.
\newblock Make your llm fully utilize the context.
\newblock \emph{arXiv preprint arXiv:2404.16811}.

\bibitem[{Babic(2023)}]{DBLP:conf/ekgllm/Babic23}
Bojan Babic. 2023.
\newblock Llms for social networks: Applications, challenges and solutions.
\newblock In \emph{CIKM Workshop}, volume 3532.

\bibitem[{Baek et~al.(2023)}]{Baek2023}
Jinheon Baek et~al. 2023.
\newblock Knowledge-augmented language model prompting for zero-shot knowledge
  graph question answering.
\newblock In \emph{ACL Workshop}.

\bibitem[{Besta et~al.(2024)Besta, Blach, Kubicek, Gerstenberger, Podstawski,
  Gianinazzi, Gajda, Lehmann, Niewiadomski, Nyczyk, and
  Hoefler}]{DBLP:conf/aaai/BestaBKGPGGLNNH24}
Maciej Besta, Nils Blach, Ales Kubicek, Robert Gerstenberger, Michal
  Podstawski, Lukas Gianinazzi, Joanna Gajda, Tomasz Lehmann, Hubert
  Niewiadomski, Piotr Nyczyk, and Torsten Hoefler. 2024.
\newblock Graph of thoughts: Solving elaborate problems with large language
  models.
\newblock In \emph{AAAI}, pages 17682--17690.

\bibitem[{Besta et~al.(2023)Besta, Gerstenberger, Peter, Fischer, Podstawski,
  Barthels, Alonso, and Hoefler}]{openproblem}
Maciej Besta, Robert Gerstenberger, Emanuel Peter, Marc Fischer, Micha\l{}
  Podstawski, Claude Barthels, Gustavo Alonso, and Torsten Hoefler. 2023.
\newblock Demystifying graph databases: Analysis and taxonomy of data
  organization, system designs, and graph queries.
\newblock \emph{ACM Computing Surveys}, 56(2).

\bibitem[{Brown et~al.(2020)Brown, Mann, Ryder, Subbiah, Kaplan, Dhariwal,
  Neelakantan, Shyam, Sastry, Askell et~al.}]{brown2020language}
Tom Brown, Benjamin Mann, Nick Ryder, Melanie Subbiah, Jared~D Kaplan, Prafulla
  Dhariwal, Arvind Neelakantan, Pranav Shyam, Girish Sastry, Amanda Askell,
  et~al. 2020.
\newblock Language models are few-shot learners.
\newblock \emph{NeurIPS}, 33:1877--1901.

\bibitem[{Chen et~al.(2024{\natexlab{a}})Chen, Li, Tang, and
  Li}]{chen2024graphwiz}
Nuo Chen, Yuhan Li, Jianheng Tang, and Jia Li. 2024{\natexlab{a}}.
\newblock Graphwiz: An instruction-following language model for graph
  computational problems.
\newblock In \emph{Proceedings of the 30th ACM SIGKDD Conference on Knowledge
  Discovery and Data Mining}, pages 353--364.

\bibitem[{Chen et~al.(2024{\natexlab{b}})Chen, Mao, Li, Jin, Wen, Wei, Wang,
  Yin, Fan, Liu et~al.}]{chen2024exploring}
Zhikai Chen, Haitao Mao, Hang Li, Wei Jin, Hongzhi Wen, Xiaochi Wei, Shuaiqiang
  Wang, Dawei Yin, Wenqi Fan, Hui Liu, et~al. 2024{\natexlab{b}}.
\newblock Exploring the potential of large language models (llms)in learning on
  graphs.
\newblock \emph{KDD}, 25(2):42--61.

\bibitem[{Das et~al.(2024)Das, Saboo, Vadrevu, Wang, and
  Xu}]{DBLP:conf/wsdm/DasSVWX24}
Sudeep Das, Raghav Saboo, Chaitanya S.~K. Vadrevu, Bruce Wang, and Steven Xu.
  2024.
\newblock Applications of llms in e-commerce search and product knowledge
  graph: The doordash case study.
\newblock In \emph{WSDM}. {ACM}.

\bibitem[{Fatemi et~al.(2023)}]{fatemi2023talk}
Fatemi et~al. 2023.
\newblock Talk like a graph: Encoding graphs for large language models.
\newblock \emph{arXiv preprint arXiv:2310.04560}.

\bibitem[{Ge et~al.(2023)Ge, Hua, Mei, Ji, Tan, Xu, Li, and
  Zhang}]{DBLP:conf/nips/GeHMJTXLZ23}
Yingqiang Ge, Wenyue Hua, Kai Mei, Jianchao Ji, Juntao Tan, Shuyuan Xu, Zelong
  Li, and Yongfeng Zhang. 2023.
\newblock Openagi: When {LLM} meets domain experts.
\newblock In \emph{NeurIPS}.

\bibitem[{Ge et~al.(2024)Ge, Liu, Mei, Chen, and Cheng}]{ge2024sequential}
Yuyao Ge, Shenghua Liu, Lingrui Mei, Lizhe Chen, and Xueqi Cheng. 2024.
\newblock Sequential ordering in textual descriptions: Impact on spatial
  perception abilities of large language models.
\newblock \emph{arXiv preprint arXiv:2402.07140}.

\bibitem[{Geng et~al.(2022)Geng, Wang, Zhuang, Ming, Du, Jiang, Guo, and
  Liu}]{DBLP:conf/cikm/GengWZMDJG022}
Hao Geng, Deqing Wang, Fuzhen Zhuang, Xuehua Ming, Chenguang Du, Ting Jiang,
  Haolong Guo, and Rui Liu. 2022.
\newblock Modeling dynamic heterogeneous graph and node importance for future
  citation prediction.
\newblock In \emph{CIKM}, pages 572--581. ACM.

\bibitem[{Ghosh et~al.(2024)Ghosh, Acharya, Jain, Saha, Chadha, and
  Sinha}]{DBLP:conf/aaai/GhoshAJ0CS24}
Akash Ghosh, Arkadeep Acharya, Raghav Jain, Sriparna Saha, Aman Chadha, and
  Setu Sinha. 2024.
\newblock Clipsyntel: {CLIP} and {LLM} synergy for multimodal question
  summarization in healthcare.
\newblock In \emph{AAAI}, pages 22031--22039.

\bibitem[{Guo et~al.(2023)Guo, Du, Liu, Zhou, He, and Han}]{guo2023gpt4graph}
Jiayan Guo, Lun Du, Hengyu Liu, Mengyu Zhou, Xinyi He, and Shi Han. 2023.
\newblock Gpt4graph: Can large language models understand graph structured
  data? an empirical evaluation and benchmarking.
\newblock \emph{arXiv preprint arXiv:2305.15066}.

\bibitem[{Hegselmann et~al.(2023)Hegselmann, Buendia, Lang, Agrawal, Jiang, and
  Sontag}]{hegselmann2023tabllm}
Stefan Hegselmann, Alejandro Buendia, Hunter Lang, Monica Agrawal, Xiaoyi
  Jiang, and David Sontag. 2023.
\newblock Tabllm: Few-shot classification of tabular data with large language
  models.
\newblock In \emph{AISTATS}, pages 5549--5581. PMLR.

\bibitem[{Hsieh et~al.(2024)Hsieh, Chuang, Li, Wang, Le, Kumar, Glass, Ratner,
  Lee, Krishna et~al.}]{hsieh2024found}
Cheng-Yu Hsieh, Yung-Sung Chuang, Chun-Liang Li, Zifeng Wang, Long~T Le,
  Abhishek Kumar, James Glass, Alexander Ratner, Chen-Yu Lee, Ranjay Krishna,
  et~al. 2024.
\newblock Found in the middle: Calibrating positional attention bias improves
  long context utilization.
\newblock \emph{arXiv preprint arXiv:2406.16008}.

\bibitem[{Ji et~al.(2013)Ji, Li, Yan, Tian, and Zhang}]{ji2013min}
Jianqiu Ji, Jianmin Li, Shuicheng Yan, Qi~Tian, and Bo~Zhang. 2013.
\newblock Min-max hash for jaccard similarity.
\newblock In \emph{ICDM}, pages 301--309. IEEE.

\bibitem[{Jin et~al.(2023)Jin, Wang, Ma, Chu, Zhang, Shi, Chen, Liang, Li, Pan
  et~al.}]{jin2023time}
Ming Jin, Shiyu Wang, Lintao Ma, Zhixuan Chu, James~Y Zhang, Xiaoming Shi,
  Pin-Yu Chen, Yuxuan Liang, Yuan-Fang Li, Shirui Pan, et~al. 2023.
\newblock Time-llm: Time series forecasting by reprogramming large language
  models.
\newblock \emph{arXiv preprint arXiv:2310.01728}.

\bibitem[{Kirk et~al.(2024)Kirk, Wray, Lindes, and Laird}]{kirk2024improving}
James~R Kirk, Robert~E Wray, Peter Lindes, and John~E Laird. 2024.
\newblock Improving knowledge extraction from llms for task learning through
  agent analysis.
\newblock In \emph{AAAI}, volume~38, pages 18390--18398.

\bibitem[{Ko et~al.(2024)Ko, Yang, Han, Kim, Lim, and
  Hormazabal}]{ko2024filling}
Hanbum Ko, Hongjun Yang, Sehui Han, Sungwoong Kim, Sungbin Lim, and Rodrigo
  Hormazabal. 2024.
\newblock Filling in the gaps: Llm-based structured data generation from
  semi-structured scientific data.
\newblock In \emph{ICML Workshop}.

\bibitem[{Lewis et~al.(2020)Lewis, Perez, Piktus, Petroni, Karpukhin, Goyal,
  K{\"u}ttler, Lewis, Yih, Rockt{\"a}schel et~al.}]{lewis2020retrieval}
Patrick Lewis, Ethan Perez, Aleksandra Piktus, Fabio Petroni, Vladimir
  Karpukhin, Naman Goyal, Heinrich K{\"u}ttler, Mike Lewis, Wen-tau Yih, Tim
  Rockt{\"a}schel, et~al. 2020.
\newblock Retrieval-augmented generation for knowledge-intensive nlp tasks.
\newblock \emph{NeurIPS}, 33:9459--9474.

\bibitem[{Li et~al.(2022)Li, Sun, Han, and Li}]{DBLP:journals/tkde/LiSHL22}
Jing Li, Aixin Sun, Jianglei Han, and Chenliang Li. 2022.
\newblock A survey on deep learning for named entity recognition.
\newblock \emph{{IEEE} Trans. Knowl. Data Eng.}, 34(1):50--70.

\bibitem[{Li et~al.(2023)Li, Li, Wang, Li, Sun, Cheng, and
  Yu}]{DBLP:journals/corr/abs-2311-12399}
Yuhan Li, Zhixun Li, Peisong Wang, Jia Li, Xiangguo Sun, Hong Cheng, and
  Jeffrey~Xu Yu. 2023.
\newblock A survey of graph meets large language model: Progress and future
  directions.
\newblock \emph{arXiv preprint arXiv:2311.12399}.

\bibitem[{Liu et~al.(2024)Liu, Lin, Hewitt, Paranjape, Bevilacqua, Petroni, and
  Liang}]{liu2024lost}
Nelson~F Liu, Kevin Lin, John Hewitt, Ashwin Paranjape, Michele Bevilacqua,
  Fabio Petroni, and Percy Liang. 2024.
\newblock Lost in the middle: How language models use long contexts.
\newblock \emph{TACL}, 12:157--173.

\bibitem[{Liu and Gao(2023)}]{DBLP:journals/entropy/LiuG23a}
Shihu Liu and Haiyan Gao. 2023.
\newblock The structure entropy-based node importance ranking method for graph
  data.
\newblock \emph{Entropy}, 25(6):941.

\bibitem[{Luo et~al.(2024)Luo, Song, Huang, Lian, Zhang, Jiang, Xie, and
  Jin}]{luo2024graphinstruct}
Zihan Luo, Xiran Song, Hong Huang, Jianxun Lian, Chenhao Zhang, Jinqi Jiang,
  Xing Xie, and Hai Jin. 2024.
\newblock Graphinstruct: Empowering large language models with graph
  understanding and reasoning capability.
\newblock \emph{arXiv preprint arXiv:2403.04483}.

\bibitem[{Naveed et~al.(2023)Naveed, Khan, Qiu, Saqib, Anwar, Usman, Barnes,
  and Mian}]{naveed2023comprehensive}
Humza Naveed, Asad~Ullah Khan, Shi Qiu, Muhammad Saqib, Saeed Anwar, Muhammad
  Usman, Nick Barnes, and Ajmal Mian. 2023.
\newblock A comprehensive overview of large language models.
\newblock \emph{arXiv preprint arXiv:2307.06435}.

\bibitem[{Pan et~al.(2024)Pan, Luo, Wang, Chen, Wang, and
  Wu}]{DBLP:journals/tkde/PanLWCWW24}
Shirui Pan, Linhao Luo, Yufei Wang, Chen Chen, Jiapu Wang, and Xindong Wu.
  2024.
\newblock Unifying large language models and knowledge graphs: {A} roadmap.
\newblock \emph{TKDE}, 36(7):3580--3599.

\bibitem[{Perozzi et~al.(2024)Perozzi, Fatemi, Zelle, Tsitsulin, Kazemi,
  Al-Rfou, and Halcrow}]{perozzi2024let}
Bryan Perozzi, Bahare Fatemi, Dustin Zelle, Anton Tsitsulin, Mehran Kazemi,
  Rami Al-Rfou, and Jonathan Halcrow. 2024.
\newblock Let your graph do the talking: Encoding structured data for llms.
\newblock \emph{arXiv preprint arXiv:2402.05862}.

\bibitem[{Rorseth et~al.(2024)Rorseth, Godfrey, Golab, Srivastava, and
  Szlichta}]{DBLP:conf/icde/RorsethGGSS24}
Joel Rorseth, Parke Godfrey, Lukasz Golab, Divesh Srivastava, and Jaroslaw
  Szlichta. 2024.
\newblock {RAGE} against the machine: Retrieval-augmented {LLM} explanations.
\newblock In \emph{ICDE}, pages 5469--5472. IEEE.

\bibitem[{Santra(2024)}]{DBLP:conf/sigir/Santra24}
Payel Santra. 2024.
\newblock Leveraging llms for detecting and modeling the propagation of
  misinformation in social networks.
\newblock In \emph{SIGIR}, page 3073. ACM.

\bibitem[{Shang and Huang(2024)}]{shang2024survey}
Wenbo Shang and Xin Huang. 2024.
\newblock A survey of large language models on generative graph analytics:
  Query, learning, and applications.
\newblock \emph{arXiv preprint arXiv:2404.14809}.

\bibitem[{Shen et~al.(2024)Shen, Logeswaran, Lee, Lee, Poria, and
  Mihalcea}]{shen2024understanding}
Siqi Shen, Lajanugen Logeswaran, Moontae Lee, Honglak Lee, Soujanya Poria, and
  Rada Mihalcea. 2024.
\newblock Understanding the capabilities and limitations of large language
  models for cultural commonsense.
\newblock pages 5668--5680.

\bibitem[{Sojitra et~al.(2024)Sojitra, Jain, Saha, Jatowt, and
  Gupta}]{DBLP:conf/sigir/SojitraJ0JG24}
Daivik Sojitra, Raghav Jain, Sriparna Saha, Adam Jatowt, and Manish Gupta.
  2024.
\newblock Timeline summarization in the era of llms.
\newblock In \emph{SIGIR}, pages 2657--2661. {ACM}.

\bibitem[{Sui et~al.(2024)Sui, Zhou, Zhou, Han, and Zhang}]{sui2024table}
Yuan Sui, Mengyu Zhou, Mingjie Zhou, Shi Han, and Dongmei Zhang. 2024.
\newblock Table meets llm: Can large language models understand structured
  table data? a benchmark and empirical study.
\newblock In \emph{WSDM}, pages 645--654.

\bibitem[{Tang et~al.(2024)Tang, Zhang, Li, and Li}]{tang2024grapharena}
Jianheng Tang, Qifan Zhang, Yuhan Li, and Jia Li. 2024.
\newblock Grapharena: Benchmarking large language models on graph computational
  problems.
\newblock \emph{arXiv preprint arXiv:2407.00379}.

\bibitem[{Tang et~al.(2023)Tang, Zhang, Ma, Lin, and Ture}]{tang2023found}
Raphael Tang, Xinyu Zhang, Xueguang Ma, Jimmy Lin, and Ferhan Ture. 2023.
\newblock Found in the middle: Permutation self-consistency improves listwise
  ranking in large language models.
\newblock \emph{arXiv preprint arXiv:2310.07712}.

\bibitem[{Wang et~al.(2024{\natexlab{a}})Wang, Feng, He, Tan, Han, and
  Tsvetkov}]{wang2024can}
Heng Wang, Shangbin Feng, Tianxing He, Zhaoxuan Tan, Xiaochuang Han, and Yulia
  Tsvetkov. 2024{\natexlab{a}}.
\newblock Can language models solve graph problems in natural language?
\newblock \emph{NeurIPS}, 36.

\bibitem[{Wang et~al.(2021)Wang, Huang, Wang, Yuan, Liu, He, and
  Chua}]{wang2021learning}
Xiang Wang, Tinglin Huang, Dingxian Wang, Yancheng Yuan, Zhenguang Liu,
  Xiangnan He, and Tat-Seng Chua. 2021.
\newblock Learning intents behind interactions with knowledge graph for
  recommendation.
\newblock In \emph{WWW}, pages 878--887.

\bibitem[{Wang et~al.(2024{\natexlab{b}})Wang, Salmani, Omidi, Ren,
  Rezagholizadeh, and Eshaghi}]{wang2024}
Xindi Wang, Mahsa Salmani, Parsa Omidi, Xiangyu Ren, Mehdi Rezagholizadeh, and
  Armaghan Eshaghi. 2024{\natexlab{b}}.
\newblock Beyond the limits: A survey of techniques to extend the context
  length in large language models.
\newblock \emph{arXiv preprint arXiv:2402.02244}.

\bibitem[{Wang et~al.(2024{\natexlab{c}})Wang, Chu, Ouyang, Wang, Hao, Shen,
  Gu, Xue, Zhang, Cui, Li, Zhou, and Li}]{DBLP:conf/aaai/WangCOWHSGXZCLZ24}
Yan Wang, Zhixuan Chu, Xin Ouyang, Simeng Wang, Hongyan Hao, Yue Shen, Jinjie
  Gu, Siqiao Xue, James Zhang, Qing Cui, Longfei Li, Jun Zhou, and Sheng Li.
  2024{\natexlab{c}}.
\newblock {LLMRG:} improving recommendations through large language model
  reasoning graphs.
\newblock In \emph{AAAI}, pages 19189--19196.

\bibitem[{Wei et~al.(2022)Wei, Wang, Schuurmans, Bosma, Xia, Chi, Le, Zhou
  et~al.}]{wei2022chain}
Jason Wei, Xuezhi Wang, Dale Schuurmans, Maarten Bosma, Fei Xia, Ed~Chi, Quoc~V
  Le, Denny Zhou, et~al. 2022.
\newblock Chain-of-thought prompting elicits reasoning in large language
  models.
\newblock \emph{NeurIPS}, 35:24824--24837.

\bibitem[{Wu et~al.(2024{\natexlab{a}})}]{wu2024never}
Wu et~al. 2024{\natexlab{a}}.
\newblock Never miss a beat: An efficient recipe for context window extension
  of large language models with consistent" middle" enhancement.
\newblock \emph{arXiv preprint arXiv:2406.07138}.

\bibitem[{Wu et~al.(2024{\natexlab{b}})Wu, Chen, Corcoran, Sra, and
  Singh}]{wu2024grapheval2000}
Qiming Wu, Zichen Chen, Will Corcoran, Misha Sra, and Ambuj~K Singh.
  2024{\natexlab{b}}.
\newblock Grapheval2000: Benchmarking and improving large language models on
  graph datasets.
\newblock \emph{arXiv preprint arXiv:2406.16176}.

\bibitem[{Xiao et~al.(2024)Xiao, Tian, Chen, Han, and
  Lewis}]{xiao2023efficient}
Guangxuan Xiao, Yuandong Tian, Beidi Chen, Song Han, and Mike Lewis. 2024.
\newblock Efficient streaming language models with attention sinks.

\bibitem[{Xu(2021)}]{xu2021understanding}
Mengjia Xu. 2021.
\newblock Understanding graph embedding methods and their applications.
\newblock \emph{SIAM Review}, 63(4):825--853.

\bibitem[{Yu et~al.(2023)Yu, Chen, Ling, Dong, Liu, and Lu}]{yu2023temporal}
Xinli Yu, Zheng Chen, Yuan Ling, Shujing Dong, Zongyi Liu, and Yanbin Lu. 2023.
\newblock Temporal data meets llm--explainable financial time series
  forecasting.
\newblock \emph{arXiv preprint arXiv:2306.11025}.

\bibitem[{Yuan et~al.(2024)Yuan, Liu, Wang, and Qin}]{yuan2024gracore}
Zike Yuan, Ming Liu, Hui Wang, and Bing Qin. 2024.
\newblock Gracore: Benchmarking graph comprehension and complex reasoning in
  large language models.
\newblock \emph{arXiv preprint arXiv:2407.02936}.

\bibitem[{Zhang et~al.(2024{\natexlab{a}})}]{zhang2024attention}
Zhang et~al. 2024{\natexlab{a}}.
\newblock Attention instruction: Amplifying attention in the middle via
  prompting.
\newblock \emph{arXiv preprint arXiv:2406.17095}.

\bibitem[{Zhang et~al.(2024{\natexlab{b}})Zhang, Wang, Zhang, Li, Qin, Wu, and
  Zhu}]{zhang2023llm4dyg}
Zeyang Zhang, Xin Wang, Ziwei Zhang, Haoyang Li, Yijian Qin, Simin Wu, and
  Wenwu Zhu. 2024{\natexlab{b}}.
\newblock Llm4dyg: Can large language models solve problems on dynamic graphs?
\newblock \emph{KDD}.

\bibitem[{Zhao et~al.(2024)Zhao, Huang, Xu, Lin, Liu, and
  Huang}]{zhao2024expel}
Andrew Zhao, Daniel Huang, Quentin Xu, Matthieu Lin, Yong-Jin Liu, and Gao
  Huang. 2024.
\newblock Expel: Llm agents are experiential learners.
\newblock In \emph{AAAI}, volume~38, pages 19632--19642.

\bibitem[{Zhu et~al.(2024)Zhu, Wang, Wu, and Zhang}]{DBLP:conf/aaai/0001WW024}
Guangming Zhu, Siyuan Wang, Tianci Wu, and Liang Zhang. 2024.
\newblock Enhance sketch recognition's explainability via semantic
  component-level parsing.
\newblock In \emph{AAAI}, pages 7731--7738. {AAAI} Press.

\end{thebibliography}

\clearpage
\appendix



\begin{table*}[h!]
	\centering
	\resizebox{\textwidth}{!}{
		\begin{tabular}{lcccc}
			\toprule
			& Task Types & Max Graph Size & Graph Structural Diversity & Multi-step Graph Reasoning \\
			\midrule
			GraphQA \cite{fatemi2023talk} & 6 & 20 & \texttimes & \texttimes \\
			GraphInstruct \cite{luo2024graphinstruct} & 21 & 35 & \texttimes & \texttimes \\
			GraphInstruct(Graphwiz) \cite{chen2024graphwiz} & 9 & 100 & \texttimes & \texttimes \\
			LLM4DyG \cite{zhang2023llm4dyg} & 9 & 20 & \texttimes & \texttimes \\
			GPT4Graph \cite{guo2023gpt4graph} & 10 & 20 & \texttimes & \texttimes \\
			GraphSQA & 20 & 200 & \checkmark & \checkmark \\
			\bottomrule
		\end{tabular}
	}
	\caption{Comparison of Benchmarks for Evaluating Graph Understanding in Large Language Models (LLMs)}
	\label{tab:benchmark}
\end{table*}

\section{Benchmark}\label{Benchmark}
In order to systematically evaluate the ability of large language models (LLMs) to understand and process graph structures, we introduce a comprehensive benchmark GraphSQA, which not only broadens the scope of evaluation by including a wider variety of graph structures but also provides a more rigorous and detailed assessment of both macro- and micro-level graph understanding tasks.

\subsection{Introduction to GraphSQA}

Existing graph understanding benchmarks suffer from several critical limitations, including restricted node coverage, ambiguous task definitions, limited structural diversity, and inadequate support for multi-step graph reasoning. These deficiencies significantly impede a comprehensive evaluation of large language models (LLMs) in their ability to understand and process graph structures effectively. To address these challenges, we introduce \textbf{GraphSQA}, a benchmark meticulously designed to evaluate LLMs across multiple dimensions of graph understanding. 

GraphSQA offers a standardized evaluation suite, enabling consistent and comprehensive assessment of LLMs' capabilities in understanding and reasoning about graph structures. Unlike existing benchmarks that primarily focus on small graphs containing up to 35 nodes, GraphSQA broadens the scope by encompassing a diverse spectrum of node quantities, ranging from 15 to 300 nodes, and includes various graph structures, such as multi-edges and self-loops. The benchmark is organized into two primary categories of graph understanding tasks: \textbf{macro-level tasks} and \textbf{micro-level tasks}, with the latter further subdivided into 15 tasks, including 8 composite tasks. This comprehensive structure provides a more challenging and extensive evaluation environment, better reflecting the complexities encountered in real-world graph-based applications. A comparative overview of existing benchmarks designed to evaluate the graph understanding capabilities of large language models (LLMs) is presented in Table~\ref{tab:benchmark}.

\subsection{Key Features of GraphSQA}
GraphSQA is meticulously designed to provide a standardized evaluation suite for the consistent and comprehensive assessment of LLMs' capabilities in understanding and reasoning about graph structures. Unlike existing benchmarks that primarily focus on small graphs containing 10-35 nodes, GraphSQA broadens the scope by including a wider range of node sizes, extending up to 300 nodes. This expansion provides a more challenging and extensive evaluation environment, better reflecting the complexities encountered in real-world graph-based applications.

\subsubsection{Macro-level Tasks}

Macro-level tasks are designed to assess LLMs' capabilities in understanding the fundamental properties of entire graphs. These tasks involve evaluating the overall graph structure, such as determining the node count, assessing graph connectivity, detecting cycles, identifying the maximum edge weight, and pinpointing nodes with the highest degrees. Each task is carefully crafted to test different aspects of global graph understanding, presenting unique challenges to the models. In total, there are 5 tasks in this category, each contributing to a comprehensive evaluation of a model's ability to grasp the graph's macrostructure.

\subsubsection{Micro-level Tasks}
Micro-level tasks focus on understanding and reasoning about specific nodes or edges within a graph. These tasks include recognizing direct connections between nodes, calculating node degrees, identifying leaf nodes, checking for even degrees, and determining specific edge weights. This category comprises 15 tasks, including 8 composite tasks, each targeting a particular aspect of local graph structure. These tasks are designed to test the models' ability to navigate and reason about graph components on a more granular scale, thus assessing their proficiency in detailed graph analysis.

\subsection{Overview of GraphSQA Categories}

\paragraph{Macro-level Tasks:}
The macro-level tasks are aimed at evaluating a model's ability to comprehend the entire graph's structure. These tasks require the model to analyze the graph as a whole, focusing on its global properties rather than individual components.

\begin{itemize}
	\item \textbf{Node Count Identification:} \textit{How many nodes are in this graph?} 
	This task evaluates the model's ability to accurately determine the total number of nodes present within a given graph. Understanding the node count is fundamental for analyzing the graph's overall size and structure, and serves as a crucial precursor to further global analysis.
	
	\item \textbf{Graph Connectivity Determination:} \textit{Is this graph a connected graph?} 
	This task assesses whether the model can determine if the graph is fully connected, meaning there is a path between every pair of nodes. Understanding graph connectivity is essential for evaluating the graph's structural integrity and the interrelationship between its components, which is critical for various applications in network analysis.
	
	\item \textbf{Cycle Detection:} \textit{Does this graph contain a cycle?} 
	This task examines the model's ability to detect the presence of cycles within the graph, a fundamental aspect of graph theory. Cycle detection is crucial for understanding complex graph behaviors and structures, such as feedback loops and circular dependencies.
	
	\item \textbf{Maximum Edge Weight Identification:} \textit{What is the maximum weight of the edges in this graph?} 
	This task measures the model's capability to identify the heaviest edge within the graph. In weighted graphs, where edge weights represent varying strengths of connections, identifying the maximum weight is important for understanding the distribution of relationships and the overall dynamics of the graph.
	
	\item \textbf{Top-Degree Nodes Identification:} \textit{What are the nodes with the top N highest degrees in this graph?} 
	This task evaluates the model's proficiency in ranking nodes based on their degrees, i.e., the number of edges connected to each node. Identifying top-degree nodes is important for understanding which nodes are most influential or central within the graph, playing key roles in its connectivity and structure.
\end{itemize}

\paragraph{Micro-level Tasks:}
Micro-level tasks require the model to analyze and understand specific elements or substructures within the graph, such as individual nodes or edges. These tasks demand a more detailed and focused approach to graph analysis.

\begin{itemize}
	\item \textbf{Direct Connection Identification:} \textit{Is there a direct connection between node U and node V?} 
	This task tests the model's ability to recognize direct adjacency between two specific nodes within the graph. Understanding direct connections is fundamental to analyzing the local structure of the graph and how nodes are interrelated.
	
	\item \textbf{Node Degree Calculation:} \textit{What is the degree of node N?} 
	This task measures the model's ability to determine the degree of a specific node, which refers to the number of edges connected to it. Node degree is a key indicator of a node's importance and centrality within the graph.
	
	\item \textbf{Leaf Node Identification:} \textit{Is node N a leaf node?} 
	This task evaluates whether the model can correctly identify leaf nodes, which are nodes with only one connection. Leaf nodes are particularly significant in tree-like structures, where they represent terminal nodes or endpoints.
	
	\item \textbf{Even Degree Check:} \textit{Does node N have an even degree?} 
	This task assesses the model's ability to determine whether a given node has an even degree. Understanding degree parity is important for analyzing the graph's structure and potential for certain types of substructures.
	
	\item \textbf{Neighbor Identification:} \textit{Who are the neighbors of node N?} 
	This task tests the model's ability to accurately identify and enumerate the neighbors of a given node. Understanding the local neighborhood is crucial for analyzing the node's immediate environment and its influence within the graph.
	
	\item \textbf{Common Neighbors Identification:} \textit{Do nodes U and V have any common neighbors?} 
	This task evaluates the model's capability to identify shared neighbors between two specific nodes. Understanding common neighbors is important for exploring the overlap in local neighborhoods, which has implications for clustering and community detection.
	
	\item \textbf{Degree Comparison:} \textit{Is the degree of node U greater than the degree of node V?} 
	This task assesses the model's comparative reasoning skills by requiring it to compare the degrees of two nodes. Understanding the relative degrees of nodes is essential for analyzing the distribution of connections and identifying influential nodes.
	
	\item \textbf{Edge Weight Determination:} \textit{What is the weight of the edge between node U and node V?} 
	This task tests the model's ability to determine the specific weight of an edge between two nodes in a weighted graph. Accurately identifying edge weights is crucial for understanding the graph's structure, especially in contexts where edge weights influence the graph's behavior.
	
	\item \textbf{Connected Edges Identification:} \textit{Given the edge (u, v), find all edges connected to it. List the answers in the format of '[(1, 2), (3, 4), ...]'.} 
	This task evaluates the model's ability to identify all edges that are connected to a given edge. Understanding local connectivity around a specified edge is essential for analyzing the graph's structure and how different parts of the graph are interlinked.
	
	\item \textbf{Subgraph Completeness Check:} \textit{Given the nodes [n1, n2, ..., nk], determine if they form a complete subgraph. List the answer directly in the format of 'Yes' or 'No'.} 
	This task assesses the model's ability to determine whether a specified set of nodes forms a complete subgraph, where every pair of nodes is connected by an edge. Understanding subgraph completeness is important for identifying densely connected clusters within the graph.
	
	\item \textbf{Highest-Degree Neighbor Identification:} \textit{Given the node n, find the neighbor's neighbor with the highest degree. List the answer directly as the node id.} 
	This task evaluates the model's ability to identify the neighbor of a given node's neighbor that has the highest degree. This task requires the model to traverse the graph's local neighborhood and perform degree comparisons, which is crucial for understanding local influence and connectivity.
	
	\item \textbf{Third-Order Neighbors Identification:} \textit{Given the node n, find all its third-order neighbors. List the answers in the format of '[1, 2, ...]'.} 
	This task tests the model's ability to identify third-order neighbors, which are nodes three steps away from the given node. Understanding extended neighborhoods is important for analyzing the broader context in which a node operates.
	
	\item \textbf{Direct Neighbor Connection Identification:} \textit{Given the node n and a specified node m, find n's neighbors that are directly connected to m. List the answers in the format of '[1, 2, ...]'.} 
	This task assesses the model's ability to identify which of a given node's neighbors are directly connected to another specified node. This task involves understanding both direct and indirect connections within the graph, which is important for analyzing the graph's local structure.
	
	\item \textbf{Triangles Identification:} \textit{Given the node n, find all triangles (sets of three nodes that are mutually connected) it forms with its neighbors. List the answers in the format of '[(1, 2, 3), (4, 5, 6), ...]'.}  
	This task evaluates the model's ability to identify all triangles that involve a given node and its neighbors. Understanding triangles is fundamental for analyzing local clustering, community structures, and the graph's overall connectivity.

	\item \textbf{Common Third-Order Neighbor Identification:} \textit{Given nodes n and m, find all common third-order neighbors. List the answers in the format of '[1, 2, ...]'.} 
	This task evaluates the model's ability to identify common third-order neighbors between two nodes. This involves analyzing the extended neighborhoods of both nodes and determining the overlap, which is crucial for understanding the broader structure and connectivity within the graph.
\end{itemize}

\begin{table*}[thb]\centering
	\caption{Analysis on Macro- and Micro-level Tasks with Different Baseline Methods and Models.}
	\label{tab:model_performance_transposed}
	\small
	\resizebox{\textwidth}{!}{
		\large
		\begin{tabular}{c|c|c|ccc|ccc|cc|c}
			\toprule
			Models & Tasks & Raw Seq. & \multicolumn{3}{c|}{Prompting Methods} & \multicolumn{3}{c|}{Reordering Methods} & \multicolumn{2}{c|}{Structural Methods} & \textbf{GraphInsight} \\
			& & & COT & FS & BAG & BFS & DFS & SP & AL & AM & \\
			\midrule
			\multirow{3}{*}{Llama-3-8B-Instruct-262k} & Overall & 3 & 9 & 7 & 4 & 5 & 2 & 5 & 8 & 10 & \textbf{1} \\
			& Macro  & 3 & 7 & 4 & 8 & 5 & 2 & 6 & 9 & 10 & \textbf{1} \\
			& Micro   & 3 & 10 & 9 & 2 & 6 & 4 & 5 & 7 & 8 & \textbf{1} \\
			\midrule
			\multirow{3}{*}{Meta-Llama-3-8B-Instruct} & Overall & 6 & 9 & 5 & 8 & 3 & 2 & 4 & 7 & 10 & \textbf{1} \\
			& Macro  & 5 & 8 & 6 & 9 & 2 & 3 & \textbf{1} & 7 & 10 & 4 \\
			& Micro   & 9 & 8 & 5 & 7 & 3 & 2 & 4 & 6 & 10 & \textbf{1} \\
			\midrule
			\multirow{3}{*}{Mistral-7B-Instruct-v0.2} & Overall & 7 & 5 & 9 & 2 & 8 & 6 & 2 & 4 & 10 & \textbf{1} \\
			& Macro  & 5 & 9 & 2 & 6 & 7 & 4 & 1 & 3 & 10 & \textbf{1} \\
			& Micro   & 8 & 3 & 9 & 2 & 6 & 5 & 4 & 7 & 10 & \textbf{1} \\
			\midrule
			\multirow{3}{*}{Qwen2-7B-Instruct} & Overall & 5 & 4 & 9 & 7 & 3 & 8 & 6 & 2 & 10 & \textbf{1} \\
			& Macro  & 7 & 8 & 9 & 6 & 1 & 5 & 3 & 4 & 10 & \textbf{1} \\
			& Micro   & 3 & 2 & 9 & 7 & 6 & 8 & 4 & 5 & 10 & \textbf{1} \\
			\midrule
			\multirow{3}{*}{vicuna-7b-v1.1} & Overall & 5 & 5 & 8 & 7 & 4 & 3 & 2 & 9 & 10 & \textbf{1} \\
			& Macro  & 2 & 3 & 8 & 7 & 5 & 4 & 6 & 9 & 10 & \textbf{1} \\
			& Micro   & 8 & 6 & 9 & 5 & 4 & 3 & 2 & 7 & 10 & \textbf{1} \\
			\bottomrule                        
		\end{tabular}
	}
	\vspace{-10pt}
\end{table*}%

\section{Experimental Settings}\label{app-experiment}

In this section, we detail the computing infrastructure used for conducting our experiments, including both hardware and software configurations.

\textbf{Hardware:} The experiments were conducted on Ubuntu 20.04.2 using four NVIDIA A100 GPUs, each with 80 GB of memory (PCIe interface).

\textbf{Software:} The software environment was configured with Python 3.11.9. The experiments were primarily implemented using PyTorch version 2.3.1. Additionally, we employed the vLLM library (version 0.5.3) for managing large-scale language model inference. The system ran on CUDA version 12.2, optimizing GPU performance for the computations.

\section{Hyperparameter Details}
\label{sec:hyperparameters}

In this section, we list the final hyperparameters used for each model/algorithm in the experiments conducted in this paper, as in Table~\ref{tab:hyperparameters}.

\begin{table}[h!]
	\centering
	\caption{Final hyperparameters used for each model/algorithm in the experiments.}
	\label{tab:hyperparameters}
	\resizebox{\columnwidth}{!}{
		\begin{tabular}{|c|c|c|}
			\hline
			\textbf{Model/Algorithm} & \textbf{Hyperparameter} & \textbf{Value} \\ \hline
			\multirow{2}{*}{\textbf{GraphInsight}} & GraphRAG Base Rate $\gamma\%$ & 0.80 \\ 
			& $\alpha\%$ (Head Memory Region) & 4.5\% \\ 
			& $\beta\%$ (Tail Memory Region) & 10.5\% \\ \hline
			
			\multirow{2}{*}{\textbf{PageRank}} & Damping Factor $\lambda$ & 0.85 \\ 
			& Max Iterations & 100 \\ \hline
			
			\multirow{1}{*}{\textbf{LLMs}} & Positional Bias $\Psi(p)$ & U-shaped \\ \hline
		\end{tabular}
	}

\end{table}

\section{Notations}\label{app-notations}
This section summarizes all notations used throughout this paper, as in Table~\ref{tab-symbols}. 

\begin{table}[htbp!]
	\scriptsize
	\setlength{\extrarowheight}{.095pt}
	\setlength\tabcolsep{2pt}
	\centering
		\caption{Notations used throughout this paper.} 
	\label{tab-symbols}
	\resizebox{\columnwidth}{!}{
		\begin{tabular}{cc}
			\Xhline{2\arrayrulewidth}
			\textbf{Notations} & \textbf{Definitions or Descriptions}\\
			\Xhline{2\arrayrulewidth}
			$G = (V, E)$ & Graph with node set $V$ and edge set $E$\\
			\hline
			$V$ & Set of nodes in the graph\\
			\hline
			$E$ & Set of edges in the graph\\
			\hline
			$v_i, v_j$ & Nodes in the graph\\
			\hline
			$e_{ij}$ & Edge between node $v_i$ and node $v_j$\\
			\hline
			$w_{ij}$ & Weight of edge $e_{ij}$\\
			\hline
			$\mathcal{T}$ & Standard graph description sequence\\
			\hline
			$\mathcal{T}_{s}$ & Subgraph descriptions sequence\\
			\hline
			$t_i$ & Subgraph description\\
			\hline
			$\mathcal{\hat{T}}$ & Reorganized graph description sequence\\
			\hline
			$C_{{\text{LLM}}}$ & Capacity of an LLM for graph understanding\\
			\hline
			$C(\mathcal{T}, p)$ & Comprehension ability at a specific position $p$ within $\mathcal{T}$\\
			\hline
			$\Psi(p)$ & Positional bias curve representing LLM's inherent comprehension at position $p$\\
			\hline
			$\Phi(p)$ & Importance of position $p$ in sequence $\mathcal{T}$\\
			\hline
			$\mathcal{I}(t_i)$ & Importance of subgraph description $t_i$\\
			\hline
			$D_{\text{KL}}$ & Kullback-Leibler divergence\\
			\hline
			$G_i$ & Subgraph centered on node $v_{c_i}$\\
			\hline
			$v_{c_i}$ & Central node of subgraph $G_i$\\
			\hline
			$\text{PR}(v)$ & PageRank score of node $v$\\
			\hline
			$\text{InNb}(v)$ & Set of nodes with edges directed towards node $v$\\
			\hline
			$\text{OutDeg}(u)$ & Number of edges leaving node $u$\\
			\hline
			$\alpha\%$, $\beta\%$ & Proportions defining the head and tail (strong memory regions) of the sequence\\
			\Xhline{2\arrayrulewidth}
		\end{tabular}
	}

\end{table}

\section{Experimental Results for different task types}
The experimental results for different task types on GraphSQA are illustrated in Figures~\ref{fig:llama3-8B} to~\ref{fig:vicuna7b}. These figures analyze macro- and micro-level similarity metrics across various LLM configurations, including Llama-3-8B-Instruct-262k, Meta-Llama-3-8B-Instruct, Mistral-7B-Instruct-v0.2, Qwen2-7B-Instruct, and Vicuna-7b-v1.1.

\begin{figure}[p] 
	\centering
	\begin{subfigure}{0.49\textwidth}
		\includegraphics[width=\linewidth]{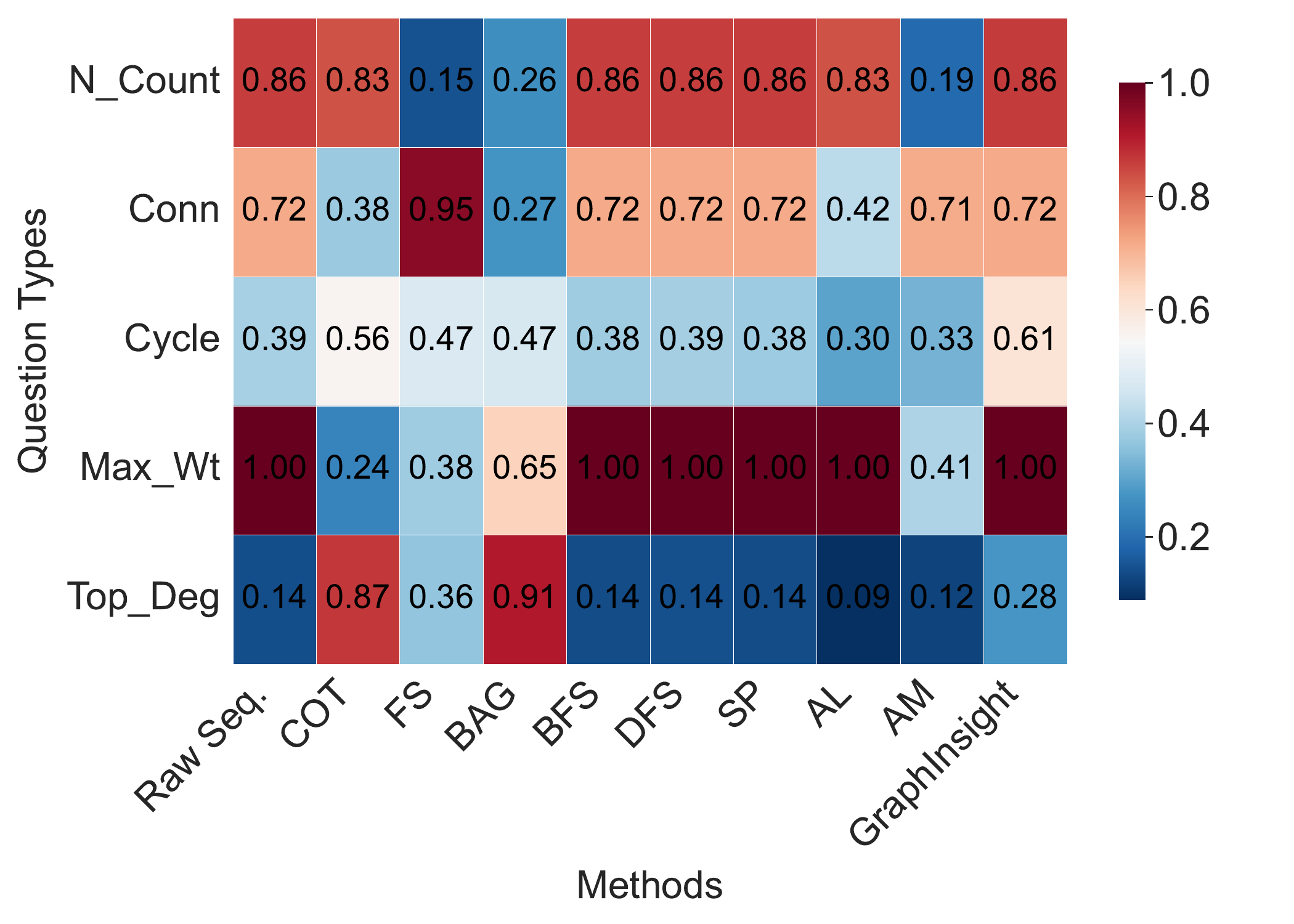}
		\caption{Macro-level (Llama-3-8B-Instruct-262k)}
		\label{fig:llama3-8B-macro}
	\end{subfigure}
	\hfill
	\begin{subfigure}{0.49\textwidth}
		\includegraphics[width=\linewidth]{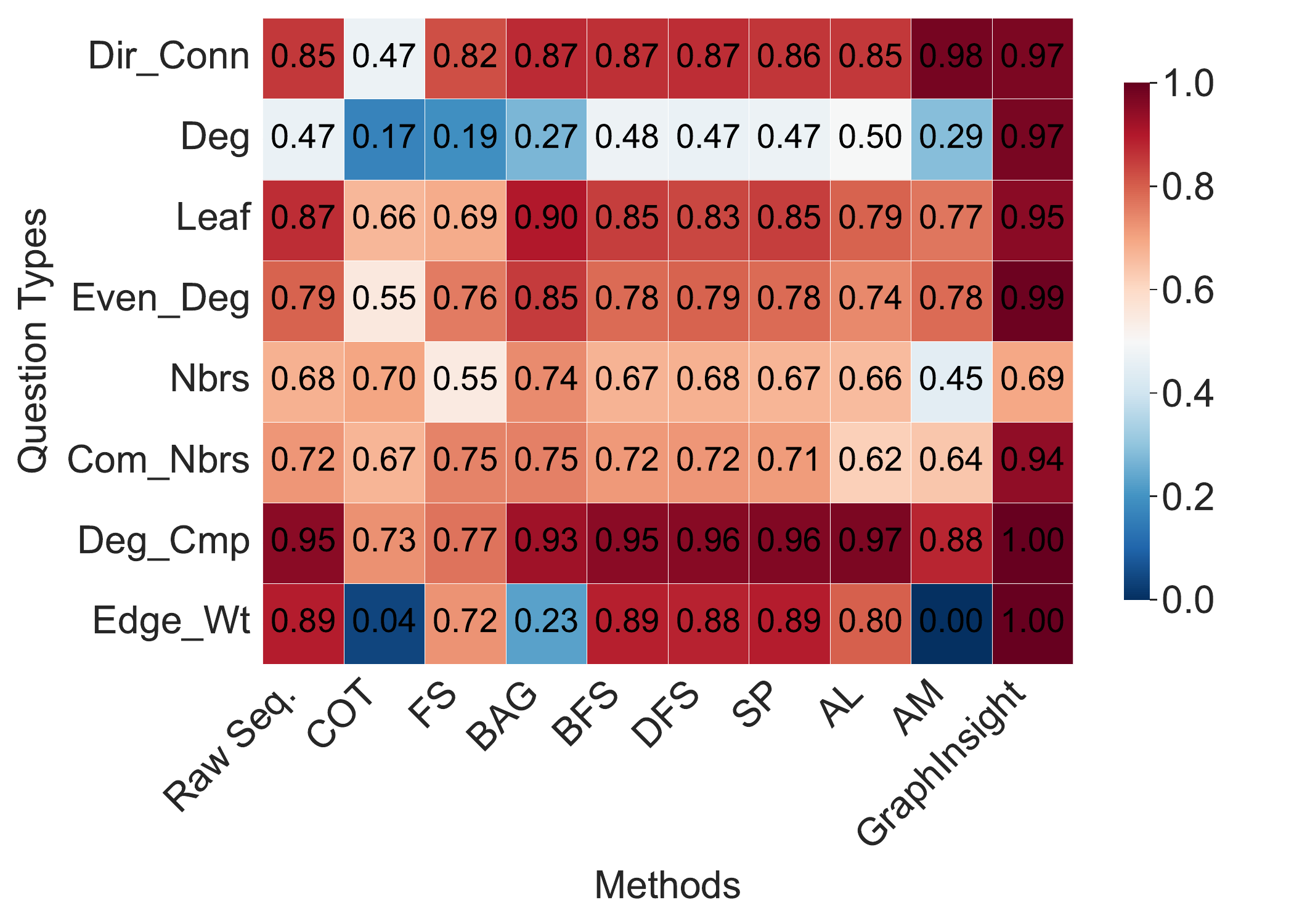}
		\caption{Micro-level (Llama-3-8B-Instruct-262k)}
		\label{fig:llama3-8B-micro}
	\end{subfigure}
	\caption{Macro- and Micro-level Similarity Analysis for Llama-3-8B-Instruct-262k}
	\label{fig:llama3-8B}
\end{figure}

\begin{figure}[p]
	\centering
	\begin{subfigure}{0.49\textwidth}
		\includegraphics[width=\linewidth]{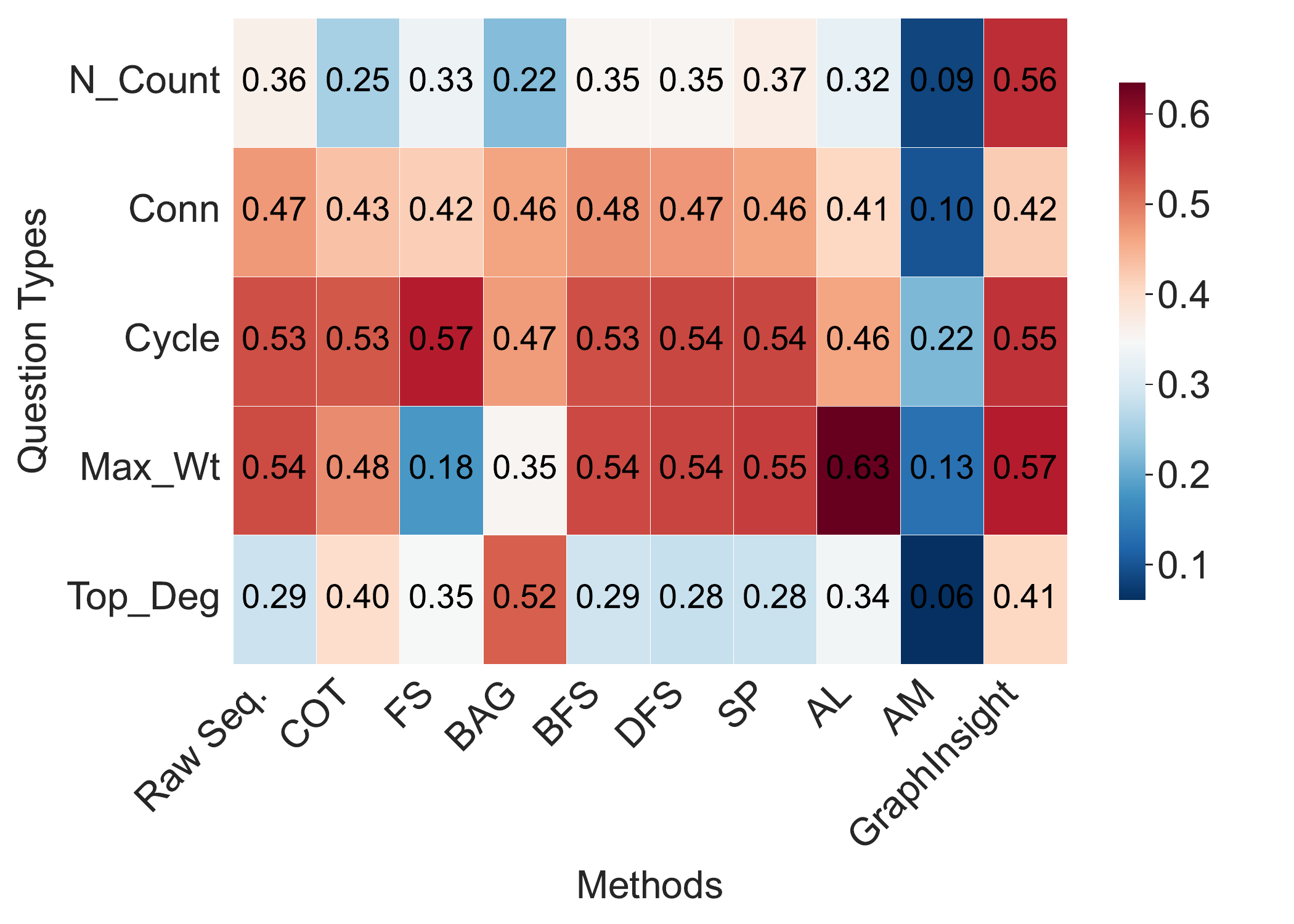}
		\caption{Macro-level (Meta-Llama-3-8B-Instruct)}
		\label{fig:meta-llama3-8B-macro}
	\end{subfigure}
	\hfill
	\begin{subfigure}{0.49\textwidth}
		\includegraphics[width=\linewidth]{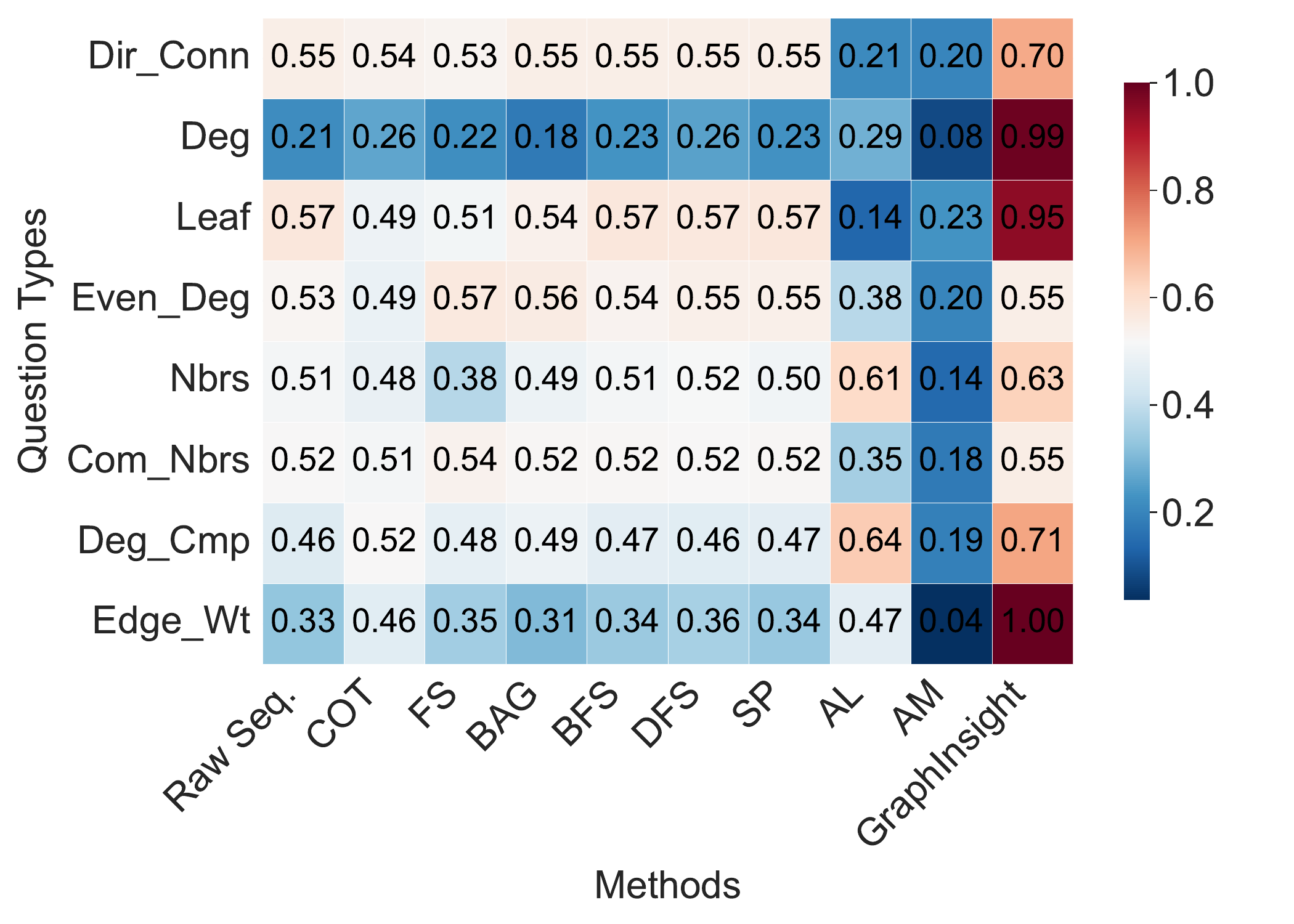}
		\caption{Micro-level (Meta-Llama-3-8B-Instruct)}
		\label{fig:meta-llama3-8B-micro}
	\end{subfigure}
	\caption{Macro- and Micro-level Similarity Analysis for Meta-Llama-3-8B-Instruct}
	\label{fig:meta-llama3-8B}
\end{figure}

\begin{figure}[t]
	\centering
	\begin{subfigure}{0.49\textwidth}
		\includegraphics[width=\linewidth]{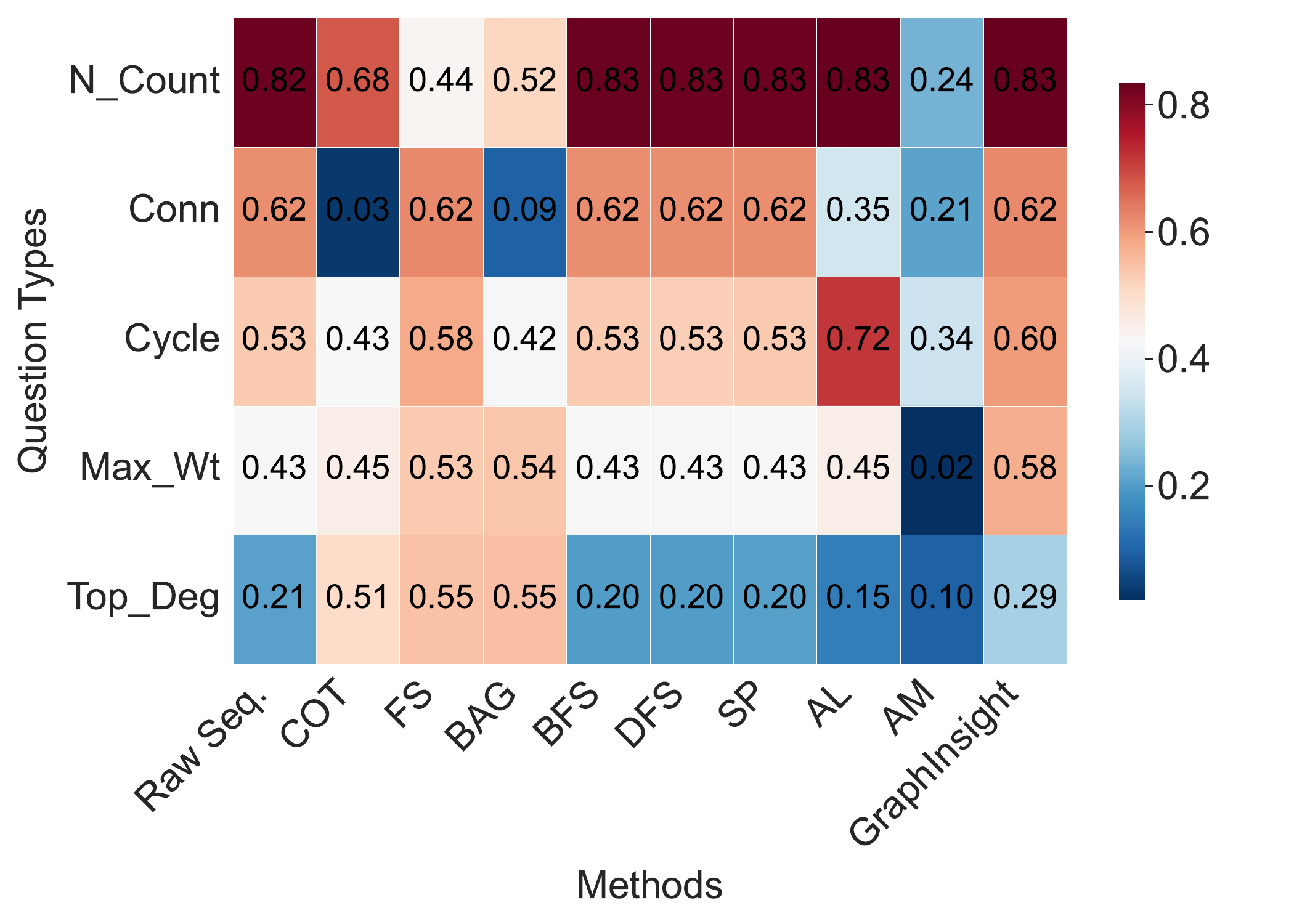}
		\caption{Macro-level (Mistral-7B-Instruct-v0.2)}
		\label{fig:mistral7B-macro}
	\end{subfigure}
	\hfill
	\begin{subfigure}{0.49\textwidth}
		\includegraphics[width=\linewidth]{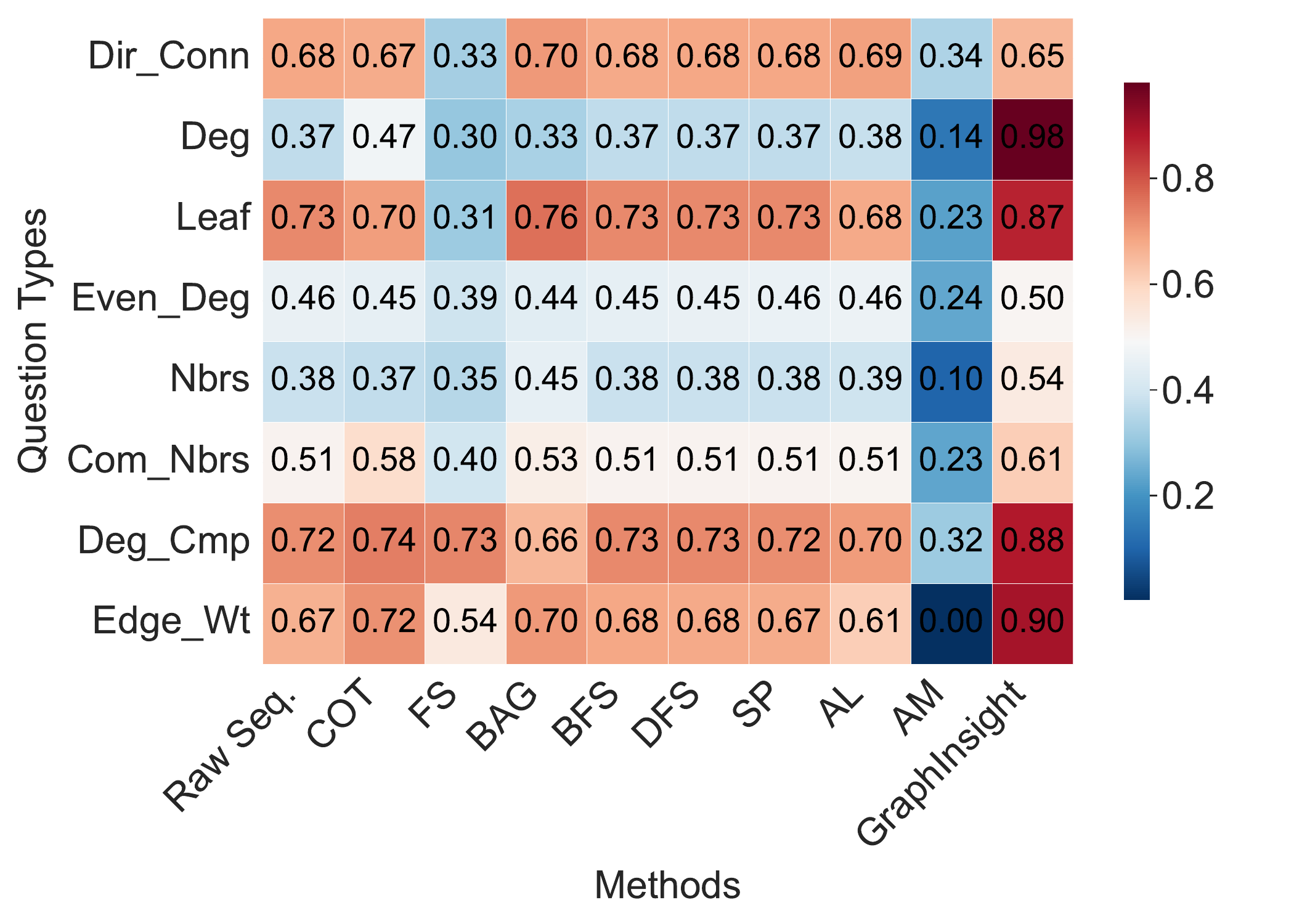}
		\caption{Micro-level (Mistral-7B-Instruct-v0.2)}
		\label{fig:mistral7B-micro}
	\end{subfigure}
	\caption{Macro- and Micro-level Similarity Analysis for Mistral-7B-Instruct-v0.2}
	\label{fig:mistral7B}
\end{figure}

\begin{figure}[t]
	\centering
	\begin{subfigure}{0.49\textwidth}
		\includegraphics[width=\linewidth]{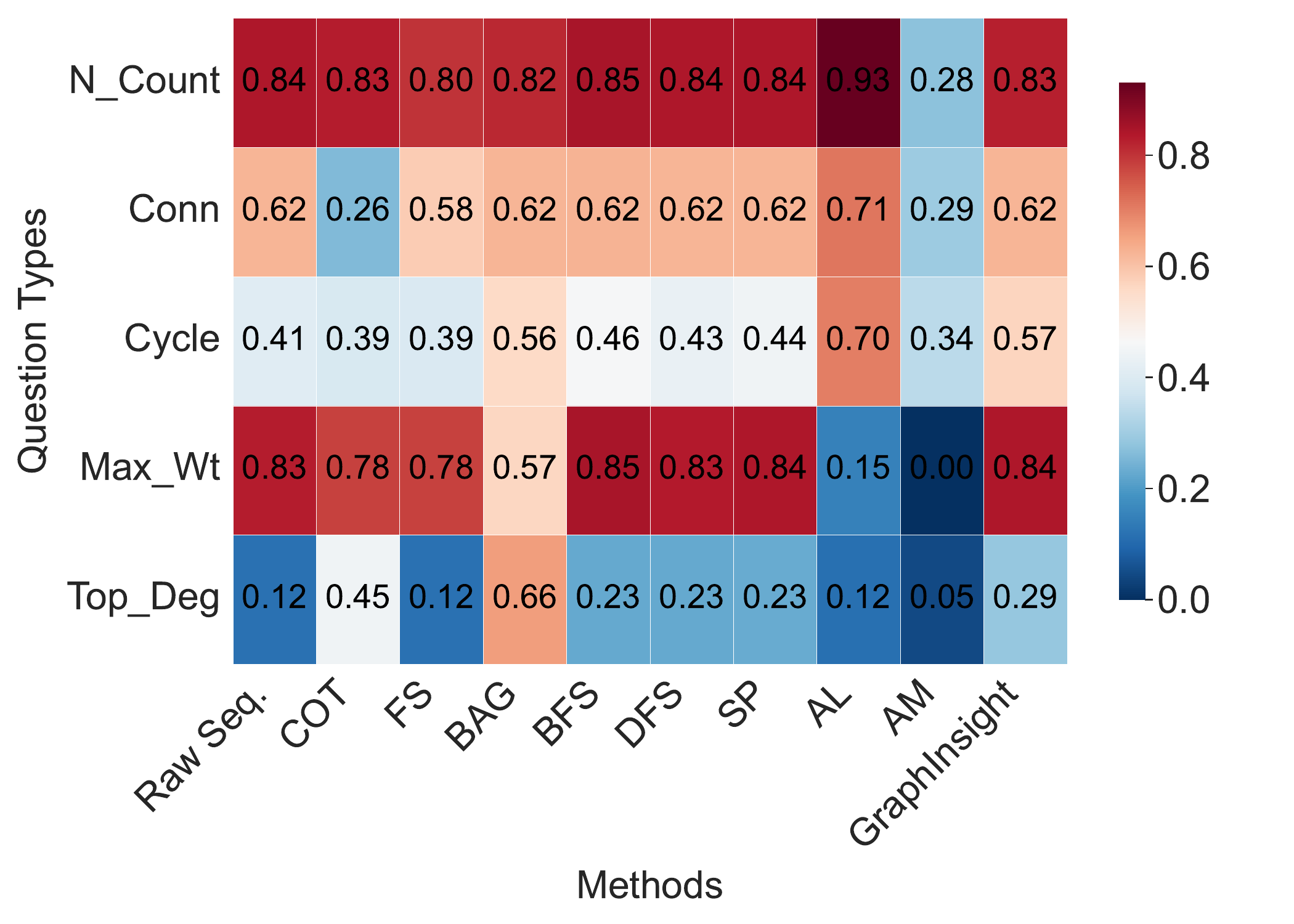}
		\caption{Macro-level (Qwen2-7B-Instruct)}
		\label{fig:qwen2-7B-macro}
	\end{subfigure}
	\hfill
	\begin{subfigure}{0.49\textwidth}
		\includegraphics[width=\linewidth]{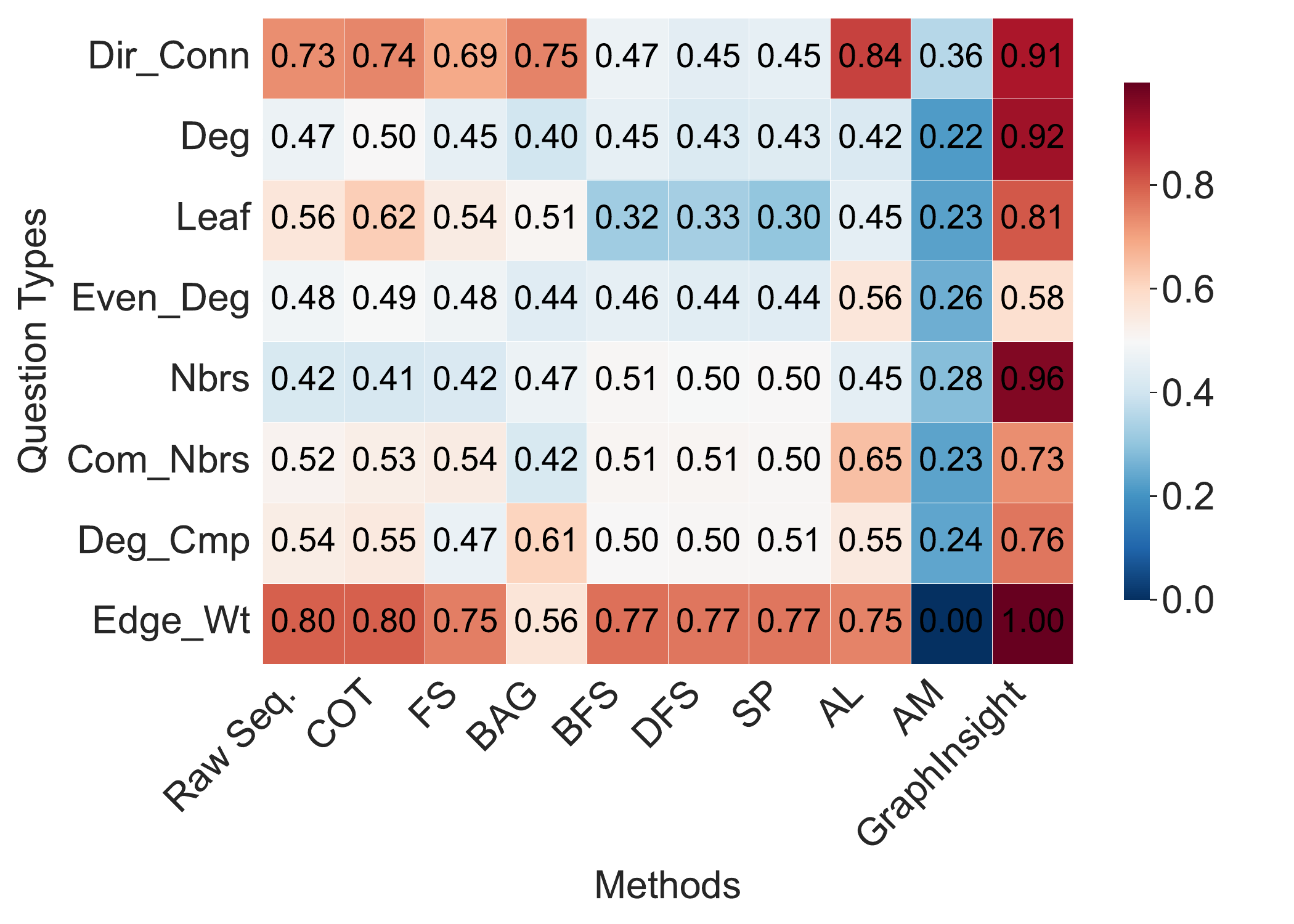}
		\caption{Micro-level (Qwen2-7B-Instruct)}
		\label{fig:qwen2-7B-micro}
	\end{subfigure}
	\caption{Macro- and Micro-level Similarity Analysis for Qwen2-7B-Instruct}
	\label{fig:qwen2-7B}
\end{figure}

\begin{figure*}[t]
	\centering
	\begin{subfigure}{0.49\textwidth}
		\includegraphics[width=\linewidth]{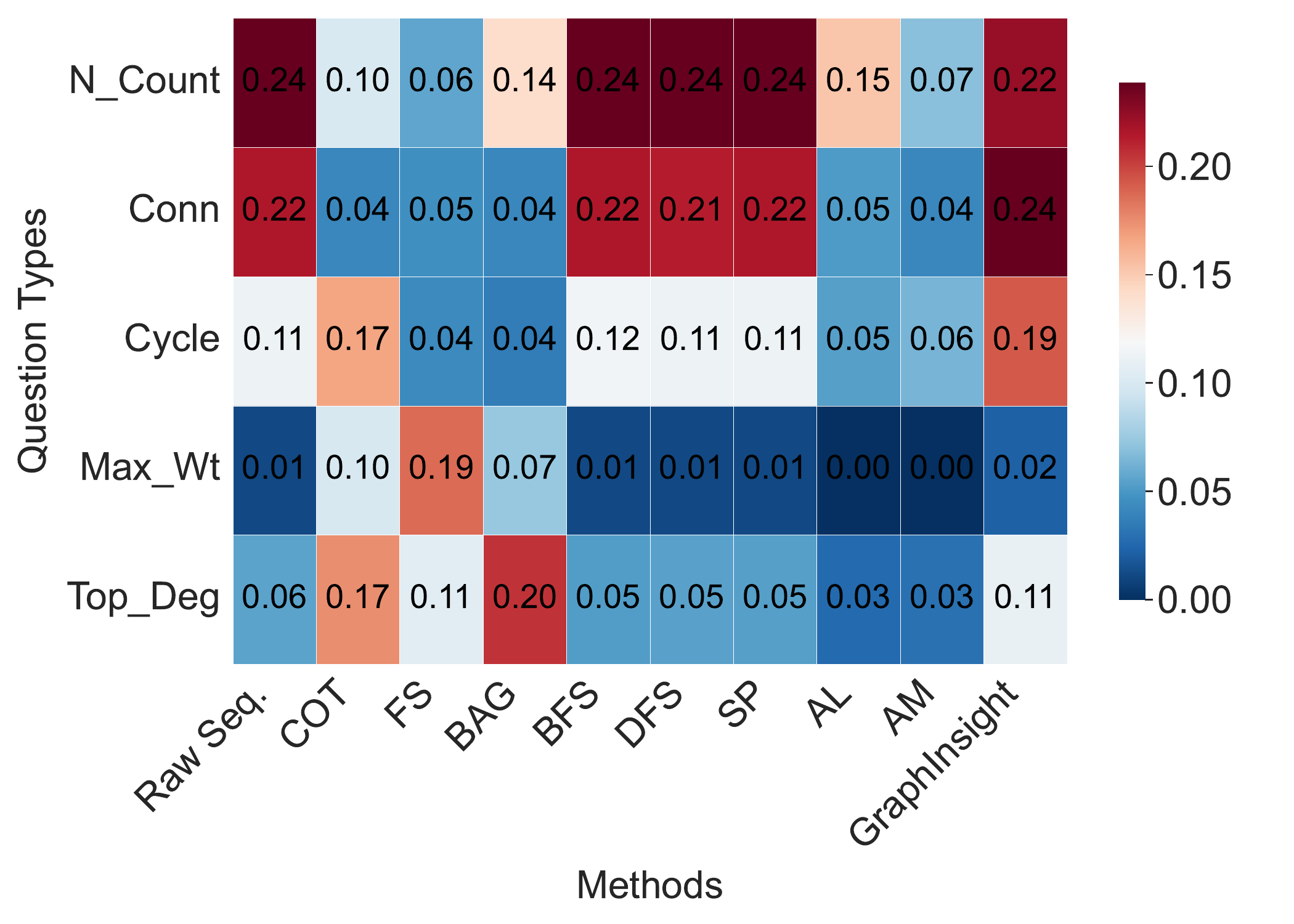}
		\caption{Macro-level (Vicuna-7b-v1.1)}
		\label{fig:vicuna7b-macro}
	\end{subfigure}
	\hfill
	\begin{subfigure}{0.49\textwidth}
		\includegraphics[width=\linewidth]{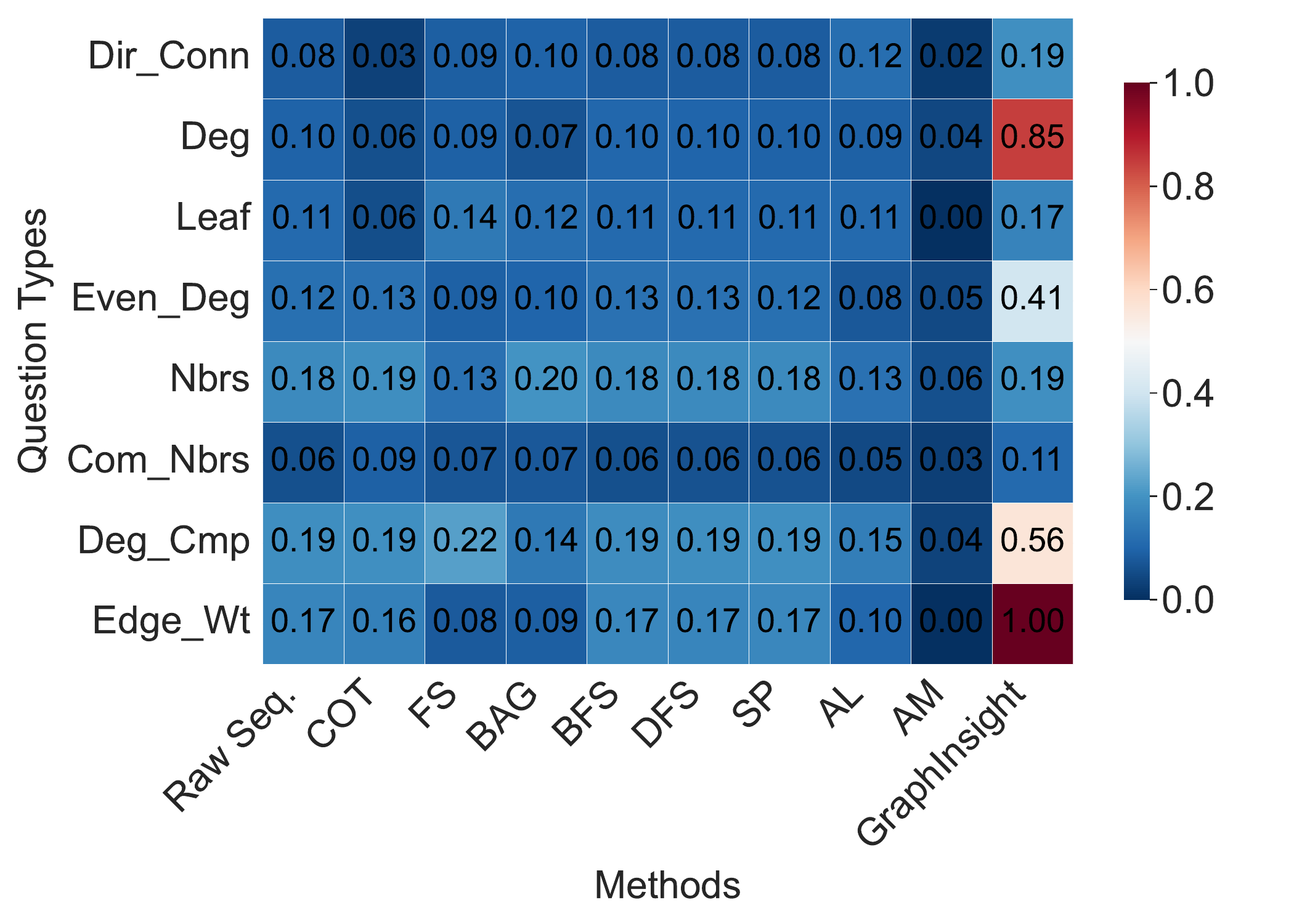}
		\caption{Micro-level (Vicuna-7b-v1.1)}
		\label{fig:vicuna7b-micro}
	\end{subfigure}
	\caption{Macro- and Micro-level Similarity Analysis for Vicuna-7b-v1.1}
	\label{fig:vicuna7b}
\end{figure*}

\section{Metric Analysis}

In the GraphSQA evaluation framework, responses are classified into three distinct categories: boolean, numerical, and set-based. Each category necessitates a specific metric to accurately evaluate the correctness of the predictions.

For \textbf{boolean answers}, which are binary in nature (e.g., true/false), the evaluation is straightforward. A score of 1 is assigned if the predicted answer exactly matches the ground truth; otherwise, the score is 0. This binary scoring system provides an unambiguous measure of correctness, ensuring clarity in the assessment of such responses.

\textbf{Numerical answers}, on the other hand, require a more nuanced scoring approach due to the potential variability in magnitude. To address this, we employ a metric based on relative error, a widely recognized method in numerical evaluations. The score is calculated as one minus the relative error between the predicted value $\tilde{y}$ and the ground truth value $y$ \cite{wang2024can}, formally defined as:

\begin{equation}
	\text{Score} = 1 - \frac{|\tilde{y} - y|}{\max(\tilde{y}, y)}
	\label{eq:relative_error}
\end{equation}

This metric effectively penalizes larger deviations from the ground truth while allowing for partial credit when the prediction is reasonably close. As shown in Equation~\ref{eq:relative_error}, it offers a graded evaluation that more accurately reflects the precision of the numerical predictions.

For \textbf{set-type answers}, which involve comparisons between predicted and ground truth sets (e.g., sets of nodes, edges, or other graph elements), we utilize the Jaccard similarity coefficient as the evaluation metric \cite{ji2013min}. The Jaccard similarity measures the degree of overlap between the predicted set $A$ and the ground truth set $B$, defined as:

\begin{equation}
	\text{Jaccard Similarity} = \frac{|A \cap B|}{|A \cup B|}
	\label{eq:jaccard_similarity}
\end{equation}

As expressed in Equation~\ref{eq:jaccard_similarity}, this similarity measure yields a score ranging from 0 to 1, where 1 indicates a perfect match between the predicted and ground truth sets, and 0 indicates no overlap. By considering both false positives and false negatives, the Jaccard similarity provides a balanced and comprehensive evaluation for set-type answers, effectively capturing both precision and recall in the predictions.

\section{Performance Analysis of the GraphInsight Framework}

The significance of any observed improvement or decline in performance between the proposed GraphInsight framework and the baseline methods is evaluated using appropriate statistical tests. In this study, the Wilcoxon signed-rank test is employed, which is a non-parametric test suitable for comparing paired samples. This test assesses whether the median differences between pairs of observations are statistically significant.

The overall performance comparison between GraphInsight and the baseline method yields a Wilcoxon signed-rank test statistic of 0.0. The corresponding p-value is \(6.103515625 \times 10^{-5}\), indicating a statistically significant improvement at the commonly accepted significance level (e.g., \(\alpha = 0.05\)). The low p-value strongly suggests that the improvements observed with GraphInsight are unlikely to be due to random chance, thereby validating the effectiveness of the proposed framework.

\section{Discussion}
\label{sec:discussion}

In this section, we discuss the implications of our framework and the results from our empirical evaluations. Our work highlights the potential of optimizing LLMs' graph comprehension through carefully designed graph description sequences, particularly when leveraging the sequential format for transforming graphs into a description form conducive to LLMs' understanding.

One of the key findings is the effectiveness of aligning the LLMs' comprehension distribution with the importance distribution of graph description sequences. As demonstrated in Section~3, by reorganizing graph descriptions to position the most important subgraphs in the strong memory regions of LLMs (i.e., the head and tail of the sequence), we significantly enhance the LLMs' performance in macro-level graph understanding tasks. This improvement stems from the better utilization of the LLMs' natural comprehension bias towards the head and tail of sequences, as illustrated by the U-shaped comprehension curve discussed in Definition~2.

Moreover, the decomposition of graph descriptions into subgraph descriptions, centered on high-importance nodes (as measured by PageRank), provides a systematic approach to structuring information in a manner that aligns with the LLMs' strengths. This method not only reinforces the LLMs' ability to process critical structural information but also opens avenues for more complex reasoning tasks that require a deep understanding of local graph structures.

However, several challenges remain. First, while our framework assumes a U-shaped comprehension curve, this may not hold universally across all LLMs or task types. Further research is needed to empirically validate the exact shape of this curve for different models and datasets. Additionally, the process of defining and quantifying the importance of graph structures, while effective in our study, is an open problem and could benefit from more sophisticated techniques beyond PageRank.

In conclusion, our findings underscore the importance of sequence organization in enhancing LLMs' graph comprehension capabilities. By strategically placing high-importance subgraph descriptions in positions that align with the LLMs' natural comprehension tendencies, we can significantly improve performance on a range of graph understanding tasks.

\section{Task Templates in GraphSQA}

In this section, we present the task templates for each category in GraphSQA, as detailed below.

\clearpage

\onecolumn
\begin{tcolorbox}[title=Node Count Identification]
	This is an undirected graph with the following edges:\\
	From node 0 to node 1 with weight 4;\\
	From node 0 to node 2 with weight 4;\\
	From node 0 to node 3 with weight 3;\\
	From node 0 to node 4 with weight 5;\\
	From node 0 to node 5 with weight 3;\\
	From node 0 to node 6 with weight 5;\\
	From node 1 to node 2 with weight 4;\\
	From node 1 to node 3 with weight 5;\\
	From node 1 to node 4 with weight 5;\\
	From node 1 to node 5 with weight 2;\\
	From node 1 to node 6 with weight 1;\\
	From node 2 to node 3 with weight 3;\\
	From node 2 to node 4 with weight 1;\\
	From node 2 to node 5 with weight 2;\\
	From node 2 to node 6 with weight 4;\\
	From node 3 to node 4 with weight 2;\\
	From node 3 to node 5 with weight 1;\\
	From node 3 to node 6 with weight 5;\\
	From node 4 to node 5 with weight 3;\\
	From node 4 to node 6 with weight 1;\\
	From node 5 to node 6 with weight 5;\\
	From node 7 to node 8 with weight 3;\\
	From node 7 to node 9 with weight 5;\\
	From node 7 to node 10 with weight 3;\\
	From node 7 to node 11 with weight 2;\\
	From node 7 to node 12 with weight 4;\\
	From node 7 to node 13 with weight 4;\\
	From node 8 to node 9 with weight 3;\\
	From node 8 to node 10 with weight 3;\\
	From node 8 to node 11 with weight 3;\\
	From node 8 to node 12 with weight 3;\\
	From node 8 to node 13 with weight 2;\\
	From node 9 to node 10 with weight 4;\\
	From node 9 to node 11 with weight 2;\\
	From node 9 to node 12 with weight 5;\\
	From node 9 to node 13 with weight 3;\\
	From node 10 to node 11 with weight 4;\\
	From node 10 to node 12 with weight 3;\\
	From node 10 to node 13 with weight 3;\\
	From node 11 to node 12 with weight 1;\\
	From node 11 to node 13 with weight 5;\\
	From node 12 to node 13 with weight 5;\\
	
	\textbf{Q}: How many nodes are in this graph?\\
	\textbf{A}: 14\\
\end{tcolorbox}

\begin{tcolorbox}[title=Graph Connectivity]
	This is an undirected graph with the following edges:\\
	From node 0 to node 1 with weight 4;\\
	From node 0 to node 2 with weight 4;\\
	From node 0 to node 3 with weight 3;\\
	From node 0 to node 4 with weight 5;\\
	From node 0 to node 5 with weight 3;\\
	From node 0 to node 6 with weight 5;\\
	From node 1 to node 2 with weight 4;\\
	From node 1 to node 3 with weight 5;\\
	From node 1 to node 4 with weight 5;\\
	From node 1 to node 5 with weight 2;\\
	From node 1 to node 6 with weight 1;\\
	From node 2 to node 3 with weight 3;\\
	From node 2 to node 4 with weight 1;\\
	From node 2 to node 5 with weight 2;\\
	From node 2 to node 6 with weight 4;\\
	From node 3 to node 4 with weight 2;\\
	From node 3 to node 5 with weight 1;\\
	From node 3 to node 6 with weight 5;\\
	From node 4 to node 5 with weight 3;\\
	From node 4 to node 6 with weight 1;\\
	From node 5 to node 6 with weight 5;\\
	From node 7 to node 8 with weight 3;\\
	From node 7 to node 9 with weight 5;\\
	From node 7 to node 10 with weight 3;\\
	From node 7 to node 11 with weight 2;\\
	From node 7 to node 12 with weight 4;\\
	From node 7 to node 13 with weight 4;\\
	From node 8 to node 9 with weight 3;\\
	From node 8 to node 10 with weight 3;\\
	From node 8 to node 11 with weight 3;\\
	From node 8 to node 12 with weight 3;\\
	From node 8 to node 13 with weight 2;\\
	From node 9 to node 10 with weight 4;\\
	From node 9 to node 11 with weight 2;\\
	From node 9 to node 12 with weight 5;\\
	From node 9 to node 13 with weight 3;\\
	From node 10 to node 11 with weight 4;\\
	From node 10 to node 12 with weight 3;\\
	From node 10 to node 13 with weight 3;\\
	From node 11 to node 12 with weight 1;\\
	From node 11 to node 13 with weight 5;\\
	From node 12 to node 13 with weight 5;\\\\
	\textbf{Q}: Is this graph a connected graph?\\
	\textbf{A}: No\\
	
\end{tcolorbox}

\begin{tcolorbox}[title=Cycle Detection]
	This is an undirected graph with the following edges:\\
	From node 0 to node 1 with weight 4;\\
	From node 0 to node 2 with weight 4;\\
	From node 0 to node 3 with weight 3;\\
	From node 0 to node 4 with weight 5;\\
	From node 0 to node 5 with weight 3;\\
	From node 0 to node 6 with weight 5;\\
	From node 1 to node 2 with weight 4;\\
	From node 1 to node 3 with weight 5;\\
	From node 1 to node 4 with weight 5;\\
	From node 1 to node 5 with weight 2;\\
	From node 1 to node 6 with weight 1;\\
	From node 2 to node 3 with weight 3;\\
	From node 2 to node 4 with weight 1;\\
	From node 2 to node 5 with weight 2;\\
	From node 2 to node 6 with weight 4;\\
	From node 3 to node 4 with weight 2;\\
	From node 3 to node 5 with weight 1;\\
	From node 3 to node 6 with weight 5;\\
	From node 4 to node 5 with weight 3;\\
	From node 4 to node 6 with weight 1;\\
	From node 5 to node 6 with weight 5;\\
	From node 7 to node 8 with weight 3;\\
	From node 7 to node 9 with weight 5;\\
	From node 7 to node 10 with weight 3;\\
	From node 7 to node 11 with weight 2;\\
	From node 7 to node 12 with weight 4;\\
	From node 7 to node 13 with weight 4;\\
	From node 8 to node 9 with weight 3;\\
	From node 8 to node 10 with weight 3;\\
	From node 8 to node 11 with weight 3;\\
	From node 8 to node 12 with weight 3;\\
	From node 8 to node 13 with weight 2;\\
	From node 9 to node 10 with weight 4;\\
	From node 9 to node 11 with weight 2;\\
	From node 9 to node 12 with weight 5;\\
	From node 9 to node 13 with weight 3;\\
	From node 10 to node 11 with weight 4;\\
	From node 10 to node 12 with weight 3;\\
	From node 10 to node 13 with weight 3;\\
	From node 11 to node 12 with weight 1;\\
	From node 11 to node 13 with weight 5;\\
	From node 12 to node 13 with weight 5;\\\\
	\textbf{Q}: Does this graph contain a cycle?\\
	\textbf{A}: Yes\\
\end{tcolorbox}

\begin{tcolorbox}[title=Maximum Weight Identification]
	This is an undirected graph with the following edges:\\
	From node 0 to node 1 with weight 4;\\
	From node 0 to node 2 with weight 4;\\
	From node 0 to node 3 with weight 3;\\
	From node 0 to node 4 with weight 5;\\
	From node 0 to node 5 with weight 3;\\
	From node 0 to node 6 with weight 5;\\
	From node 1 to node 2 with weight 4;\\
	From node 1 to node 3 with weight 5;\\
	From node 1 to node 4 with weight 5;\\
	From node 1 to node 5 with weight 2;\\
	From node 1 to node 6 with weight 1;\\
	From node 2 to node 3 with weight 3;\\
	From node 2 to node 4 with weight 1;\\
	From node 2 to node 5 with weight 2;\\
	From node 2 to node 6 with weight 4;\\
	From node 3 to node 4 with weight 2;\\
	From node 3 to node 5 with weight 1;\\
	From node 3 to node 6 with weight 5;\\
	From node 4 to node 5 with weight 3;\\
	From node 4 to node 6 with weight 1;\\
	From node 5 to node 6 with weight 5;\\
	From node 7 to node 8 with weight 3;\\
	From node 7 to node 9 with weight 5;\\
	From node 7 to node 10 with weight 3;\\
	From node 7 to node 11 with weight 2;\\
	From node 7 to node 12 with weight 4;\\
	From node 7 to node 13 with weight 4;\\
	From node 8 to node 9 with weight 3;\\
	From node 8 to node 10 with weight 3;\\
	From node 8 to node 11 with weight 3;\\
	From node 8 to node 12 with weight 3;\\
	From node 8 to node 13 with weight 2;\\
	From node 9 to node 10 with weight 4;\\
	From node 9 to node 11 with weight 2;\\
	From node 9 to node 12 with weight 5;\\
	From node 9 to node 13 with weight 3;\\
	From node 10 to node 11 with weight 4;\\
	From node 10 to node 12 with weight 3;\\
	From node 10 to node 13 with weight 3;\\
	From node 11 to node 12 with weight 1;\\
	From node 11 to node 13 with weight 5;\\
	From node 12 to node 13 with weight 5;\\\\
	\textbf{Q}: What is the maximum weight of the edges in this graph?\\
	\textbf{A}: 5\\
\end{tcolorbox}

\begin{tcolorbox}[title=Highest Degree Nodes Identification]
	This is an undirected graph with the following edges:\\
	From node 0 to node 1 with weight 4;\\
	From node 0 to node 2 with weight 4;\\
	From node 0 to node 3 with weight 3;\\
	From node 0 to node 4 with weight 5;\\
	From node 0 to node 5 with weight 3;\\
	From node 0 to node 6 with weight 5;\\
	From node 1 to node 2 with weight 4;\\
	From node 1 to node 3 with weight 5;\\
	From node 1 to node 4 with weight 5;\\
	From node 1 to node 5 with weight 2;\\
	From node 1 to node 6 with weight 1;\\
	From node 2 to node 3 with weight 3;\\
	From node 2 to node 4 with weight 1;\\
	From node 2 to node 5 with weight 2;\\
	From node 2 to node 6 with weight 4;\\
	From node 3 to node 4 with weight 2;\\
	From node 3 to node 5 with weight 1;\\
	From node 3 to node 6 with weight 5;\\
	From node 4 to node 5 with weight 3;\\
	From node 4 to node 6 with weight 1;\\
	From node 5 to node 6 with weight 5;\\
	From node 7 to node 8 with weight 3;\\
	From node 7 to node 9 with weight 5;\\
	From node 7 to node 10 with weight 3;\\
	From node 7 to node 11 with weight 2;\\
	From node 7 to node 12 with weight 4;\\
	From node 7 to node 13 with weight 4;\\
	From node 8 to node 9 with weight 3;\\
	From node 8 to node 10 with weight 3;\\
	From node 8 to node 11 with weight 3;\\
	From node 8 to node 12 with weight 3;\\
	From node 8 to node 13 with weight 2;\\
	From node 9 to node 10 with weight 4;\\
	From node 9 to node 11 with weight 2;\\
	From node 9 to node 12 with weight 5;\\
	From node 9 to node 13 with weight 3;\\
	From node 10 to node 11 with weight 4;\\
	From node 10 to node 12 with weight 3;\\
	From node 10 to node 13 with weight 3;\\
	From node 11 to node 12 with weight 1;\\
	From node 11 to node 13 with weight 5;\\
	From node 12 to node 13 with weight 5;\\\\
	\textbf{Q}: What are the nodes with the top 3 highest degrees in this graph?\\
	\textbf{A}: [(0, 6), (1, 6), (2, 6)]\\
	
\end{tcolorbox}

\begin{tcolorbox}[title=Direct Connection Check]
	This is an undirected graph with the following edges:\\
	From node 0 to node 1 with weight 4;\\
	From node 0 to node 2 with weight 4;\\
	From node 0 to node 3 with weight 3;\\
	From node 0 to node 4 with weight 5;\\
	From node 0 to node 5 with weight 3;\\
	From node 0 to node 6 with weight 5;\\
	From node 1 to node 2 with weight 4;\\
	From node 1 to node 3 with weight 5;\\
	From node 1 to node 4 with weight 5;\\
	From node 1 to node 5 with weight 2;\\
	From node 1 to node 6 with weight 1;\\
	From node 2 to node 3 with weight 3;\\
	From node 2 to node 4 with weight 1;\\
	From node 2 to node 5 with weight 2;\\
	From node 2 to node 6 with weight 4;\\
	From node 3 to node 4 with weight 2;\\
	From node 3 to node 5 with weight 1;\\
	From node 3 to node 6 with weight 5;\\
	From node 4 to node 5 with weight 3;\\
	From node 4 to node 6 with weight 1;\\
	From node 5 to node 6 with weight 5;\\
	From node 7 to node 8 with weight 3;\\
	From node 7 to node 9 with weight 5;\\
	From node 7 to node 10 with weight 3;\\
	From node 7 to node 11 with weight 2;\\
	From node 7 to node 12 with weight 4;\\
	From node 7 to node 13 with weight 4;\\
	From node 8 to node 9 with weight 3;\\
	From node 8 to node 10 with weight 3;\\
	From node 8 to node 11 with weight 3;\\
	From node 8 to node 12 with weight 3;\\
	From node 8 to node 13 with weight 2;\\
	From node 9 to node 10 with weight 4;\\
	From node 9 to node 11 with weight 2;\\
	From node 9 to node 12 with weight 5;\\
	From node 9 to node 13 with weight 3;\\
	From node 10 to node 11 with weight 4;\\
	From node 10 to node 12 with weight 3;\\
	From node 10 to node 13 with weight 3;\\
	From node 11 to node 12 with weight 1;\\
	From node 11 to node 13 with weight 5;\\
	From node 12 to node 13 with weight 5;\\\\
	\textbf{Q}: Is there a direct connection between node 8 and node 2?\\
	\textbf{A}: No\\
\end{tcolorbox}

\begin{tcolorbox}[title=Node Degree Calculation]
	This is an undirected graph with the following edges:\\
	From node 0 to node 1 with weight 4;\\
	From node 0 to node 2 with weight 4;\\
	From node 0 to node 3 with weight 3;\\
	From node 0 to node 4 with weight 5;\\
	From node 0 to node 5 with weight 3;\\
	From node 0 to node 6 with weight 5;\\
	From node 1 to node 2 with weight 4;\\
	From node 1 to node 3 with weight 5;\\
	From node 1 to node 4 with weight 5;\\
	From node 1 to node 5 with weight 2;\\
	From node 1 to node 6 with weight 1;\\
	From node 2 to node 3 with weight 3;\\
	From node 2 to node 4 with weight 1;\\
	From node 2 to node 5 with weight 2;\\
	From node 2 to node 6 with weight 4;\\
	From node 3 to node 4 with weight 2;\\
	From node 3 to node 5 with weight 1;\\
	From node 3 to node 6 with weight 5;\\
	From node 4 to node 5 with weight 3;\\
	From node 4 to node 6 with weight 1;\\
	From node 5 to node 6 with weight 5;\\
	From node 7 to node 8 with weight 3;\\
	From node 7 to node 9 with weight 5;\\
	From node 7 to node 10 with weight 3;\\
	From node 7 to node 11 with weight 2;\\
	From node 7 to node 12 with weight 4;\\
	From node 7 to node 13 with weight 4;\\
	From node 8 to node 9 with weight 3;\\
	From node 8 to node 10 with weight 3;\\
	From node 8 to node 11 with weight 3;\\
	From node 8 to node 12 with weight 3;\\
	From node 8 to node 13 with weight 2;\\
	From node 9 to node 10 with weight 4;\\
	From node 9 to node 11 with weight 2;\\
	From node 9 to node 12 with weight 5;\\
	From node 9 to node 13 with weight 3;\\
	From node 10 to node 11 with weight 4;\\
	From node 10 to node 12 with weight 3;\\
	From node 10 to node 13 with weight 3;\\
	From node 11 to node 12 with weight 1;\\
	From node 11 to node 13 with weight 5;\\
	From node 12 to node 13 with weight 5;\\\\
	\textbf{Q}: What is the degree of node 12?\\
	\textbf{A}: 6\\
\end{tcolorbox}

\begin{tcolorbox}[title=Leaf Node Check]
	This is an undirected graph with the following edges:\\
	From node 0 to node 1 with weight 4;\\
	From node 0 to node 2 with weight 4;\\
	From node 0 to node 3 with weight 3;\\
	From node 0 to node 4 with weight 5;\\
	From node 0 to node 5 with weight 3;\\
	From node 0 to node 6 with weight 5;\\
	From node 1 to node 2 with weight 4;\\
	From node 1 to node 3 with weight 5;\\
	From node 1 to node 4 with weight 5;\\
	From node 1 to node 5 with weight 2;\\
	From node 1 to node 6 with weight 1;\\
	From node 2 to node 3 with weight 3;\\
	From node 2 to node 4 with weight 1;\\
	From node 2 to node 5 with weight 2;\\
	From node 2 to node 6 with weight 4;\\
	From node 3 to node 4 with weight 2;\\
	From node 3 to node 5 with weight 1;\\
	From node 3 to node 6 with weight 5;\\
	From node 4 to node 5 with weight 3;\\
	From node 4 to node 6 with weight 1;\\
	From node 5 to node 6 with weight 5;\\
	From node 7 to node 8 with weight 3;\\
	From node 7 to node 9 with weight 5;\\
	From node 7 to node 10 with weight 3;\\
	From node 7 to node 11 with weight 2;\\
	From node 7 to node 12 with weight 4;\\
	From node 7 to node 13 with weight 4;\\
	From node 8 to node 9 with weight 3;\\
	From node 8 to node 10 with weight 3;\\
	From node 8 to node 11 with weight 3;\\
	From node 8 to node 12 with weight 3;\\
	From node 8 to node 13 with weight 2;\\
	From node 9 to node 10 with weight 4;\\
	From node 9 to node 11 with weight 2;\\
	From node 9 to node 12 with weight 5;\\
	From node 9 to node 13 with weight 3;\\
	From node 10 to node 11 with weight 4;\\
	From node 10 to node 12 with weight 3;\\
	From node 10 to node 13 with weight 3;\\
	From node 11 to node 12 with weight 1;\\
	From node 11 to node 13 with weight 5;\\
	From node 12 to node 13 with weight 5;\\\\
	\textbf{Q}: Is node 0 a leaf node?\\
	\textbf{A}: No\\
\end{tcolorbox}

\begin{tcolorbox}[title=Even Degree Check]
	This is an undirected graph with the following edges:\\
	From node 0 to node 1 with weight 4;\\
	From node 0 to node 2 with weight 4;\\
	From node 0 to node 3 with weight 3;\\
	From node 0 to node 4 with weight 5;\\
	From node 0 to node 5 with weight 3;\\
	From node 0 to node 6 with weight 5;\\
	From node 1 to node 2 with weight 4;\\
	From node 1 to node 3 with weight 5;\\
	From node 1 to node 4 with weight 5;\\
	From node 1 to node 5 with weight 2;\\
	From node 1 to node 6 with weight 1;\\
	From node 2 to node 3 with weight 3;\\
	From node 2 to node 4 with weight 1;\\
	From node 2 to node 5 with weight 2;\\
	From node 2 to node 6 with weight 4;\\
	From node 3 to node 4 with weight 2;\\
	From node 3 to node 5 with weight 1;\\
	From node 3 to node 6 with weight 5;\\
	From node 4 to node 5 with weight 3;\\
	From node 4 to node 6 with weight 1;\\
	From node 5 to node 6 with weight 5;\\
	From node 7 to node 8 with weight 3;\\
	From node 7 to node 9 with weight 5;\\
	From node 7 to node 10 with weight 3;\\
	From node 7 to node 11 with weight 2;\\
	From node 7 to node 12 with weight 4;\\
	From node 7 to node 13 with weight 4;\\
	From node 8 to node 9 with weight 3;\\
	From node 8 to node 10 with weight 3;\\
	From node 8 to node 11 with weight 3;\\
	From node 8 to node 12 with weight 3;\\
	From node 8 to node 13 with weight 2;\\
	From node 9 to node 10 with weight 4;\\
	From node 9 to node 11 with weight 2;\\
	From node 9 to node 12 with weight 5;\\
	From node 9 to node 13 with weight 3;\\
	From node 10 to node 11 with weight 4;\\
	From node 10 to node 12 with weight 3;\\
	From node 10 to node 13 with weight 3;\\
	From node 11 to node 12 with weight 1;\\
	From node 11 to node 13 with weight 5;\\
	From node 12 to node 13 with weight 5;\\\\
	\textbf{Q}: Does node 12 have an even degree?\\
	\textbf{A}: Yes\\
\end{tcolorbox}

\begin{tcolorbox}[title=Neighbor Nodes Identification]
	This is an undirected graph with the following edges:\\
	From node 0 to node 1 with weight 4;\\
	From node 0 to node 2 with weight 4;\\
	From node 0 to node 3 with weight 3;\\
	From node 0 to node 4 with weight 5;\\
	From node 0 to node 5 with weight 3;\\
	From node 0 to node 6 with weight 5;\\
	From node 1 to node 2 with weight 4;\\
	From node 1 to node 3 with weight 5;\\
	From node 1 to node 4 with weight 5;\\
	From node 1 to node 5 with weight 2;\\
	From node 1 to node 6 with weight 1;\\
	From node 2 to node 3 with weight 3;\\
	From node 2 to node 4 with weight 1;\\
	From node 2 to node 5 with weight 2;\\
	From node 2 to node 6 with weight 4;\\
	From node 3 to node 4 with weight 2;\\
	From node 3 to node 5 with weight 1;\\
	From node 3 to node 6 with weight 5;\\
	From node 4 to node 5 with weight 3;\\
	From node 4 to node 6 with weight 1;\\
	From node 5 to node 6 with weight 5;\\
	From node 7 to node 8 with weight 3;\\
	From node 7 to node 9 with weight 5;\\
	From node 7 to node 10 with weight 3;\\
	From node 7 to node 11 with weight 2;\\
	From node 7 to node 12 with weight 4;\\
	From node 7 to node 13 with weight 4;\\
	From node 8 to node 9 with weight 3;\\
	From node 8 to node 10 with weight 3;\\
	From node 8 to node 11 with weight 3;\\
	From node 8 to node 12 with weight 3;\\
	From node 8 to node 13 with weight 2;\\
	From node 9 to node 10 with weight 4;\\
	From node 9 to node 11 with weight 2;\\
	From node 9 to node 12 with weight 5;\\
	From node 9 to node 13 with weight 3;\\
	From node 10 to node 11 with weight 4;\\
	From node 10 to node 12 with weight 3;\\
	From node 10 to node 13 with weight 3;\\
	From node 11 to node 12 with weight 1;\\
	From node 11 to node 13 with weight 5;\\
	From node 12 to node 13 with weight 5;\\\\
	\textbf{Q}: Who are the neighbors of node 12?\\
	\textbf{A}: [7, 8, 9, 10, 11, 13]\\
\end{tcolorbox}

\begin{tcolorbox}[title=Common Neighbors Identification]
	This is an undirected graph with the following edges:\\
	From node 0 to node 1 with weight 4;\\
	From node 0 to node 2 with weight 4;\\
	From node 0 to node 3 with weight 3;\\
	From node 0 to node 4 with weight 5;\\
	From node 0 to node 5 with weight 3;\\
	From node 0 to node 6 with weight 5;\\
	From node 1 to node 2 with weight 4;\\
	From node 1 to node 3 with weight 5;\\
	From node 1 to node 4 with weight 5;\\
	From node 1 to node 5 with weight 2;\\
	From node 1 to node 6 with weight 1;\\
	From node 2 to node 3 with weight 3;\\
	From node 2 to node 4 with weight 1;\\
	From node 2 to node 5 with weight 2;\\
	From node 2 to node 6 with weight 4;\\
	From node 3 to node 4 with weight 2;\\
	From node 3 to node 5 with weight 1;\\
	From node 3 to node 6 with weight 5;\\
	From node 4 to node 5 with weight 3;\\
	From node 4 to node 6 with weight 1;\\
	From node 5 to node 6 with weight 5;\\
	From node 7 to node 8 with weight 3;\\
	From node 7 to node 9 with weight 5;\\
	From node 7 to node 10 with weight 3;\\
	From node 7 to node 11 with weight 2;\\
	From node 7 to node 12 with weight 4;\\
	From node 7 to node 13 with weight 4;\\
	From node 8 to node 9 with weight 3;\\
	From node 8 to node 10 with weight 3;\\
	From node 8 to node 11 with weight 3;\\
	From node 8 to node 12 with weight 3;\\
	From node 8 to node 13 with weight 2;\\
	From node 9 to node 10 with weight 4;\\
	From node 9 to node 11 with weight 2;\\
	From node 9 to node 12 with weight 5;\\
	From node 9 to node 13 with weight 3;\\
	From node 10 to node 11 with weight 4;\\
	From node 10 to node 12 with weight 3;\\
	From node 10 to node 13 with weight 3;\\
	From node 11 to node 12 with weight 1;\\
	From node 11 to node 13 with weight 5;\\
	From node 12 to node 13 with weight 5;\\\\
	\textbf{Q}: Do nodes 7 and 11 have any common neighbors?\\
	\textbf{A}: Yes, node 8.\\
\end{tcolorbox}

\begin{tcolorbox}[title=Degree Comparison]
	This is an undirected graph with the following edges:\\
	From node 0 to node 1 with weight 4;\\
	From node 0 to node 2 with weight 4;\\
	From node 0 to node 3 with weight 3;\\
	From node 0 to node 4 with weight 5;\\
	From node 0 to node 5 with weight 3;\\
	From node 0 to node 6 with weight 5;\\
	From node 1 to node 2 with weight 4;\\
	From node 1 to node 3 with weight 5;\\
	From node 1 to node 4 with weight 5;\\
	From node 1 to node 5 with weight 2;\\
	From node 1 to node 6 with weight 1;\\
	From node 2 to node 3 with weight 3;\\
	From node 2 to node 4 with weight 1;\\
	From node 2 to node 5 with weight 2;\\
	From node 2 to node 6 with weight 4;\\
	From node 3 to node 4 with weight 2;\\
	From node 3 to node 5 with weight 1;\\
	From node 3 to node 6 with weight 5;\\
	From node 4 to node 5 with weight 3;\\
	From node 4 to node 6 with weight 1;\\
	From node 5 to node 6 with weight 5;\\
	From node 7 to node 8 with weight 3;\\
	From node 7 to node 9 with weight 5;\\
	From node 7 to node 10 with weight 3;\\
	From node 7 to node 11 with weight 2;\\
	From node 7 to node 12 with weight 4;\\
	From node 7 to node 13 with weight 4;\\
	From node 8 to node 9 with weight 3;\\
	From node 8 to node 10 with weight 3;\\
	From node 8 to node 11 with weight 3;\\
	From node 8 to node 12 with weight 3;\\
	From node 8 to node 13 with weight 2;\\
	From node 9 to node 10 with weight 4;\\
	From node 9 to node 11 with weight 2;\\
	From node 9 to node 12 with weight 5;\\
	From node 9 to node 13 with weight 3;\\
	From node 10 to node 11 with weight 4;\\
	From node 10 to node 12 with weight 3;\\
	From node 10 to node 13 with weight 3;\\
	From node 11 to node 12 with weight 1;\\
	From node 11 to node 13 with weight 5;\\
	From node 12 to node 13 with weight 5;\\\\
	\textbf{Q}: Is the degree of node 2 greater than the degree of node 6?\\
	\textbf{A}: No\\
	
\end{tcolorbox}

\begin{tcolorbox}[title=Edge Weight Identification]
	This is an undirected graph with the following edges:\\
	From node 0 to node 1 with weight 4;\\
	From node 0 to node 2 with weight 4;\\
	From node 0 to node 3 with weight 3;\\
	From node 0 to node 4 with weight 5;\\
	From node 0 to node 5 with weight 3;\\
	From node 0 to node 6 with weight 5;\\
	From node 1 to node 2 with weight 4;\\
	From node 1 to node 3 with weight 5;\\
	From node 1 to node 4 with weight 5;\\
	From node 1 to node 5 with weight 2;\\
	From node 1 to node 6 with weight 1;\\
	From node 2 to node 3 with weight 3;\\
	From node 2 to node 4 with weight 1;\\
	From node 2 to node 5 with weight 2;\\
	From node 2 to node 6 with weight 4;\\
	From node 3 to node 4 with weight 2;\\
	From node 3 to node 5 with weight 1;\\
	From node 3 to node 6 with weight 5;\\
	From node 4 to node 5 with weight 3;\\
	From node 4 to node 6 with weight 1;\\
	From node 5 to node 6 with weight 5;\\
	From node 7 to node 8 with weight 3;\\
	From node 7 to node 9 with weight 5;\\
	From node 7 to node 10 with weight 3;\\
	From node 7 to node 11 with weight 2;\\
	From node 7 to node 12 with weight 4;\\
	From node 7 to node 13 with weight 4;\\
	From node 8 to node 9 with weight 3;\\
	From node 8 to node 10 with weight 3;\\
	From node 8 to node 11 with weight 3;\\
	From node 8 to node 12 with weight 3;\\
	From node 8 to node 13 with weight 2;\\
	From node 9 to node 10 with weight 4;\\
	From node 9 to node 11 with weight 2;\\
	From node 9 to node 12 with weight 5;\\
	From node 9 to node 13 with weight 3;\\
	From node 10 to node 11 with weight 4;\\
	From node 10 to node 12 with weight 3;\\
	From node 10 to node 13 with weight 3;\\
	From node 11 to node 12 with weight 1;\\
	From node 11 to node 13 with weight 5;\\
	From node 12 to node 13 with weight 5;\\\\
	\textbf{Q}: What is the weight of the edge between node 12 and node 13?\\
	\textbf{A}: 5\\
\end{tcolorbox}

\begin{tcolorbox}[title=Find All Connected Edges]
	This is an undirected graph with the following edges:\\
	From node 0 to node 1 with weight 4;\\
	From node 0 to node 2 with weight 1;\\
	From node 0 to node 3 with weight 2;\\
	From node 0 to node 4 with weight 5;\\
	From node 0 to node 5 with weight 4;\\
	From node 0 to node 6 with weight 3;\\
	From node 1 to node 2 with weight 1;\\
	From node 1 to node 3 with weight 2;\\
	From node 1 to node 4 with weight 2;\\
	From node 1 to node 5 with weight 1;\\
	From node 1 to node 6 with weight 1;\\
	From node 2 to node 3 with weight 1;\\
	From node 2 to node 4 with weight 4;\\
	From node 2 to node 5 with weight 3;\\
	From node 2 to node 6 with weight 4;\\
	From node 3 to node 4 with weight 1;\\
	From node 3 to node 5 with weight 1;\\
	From node 3 to node 6 with weight 5;\\
	From node 4 to node 5 with weight 4;\\
	From node 4 to node 6 with weight 2;\\
	From node 5 to node 6 with weight 1;\\
	From node 7 to node 8 with weight 3;\\
	From node 7 to node 9 with weight 5;\\
	From node 7 to node 10 with weight 5;\\
	From node 7 to node 11 with weight 2;\\
	From node 7 to node 12 with weight 2;\\
	From node 7 to node 13 with weight 4;\\
	From node 8 to node 9 with weight 5;\\
	From node 8 to node 10 with weight 1;\\
	From node 8 to node 11 with weight 4;\\
	From node 8 to node 12 with weight 5;\\
	From node 8 to node 13 with weight 3;\\
	From node 9 to node 10 with weight 3;\\
	From node 9 to node 11 with weight 2;\\
	From node 9 to node 12 with weight 2;\\
	From node 9 to node 13 with weight 1;\\
	From node 10 to node 11 with weight 1;\\
	From node 10 to node 12 with weight 2;\\
	From node 10 to node 13 with weight 4;\\
	From node 11 to node 12 with weight 5;\\
	From node 11 to node 13 with weight 5;\\
	From node 12 to node 13 with weight 5;\\\\
	\textbf{Q}: Given the edge (8, 11), find all edges connected to it. List the answers in the format of '[(1, 2), (3, 4), ...]'.\\
	\textbf{A}: \[
	\left[
	\begin{array}{c}
		(8, 7), (8, 9), (8, 10), (8, 11), (8, 12), (8, 13),\\
		(11, 7), (11, 8), (11, 9), (11, 10), (11, 12), (11, 13)
	\end{array}
	\right]
	\]
\end{tcolorbox}

\begin{tcolorbox}[title=Is Complete Subgraph]
	This is an undirected graph with the following edges:\\
	From node 0 to node 1 with weight 4;\\
	From node 0 to node 2 with weight 1;\\
	From node 0 to node 3 with weight 2;\\
	From node 0 to node 4 with weight 5;\\
	From node 0 to node 5 with weight 4;\\
	From node 0 to node 6 with weight 3;\\
	From node 1 to node 2 with weight 1;\\
	From node 1 to node 3 with weight 2;\\
	From node 1 to node 4 with weight 2;\\
	From node 1 to node 5 with weight 1;\\
	From node 1 to node 6 with weight 1;\\
	From node 2 to node 3 with weight 1;\\
	From node 2 to node 4 with weight 4;\\
	From node 2 to node 5 with weight 3;\\
	From node 2 to node 6 with weight 4;\\
	From node 3 to node 4 with weight 1;\\
	From node 3 to node 5 with weight 1;\\
	From node 3 to node 6 with weight 5;\\
	From node 4 to node 5 with weight 4;\\
	From node 4 to node 6 with weight 2;\\
	From node 5 to node 6 with weight 1;\\
	From node 7 to node 8 with weight 3;\\
	From node 7 to node 9 with weight 5;\\
	From node 7 to node 10 with weight 5;\\
	From node 7 to node 11 with weight 2;\\
	From node 7 to node 12 with weight 2;\\
	From node 7 to node 13 with weight 4;\\
	From node 8 to node 9 with weight 5;\\
	From node 8 to node 10 with weight 1;\\
	From node 8 to node 11 with weight 4;\\
	From node 8 to node 12 with weight 5;\\
	From node 8 to node 13 with weight 3;\\
	From node 9 to node 10 with weight 3;\\
	From node 9 to node 11 with weight 2;\\
	From node 9 to node 12 with weight 2;\\
	From node 9 to node 13 with weight 1;\\
	From node 10 to node 11 with weight 1;\\
	From node 10 to node 12 with weight 2;\\
	From node 10 to node 13 with weight 4;\\
	From node 11 to node 12 with weight 5;\\
	From node 11 to node 13 with weight 5;\\
	From node 12 to node 13 with weight 5;\\\\
	\textbf{Q}: Given the nodes [0, 1, 6], determine if they form a complete subgraph. List the answer directly in the format of 'Yes' or 'No'.\\
	\textbf{A}: Yes
\end{tcolorbox}

\begin{tcolorbox}[title=Find Highest Degree Neighbor of Neighbors]
	This is an undirected graph with the following edges:\\
	From node 0 to node 1 with weight 4;\\
	From node 0 to node 2 with weight 1;\\
	From node 0 to node 3 with weight 2;\\
	From node 0 to node 4 with weight 5;\\
	From node 0 to node 5 with weight 4;\\
	From node 0 to node 6 with weight 3;\\
	From node 1 to node 2 with weight 1;\\
	From node 1 to node 3 with weight 2;\\
	From node 1 to node 4 with weight 2;\\
	From node 1 to node 5 with weight 1;\\
	From node 1 to node 6 with weight 1;\\
	From node 2 to node 3 with weight 1;\\
	From node 2 to node 4 with weight 4;\\
	From node 2 to node 5 with weight 3;\\
	From node 2 to node 6 with weight 4;\\
	From node 3 to node 4 with weight 1;\\
	From node 3 to node 5 with weight 1;\\
	From node 3 to node 6 with weight 5;\\
	From node 4 to node 5 with weight 4;\\
	From node 4 to node 6 with weight 2;\\
	From node 5 to node 6 with weight 1;\\
	From node 7 to node 8 with weight 3;\\
	From node 7 to node 9 with weight 5;\\
	From node 7 to node 10 with weight 5;\\
	From node 7 to node 11 with weight 2;\\
	From node 7 to node 12 with weight 2;\\
	From node 7 to node 13 with weight 4;\\
	From node 8 to node 9 with weight 5;\\
	From node 8 to node 10 with weight 1;\\
	From node 8 to node 11 with weight 4;\\
	From node 8 to node 12 with weight 5;\\
	From node 8 to node 13 with weight 3;\\
	From node 9 to node 10 with weight 3;\\
	From node 9 to node 11 with weight 2;\\
	From node 9 to node 12 with weight 2;\\
	From node 9 to node 13 with weight 1;\\
	From node 10 to node 11 with weight 1;\\
	From node 10 to node 12 with weight 2;\\
	From node 10 to node 13 with weight 4;\\
	From node 11 to node 12 with weight 5;\\
	From node 11 to node 13 with weight 5;\\
	From node 12 to node 13 with weight 5;\\\\
	\textbf{Q}: Given the node 9, find the neighbor's neighbor with the highest degree. List the answer directly as the node id.\\
	\textbf{A}: 8
\end{tcolorbox}

\begin{tcolorbox}[title=Find K-Order Neighbors]
	This is an undirected graph with the following edges:\\
	From node 0 to node 1 with weight 4;\\
	From node 0 to node 2 with weight 1;\\
	From node 0 to node 3 with weight 2;\\
	From node 0 to node 4 with weight 5;\\
	From node 0 to node 5 with weight 4;\\
	From node 0 to node 6 with weight 3;\\
	From node 1 to node 2 with weight 1;\\
	From node 1 to node 3 with weight 2;\\
	From node 1 to node 4 with weight 2;\\
	From node 1 to node 5 with weight 1;\\
	From node 1 to node 6 with weight 1;\\
	From node 2 to node 3 with weight 1;\\
	From node 2 to node 4 with weight 4;\\
	From node 2 to node 5 with weight 3;\\
	From node 2 to node 6 with weight 4;\\
	From node 3 to node 4 with weight 1;\\
	From node 3 to node 5 with weight 1;\\
	From node 3 to node 6 with weight 5;\\
	From node 4 to node 5 with weight 4;\\
	From node 4 to node 6 with weight 2;\\
	From node 5 to node 6 with weight 1;\\
	From node 7 to node 8 with weight 3;\\
	From node 7 to node 9 with weight 5;\\
	From node 7 to node 10 with weight 5;\\
	From node 7 to node 11 with weight 2;\\
	From node 7 to node 12 with weight 2;\\
	From node 7 to node 13 with weight 4;\\
	From node 8 to node 9 with weight 5;\\
	From node 8 to node 10 with weight 1;\\
	From node 8 to node 11 with weight 4;\\
	From node 8 to node 12 with weight 5;\\
	From node 8 to node 13 with weight 3;\\
	From node 9 to node 10 with weight 3;\\
	From node 9 to node 11 with weight 2;\\
	From node 9 to node 12 with weight 2;\\
	From node 9 to node 13 with weight 1;\\
	From node 10 to node 11 with weight 1;\\
	From node 10 to node 12 with weight 2;\\
	From node 10 to node 13 with weight 4;\\
	From node 11 to node 12 with weight 5;\\
	From node 11 to node 13 with weight 5;\\
	From node 12 to node 13 with weight 5;\\\\
	\textbf{Q}: Given the node 7, find all its 2-order neighbors. Note that the 2-order neighbors do not include the 1-order neighbors, and so on. List the answers in the format of '[1, 2, ...]'.\\
	\textbf{A}: []
\end{tcolorbox}

\begin{tcolorbox}[title=Find Direct Neighbors of Specified Node]
	This is an undirected graph with the following edges:\\
	From node 0 to node 1 with weight 4;\\
	From node 0 to node 2 with weight 1;\\
	From node 0 to node 3 with weight 2;\\
	From node 0 to node 4 with weight 5;\\
	From node 0 to node 5 with weight 4;\\
	From node 0 to node 6 with weight 3;\\
	From node 1 to node 2 with weight 1;\\
	From node 1 to node 3 with weight 2;\\
	From node 1 to node 4 with weight 2;\\
	From node 1 to node 5 with weight 1;\\
	From node 1 to node 6 with weight 1;\\
	From node 2 to node 3 with weight 1;\\
	From node 2 to node 4 with weight 4;\\
	From node 2 to node 5 with weight 3;\\
	From node 2 to node 6 with weight 4;\\
	From node 3 to node 4 with weight 1;\\
	From node 3 to node 5 with weight 1;\\
	From node 3 to node 6 with weight 5;\\
	From node 4 to node 5 with weight 4;\\
	From node 4 to node 6 with weight 2;\\
	From node 5 to node 6 with weight 1;\\
	From node 7 to node 8 with weight 3;\\
	From node 7 to node 9 with weight 5;\\
	From node 7 to node 10 with weight 5;\\
	From node 7 to node 11 with weight 2;\\
	From node 7 to node 12 with weight 2;\\
	From node 7 to node 13 with weight 4;\\
	From node 8 to node 9 with weight 5;\\
	From node 8 to node 10 with weight 1;\\
	From node 8 to node 11 with weight 4;\\
	From node 8 to node 12 with weight 5;\\
	From node 8 to node 13 with weight 3;\\
	From node 9 to node 10 with weight 3;\\
	From node 9 to node 11 with weight 2;\\
	From node 9 to node 12 with weight 2;\\
	From node 9 to node 13 with weight 1;\\
	From node 10 to node 11 with weight 1;\\
	From node 10 to node 12 with weight 2;\\
	From node 10 to node 13 with weight 4;\\
	From node 11 to node 12 with weight 5;\\
	From node 11 to node 13 with weight 5;\\
	From node 12 to node 13 with weight 5;\\\\
	\textbf{Q}: Given the node 11, find its neighbors that are directly connected to node 3. List the answers in the format of '[1, 2, ...]'.\\
	\textbf{A}: []
\end{tcolorbox}

\begin{tcolorbox}[title=Find Connected Neighbor Pairs]
	This is an undirected graph with the following edges:\\
	From node 0 to node 1 with weight 4;\\
	From node 0 to node 2 with weight 1;\\
	From node 0 to node 3 with weight 2;\\
	From node 0 to node 4 with weight 5;\\
	From node 0 to node 5 with weight 4;\\
	From node 0 to node 6 with weight 3;\\
	From node 1 to node 2 with weight 1;\\
	From node 1 to node 3 with weight 2;\\
	From node 1 to node 4 with weight 2;\\
	From node 1 to node 5 with weight 1;\\
	From node 1 to node 6 with weight 1;\\
	From node 2 to node 3 with weight 1;\\
	From node 2 to node 4 with weight 4;\\
	From node 2 to node 5 with weight 3;\\
	From node 2 to node 6 with weight 4;\\
	From node 3 to node 4 with weight 1;\\
	From node 3 to node 5 with weight 1;\\
	From node 3 to node 6 with weight 5;\\
	From node 4 to node 5 with weight 4;\\
	From node 4 to node 6 with weight 2;\\
	From node 5 to node 6 with weight 1;\\
	From node 7 to node 8 with weight 3;\\
	From node 7 to node 9 with weight 5;\\
	From node 7 to node 10 with weight 5;\\
	From node 7 to node 11 with weight 2;\\
	From node 7 to node 12 with weight 2;\\
	From node 7 to node 13 with weight 4;\\
	From node 8 to node 9 with weight 5;\\
	From node 8 to node 10 with weight 1;\\
	From node 8 to node 11 with weight 4;\\
	From node 8 to node 12 with weight 5;\\
	From node 8 to node 13 with weight 3;\\
	From node 9 to node 10 with weight 3;\\
	From node 9 to node 11 with weight 2;\\
	From node 9 to node 12 with weight 2;\\
	From node 9 to node 13 with weight 1;\\
	From node 10 to node 11 with weight 1;\\
	From node 10 to node 12 with weight 2;\\
	From node 10 to node 13 with weight 4;\\
	From node 11 to node 12 with weight 5;\\
	From node 11 to node 13 with weight 5;\\
	From node 12 to node 13 with weight 5;\\\\
	\textbf{Q}: Given the node 2, find all connected pairs among its neighbors. List the answers in the format of '[(1, 2), (3, 4), ...]'.\\
	\textbf{A}: \[
	\left[
	\begin{array}{c}
		(0, 1), (0, 3), (0, 4), (0, 5), (0, 6),\\
		(1, 3), (1, 4), (1, 5), (1, 6),\\
		(3, 4), (3, 5), (3, 6),\\
		(4, 5), (4, 6),\\
		(5, 6)
	\end{array}
	\right]
	\]
\end{tcolorbox}

\begin{tcolorbox}[title=Find Common Neighbors of Edge Nodes]
	This is an undirected graph with the following edges:\\
	From node 0 to node 1 with weight 4;\\
	From node 0 to node 2 with weight 1;\\
	From node 0 to node 3 with weight 2;\\
	From node 0 to node 4 with weight 5;\\
	From node 0 to node 5 with weight 4;\\
	From node 0 to node 6 with weight 3;\\
	From node 1 to node 2 with weight 1;\\
	From node 1 to node 3 with weight 2;\\
	From node 1 to node 4 with weight 2;\\
	From node 1 to node 5 with weight 1;\\
	From node 1 to node 6 with weight 1;\\
	From node 2 to node 3 with weight 1;\\
	From node 2 to node 4 with weight 4;\\
	From node 2 to node 5 with weight 3;\\
	From node 2 to node 6 with weight 4;\\
	From node 3 to node 4 with weight 1;\\
	From node 3 to node 5 with weight 1;\\
	From node 3 to node 6 with weight 5;\\
	From node 4 to node 5 with weight 4;\\
	From node 4 to node 6 with weight 2;\\
	From node 5 to node 6 with weight 1;\\
	From node 7 to node 8 with weight 3;\\
	From node 7 to node 9 with weight 5;\\
	From node 7 to node 10 with weight 5;\\
	From node 7 to node 11 with weight 2;\\
	From node 7 to node 12 with weight 2;\\
	From node 7 to node 13 with weight 4;\\
	From node 8 to node 9 with weight 5;\\
	From node 8 to node 10 with weight 1;\\
	From node 8 to node 11 with weight 4;\\
	From node 8 to node 12 with weight 5;\\
	From node 8 to node 13 with weight 3;\\
	From node 9 to node 10 with weight 3;\\
	From node 9 to node 11 with weight 2;\\
	From node 9 to node 12 with weight 2;\\
	From node 9 to node 13 with weight 1;\\
	From node 10 to node 11 with weight 1;\\
	From node 10 to node 12 with weight 2;\\
	From node 10 to node 13 with weight 4;\\
	From node 11 to node 12 with weight 5;\\
	From node 11 to node 13 with weight 5;\\
	From node 12 to node 13 with weight 5;\\\\
	\textbf{Q}: Given the edge (1, 5), find all common neighbors of its two end nodes. List the answers in the format of '[1, 2, ...]'.\\
	\textbf{A}: [0, 2, 3, 4, 6]
\end{tcolorbox}

\end{document}